\documentclass[runningheads]{llncs}

\usepackage{xcolor}
\usepackage{hyperref}
\usepackage{graphicx}
\usepackage[noend,ruled,vlined,linesnumbered]{algorithm2e}

\usepackage{amsmath}
\usepackage{amssymb}
\usepackage{mathtools}

\newcommand{\Q}{\mathcal{Q}}
\newcommand{\range}{\mathrm{range}}
\newcommand{\opt}{\mathrm{opt}}
\newcommand{\ranges}{\mathrm{ranges}}
\newcommand{\loss}{\mathrm{loss}}
\newcommand{\abs}[1]{\left| #1\right|}

\newcommand{\R}{\mathcal{R}}
\newcommand{\Hstyled}{\mathcal{H}}

\newcommand{\eps}{\varepsilon}
\newcommand{\of}[1]{\left(#1\right)}
\newcommand{\br}[1]{\left\{#1\right\}}
\newcommand{\ellt}{\tilde{\ell}}
\newcommand{\REAL}{\ensuremath{\mathbb{R}}}
\DeclareMathOperator*{\argmin}{arg\,min}
\newcommand{\Approx}{\textsc{Approx}}

\newcommand{\Coreset}{\textsc{Coreset}}
\newcommand{\kmeanspp}{\textsc{kmeans++}}
\newcommand{\Bicriteria}{\textsc{Bi-Criteria}}
\newcommand{\closest}{\texttt{closest}}
\newcommand{\kmeans}{\texttt{Kmeans}}
\newcommand{\kmeansgpu}{\texttt{Kmeans-Gpu}}
\newcommand{\approxoncore}{\texttt{Approx-on-coreset}}
\newcommand{\gaussian}{\texttt{gaussian}}

\newcommand{\randU}{\texttt{RandomUnderSampler}}
\newcommand{\randO}{\texttt{RandomOverSampler}}
\newcommand{\nearO}{\texttt{NearMiss-1}}
\newcommand{\nearTw}{\texttt{NearMiss-2}}
\newcommand{\nearTh}{\texttt{NearMiss-3}}

\newcommand{\KMeansSMOTE}{\texttt{KMeansSMOTE}}
\newcommand{\SMOTE}{\texttt{SMOTE}}
\newcommand{\SMOTETomek}{\texttt{SMOTETomek}}

\newcommand{\spectral}{\texttt{spectral}}
\newcommand{\dePDDP}{\texttt{dePDDP}}
\newcommand{\ward}{\texttt{ward}}
\usepackage{fullpage}
\begin{document}
\title{Provable Imbalanced Point Clustering}


%
%
\author{David Denisov \inst{1} \and
Dan Feldman\inst{1} \and
Shlomi Dolev\inst{2} \and
Michael Segal\inst{2}
}
\authorrunning{David. et al.}
%
\institute{University of Haifa, Haifa, Israel.\\
E-mail: David Denisov: \email{daviddenisovphd@gmail.com}. \and
Ben-Gurion University of the Negev, Beer-Sheva, Israel}
\maketitle     
\begin{abstract}
We suggest efficient and provable methods to compute an approximation for imbalanced point clustering, that is, fitting $k$-centers to a set of points in $\REAL^d$, for any $d,k\geq 1$.
To this end, we utilize \emph{coresets}, which, in the context of the paper, are essentially weighted sets of points in $\REAL^d$ that approximate the fitting loss for every model in a given set, up to a multiplicative factor of $1\pm\varepsilon$.
We provide [Section~\ref{sec: tests} and Section~\ref{sec: appedix tests} in the appendix] experiments that show the empirical contribution of our suggested methods for real images (novel and reference), synthetic data, and real-world data.
We also propose choice clustering, which by combining clustering algorithms yields better performance than each one separately.
\end{abstract}

\section{Introduction}
Imbalanced clustering considers the problem of clustering data, with severe class distribution skews.
This problem is of significant importance since this event can occur in practice, and if it does occur algorithms that do not support this, such as $k$-means, would give faulty results.
An example of clustering with class imbalance and its effect on clustering via $k$-means (along with our proposed methods) is provided in Section~\ref{sec: motivation}.

\textbf{Previous work.}
A common method to tackle the problem of imbalanced clustering is by under-sampling or over-sampling, where, informally, the points from the too-large or too-small clusters are sampled to obtain equal-sized clusters synthetically.
To this end, the methods often require labels (or targets) for the points, which are not always given.
Specifically, in Section~\ref{test: clustering} we compare our methods to the implementation in~\cite{Imbalanced_learn} of works~\cite{kNN_imbalanced},\cite{SMOTE},\cite{kmeans_smote},\cite{SMOTETomek}, which all require labels (or targets).
For a thorough review of such methods, see~\cite{sampleing} and \cite{imbalanced_clustering_new}.

The problem of weighted clustering was previously considered, e.g. in~\cite{outliers-resistance}, where each centroid point has its corresponding weight.
Nonetheless, as in~\cite{outliers-resistance}, the weights of all the centroid points fitted are often set before the fitting, while in our case the the weight is a function of the cluster size and as such can not be preemptively set.
Moreover, the coreset proposed in~\cite{outliers-resistance} is significantly larger than the coreset which we provide in Theorem~\ref{th: coreset}; in~\cite{outliers-resistance} the coreset size has an exponential dependency on the number of clusters, while in our case the coreset size has a sub-quadratic dependency on the number of clusters.

\textbf{Novelty.}
In contrast to the aforementioned prior works, we propose to solve the problem by considering loss functions that attempt to give each cluster equal importance.
In particular, instead of attempting to minimize the total Euclidean distance to all the points (with square loss this is the goal of k-means), we attempt to minimize the sum (over the clusters) of the mean Euclidean distance to the points assigned to it.
This is formalized in Section~\ref{sec: Objective functions}.
As a result, our method does not require targets like most previous works (\cite{kNN_imbalanced},\cite{SMOTE},\cite{kmeans_smote},\cite{SMOTETomek}, e.t.c.) and has essentially identical and even better performance as demonstrated in the additional tests at Section~\ref{sec: appedix tests} of the appendix.

\subsection{Paper structure}
Our work has the following structure:

In Section~\ref{sec: Objective functions} we state the objective functions that we aim to minimize and provide motivation for this task in Section~\ref{sec: motivation}.
In Section~\ref{sec: approx} we state our main theoretical results, notably Lemma~\ref{l: Approx} and Theorem~\ref{th: coreset: premtive}, whose proof is provided in the appendix.

In Section~\ref{sec: tests} we provide experimental results, mostly on images.
Specifically:
In Section~\ref{mot: image} we demonstrate our results for image quantization of a ``semi-synthetic" image (a photo of a rectangle drawn on paper).
In Section~\ref{sec: Sk-learn comp} we provide a comparison to various clustering algorithms at Scikit-learn~\cite{scikit-learn}, inspired by a demonstration therein.
In Section~\ref{sec: choice} we present \emph{choice clustering} that entails computing clusters in various methods and choosing the best from the ones computed via some measure, such as the silhouette-score~\cite{Silhouettes}.
In Section~\ref{sec: choice examples} we demonstrate the effectiveness of \emph{choice clustering} by comparing it to our suggested clustering and $k$-means, where those are also the clustering methods considered in the \emph{choice clustering}, over real-world images (one provided by the authors and one a reference from the ``USC-SIPI Image Database"~\url{https://sipi.usc.edu/database/}).

In Section~\ref{sec: conclusion} we provide a conclusion to the paper, followed by Section~\ref{sec: future work} where we state future work and limitations.

In Section~\ref{sec: alg} we state our algorithms, divided as follows: 
Section~\ref{sec: bi}, where we state our proposed approximations to the loss function from Definition~\ref{def: loss function} in Algorithm~\ref{alg: Bicriteria}, Section~\ref{sec: core}, where we state our proposed coreset to the loss function from Definition~\ref{def: loss function 2} in  Algorithm~\ref{alg: Coreset}, and Section~\ref{sec: gap}, where we state the practical modifications in code done to our theoretically proven algorithms to improve their running time.

In Section~\ref{sec: analysis of approx} we prove Lemma~\ref{l: Approx}, i.e., the desired properties of Definition~\ref{def: approx}.

In Section~\ref{sec: bi proof} we provide proof of the desired properties of Algorithm~\ref{alg: Bicriteria}, split into Section~\ref{sec: bi proof robust}, where we state previous work on robust medians and prove efficient computation for our case, and Section~\ref{sec: bi proof 2} where utilizing Section~\ref{sec: bi proof robust} we prove desired properties of Algorithm~\ref{alg: Bicriteria}.

In Section~\ref{sec: analysis of Core} we prove Theorem~\ref{th: coreset: premtive} that states the desired properties of Algorithm~\ref{alg: Coreset}.
To this end, we define and bound the VC-dimension of the problem in Section~\ref{subsection - VC dim}, state previous works on the sensitivity of functions in Section~\ref{sec: sen}, and then prove Theorem~\ref{th: coreset: premtive} in Section~\ref{sec: analysis of Core}.

In Section~\ref{sec: appedix tests} we provide additional tests of our methods both on synthetic and real-world, as follows.
In Section~\ref{test: clustering} we provide an extensive comparison of our methods to various clustering methods including (but not limited to) ones from the imbalanced-learn library~\cite{Imbalanced_learn} which require labels/targets, which are supplied.
In Section~\ref{test: hierarchical} we provide a comparison of our methods incorporated into hierarchical clustering and existing hierarchical clustering methods.
\subsection{Objective functions} \label{sec: Objective functions}
\textbf{Notations. }
Throughout this paper we assume $k,d\geq 1$ are integers, and denote by $\REAL^d$ the union of $d$-dimensional real column vectors.
A \emph{weighted set} is a pair $(P,w)$ where $P$ is a finite set of points in $\REAL^d$ and $w: P \to [0,\infty)$ is a \emph{weights function}.
For simplicity, we denote $\log(x):=\log_2(x)$.
In this work, we assume that all the input sets are finite and non-empty.

A method to define the problem of imbalanced clustering is to minimize the mean (over the clusters) of the cluster's variance.
That is formally stated as follows.
Note that we define all the loss functions for general sets $C\subset \REAL^d$, but aim to minimize the loss over such sets of size at most $k$.
\begin{definition}[Loss function]\label{def: loss function}
Let $P\subset \REAL^d$ be a set of points.
We define its \emph{fitting loss} to a set $C\subset \REAL^d$, as
\begin{equation}
    \ell\big(P,C\big) = 
    \sum_{c\in C} \frac{1}{|P_c|+1} \sum_{p\in P_c} \| p-c\|_2,
\end{equation}
where $\displaystyle \br{P_c}_{c\in C}$ is a partition of $P$, such that every $p\in P$ is in $P_i$ if $\displaystyle c \in \argmin_{c\in C} \| p-c\|_2$. Ties broken arbitrarily. That is, for every $c\in C$ we set $P_c$ as the points in $P$ that are closest to $c$.
Our goal is to minimize $\ell\big(P,C\big)$ over every set $C$ of $|C|=k$ centers.
\end{definition}

To obtain a provable compression and more refined approximation, we introduce the following relaxation to the problem.
The use of the following poly-logarithmic function enables a poly-logarithmic sized compression scheme (i.e., coreset, see Definition~\ref{Def:coreset}) for any given $d$ and $k$ (it is used in the proof of Theorem~\ref{th: coreset proof} specifically at E.q.~[\ref{sen:2}--\ref{sen:6.2}]) while preserving that the impact of each point $p$ on the loss is inversely (but not linearly) correlated to the size (in number of points) of the cluster that $p$ belongs to.
\begin{definition}[Relaxed loss function]\label{def: loss function 2}
Let $P\subset \REAL^d$ be a set of points.
We define the \emph{relaxed fitting loss} of a set $C\subset \REAL^d$, as
\begin{equation}
    \ellt\big(P,C\big) = 
    \sum_{c\in C} \frac{1}{\log^2(|P_c|+1)} \sum_{p\in P_c} \| p-c\|_2,
\end{equation}
where $\displaystyle \br{P_c}_{c\in C}$ is a partition of $P$, such that every $p\in P$ is in $P_i$ if $\displaystyle c \in \argmin_{c\in C} \| p-c\|_2$. Ties broken arbitrarily. That is, for every $c\in C$ we set $P_c$ as the points in $P$ that are closest to $c$.
Our goal is to minimize $\ellt\big(P,C\big)$ over every set $C$ of $|C|=k$ centers.

More generally, for a weighted set $(C,w)$ we set the \emph{relaxed fitting loss} of $(C,w)$ as 
\begin{equation}
    \ellt\big(P,(C,w)\big) = 
    \sum_{c\in C} w(c)\frac{1}{\log^2(|P_c|+1)} \sum_{p\in P_c} \| p-c\|_2,
\end{equation}
where $\displaystyle \br{P_c}_{c\in C}$ is a partition of $P$ defined as above.
\end{definition}

\subsection{Motivation}\label{sec: motivation}
In the following section, we demonstrate a motivation for the suggested clustering method, and its relaxation via comparison to $k$-means, specifically $k$-means++~\cite{kmeans++}, as implemented by Scikit-Learn~\cite{scikit-learn}.
This would be done by considering the problem of clustering points in $\REAL^2$ where the clusters are very imbalanced, which would result in miss-classifying by $k$-means.

We utilized the following approximation, which is inspired by the centroid sets from~\cite{newframework}, to the problem stated in Definition~\ref{def: loss function}.
For the properties of this method see Section~\ref{sec: approx}.

\begin{definition}\label{alg: approx}
Let $(P,w)$ be a weighted set of size $n\geq k$.
That is $P\subset\REAL^d,|P|=n$ and $w:P\to \REAL$.
Let $\mathcal{P}_k$ be the union over all the subsets of size $k$ from $P$.
A set that minimized $\ellt\big((P,w),C\big)$ over every set $C\in \mathcal{P}_k$ is called the \emph{optimal solution in $\mathcal{P}_k$}.
We denote by $\opt(P,w,k)$ this \emph{optimal solution in $\mathcal{P}_k$}. Ties broken arbitrarily.
\end{definition}

Simply replacing $\ellt$ by $\ell$ in Definition~\ref{alg: approx} would not yield a provable approximation in general.
Nonetheless, in this section, we approximate a minimizer to the loss function in Definition~\ref{def: loss function} via such substitution.
In this section, for simplicity, we set $k=2$.

We consider the following data sets in $\REAL^2$, where we use different values for $n$, which are the union of $2$ uniform samples from discs generated as follows.\\
(i). A sample of $n$ ``inliers" points, where each point is chosen uniformly inside the unit disc.\\
(ii). A sample of $25$ ``outliers" points, where each point is chosen uniformly inside a disc of radius $0.1$, centered at $(2,0)$.

\begin{figure}[h!]
    \centering
    \includegraphics[width=0.3275\textwidth]{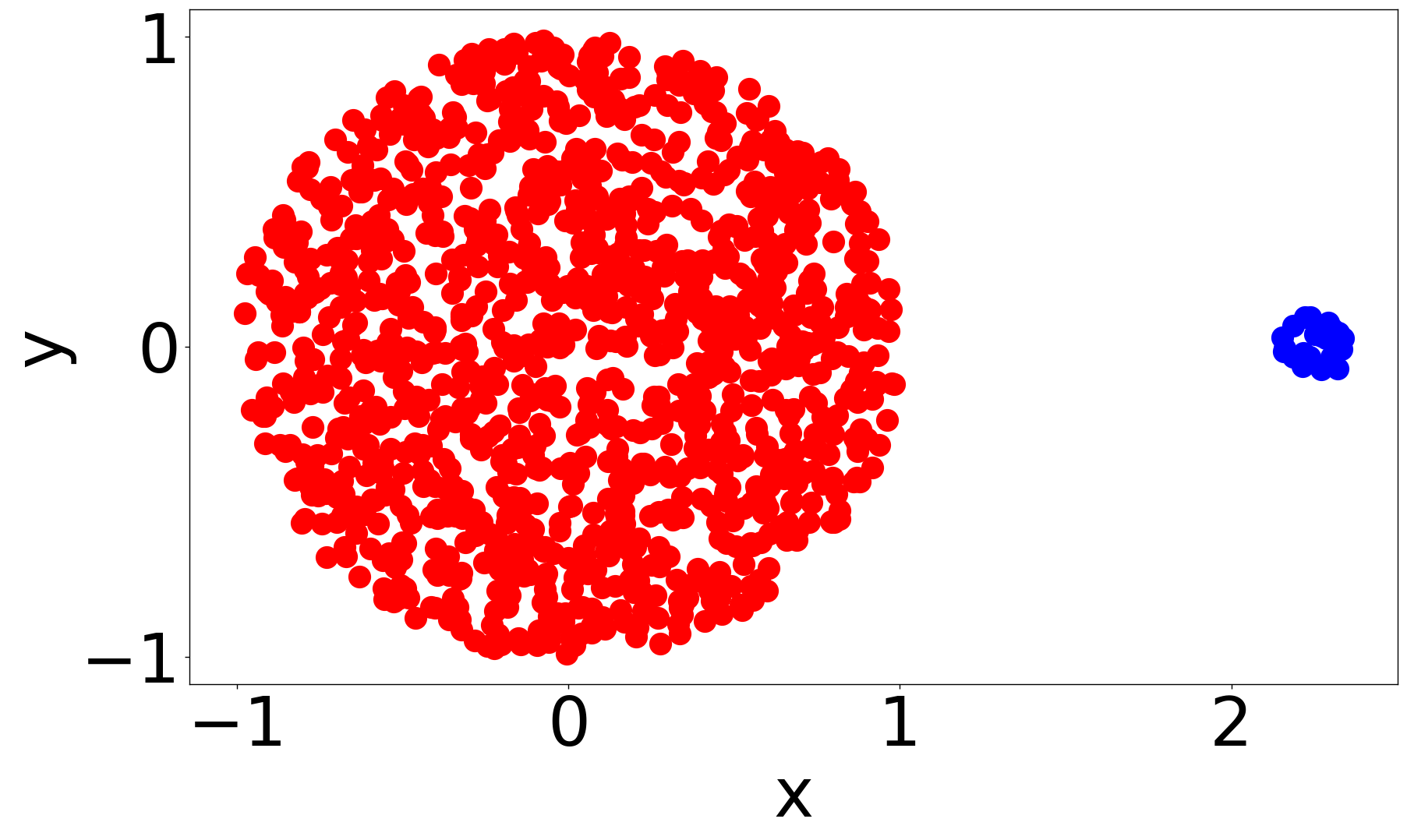}
    \includegraphics[width=0.3275\textwidth]{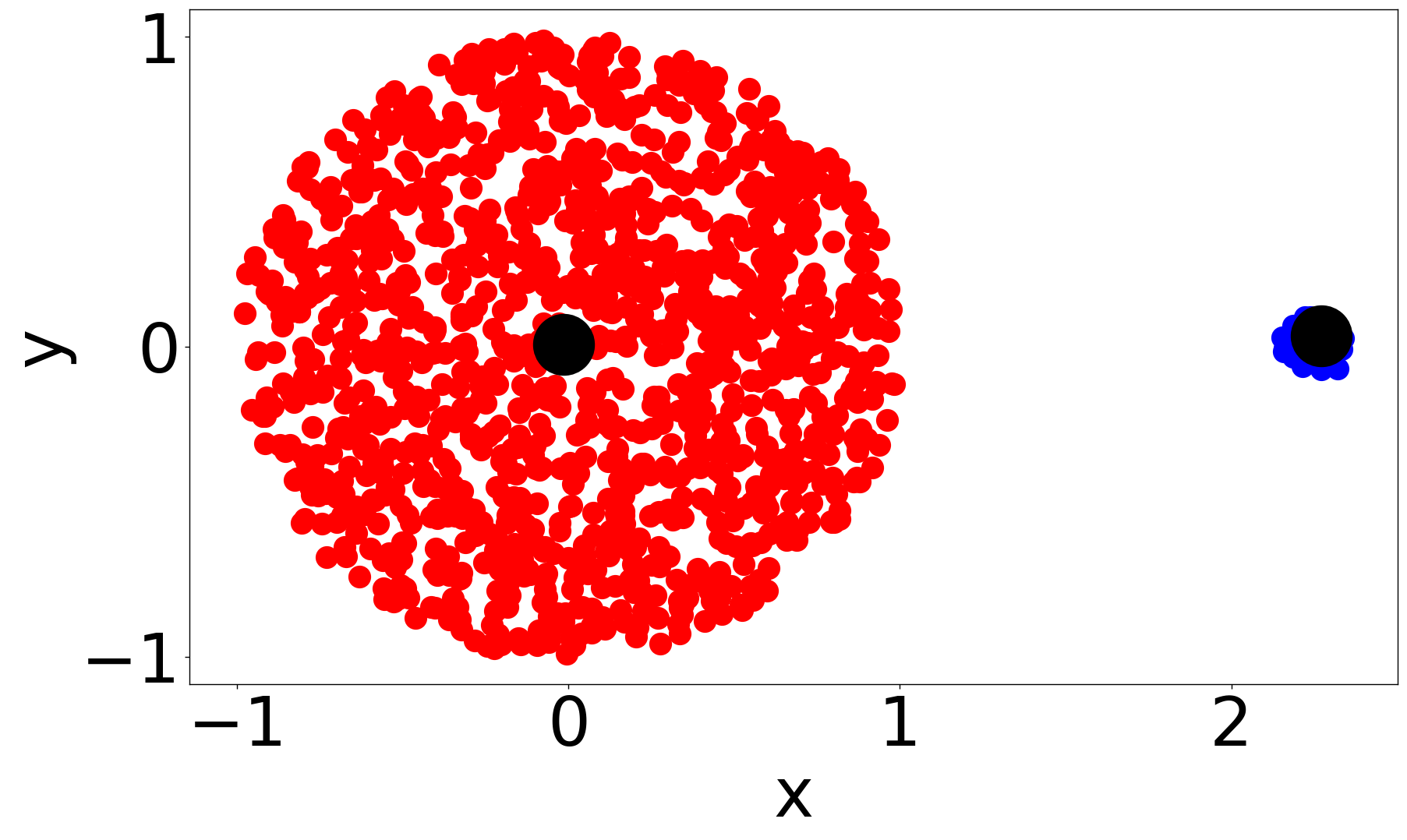}
    \includegraphics[width=0.3275\textwidth]{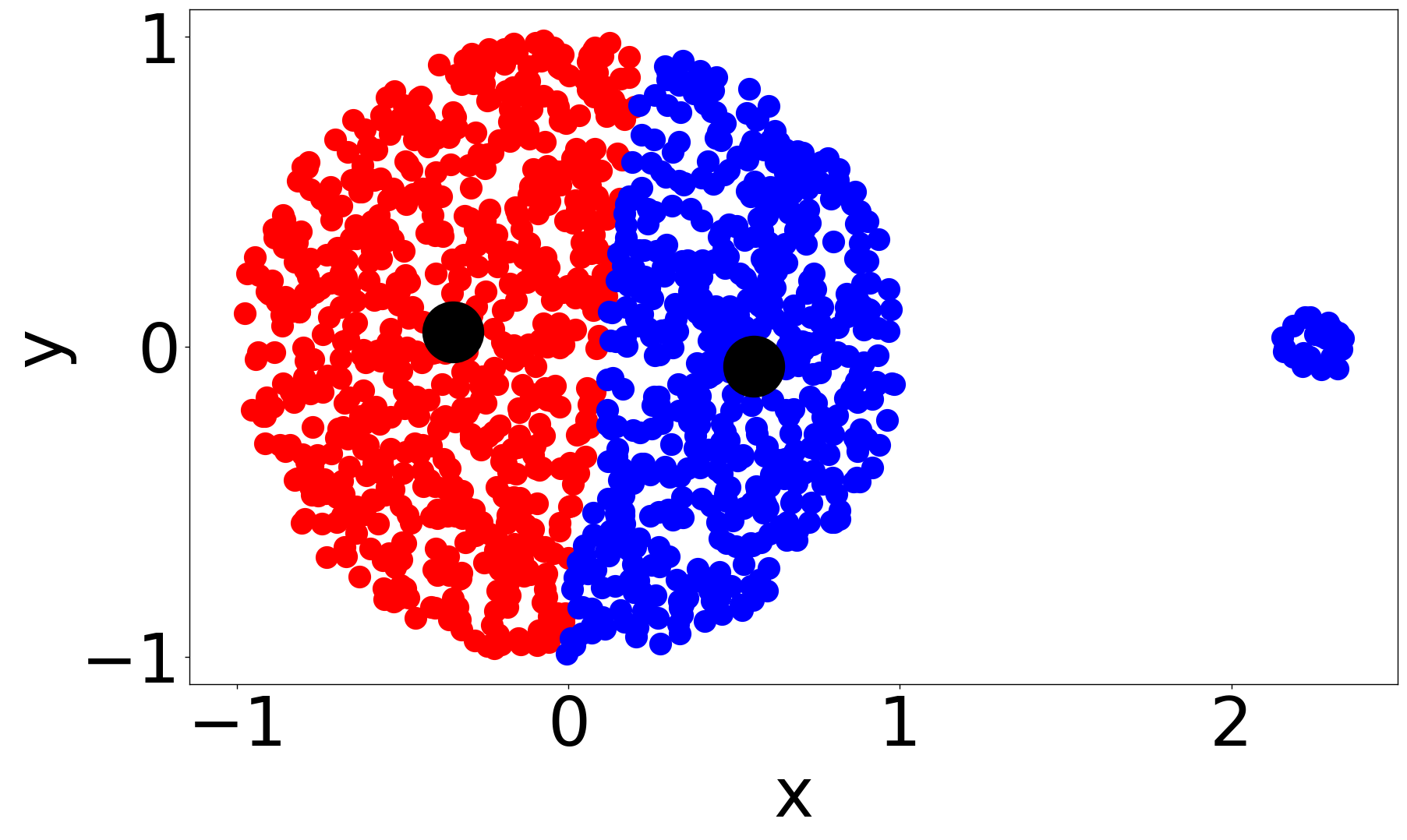}
    \caption{Motivation for minimizing the loss function of Definition~\ref{def: loss function}.
    \textbf{(left)}
    The left figure is the data generated for $n=1250$ ``inliers" points and 25 ``outliers", the color of each point corresponds to the set from which it was generated (outlier or inlier).
    \textbf{(middle)} the black dots are the two optimal centers according to our approximation to the minimizer of the loss function in Definition Definition~\ref{def: loss function}. The red points are closest to the first center, while the blue points are closest to the second center.
    \textbf{(middle)} the black dots are the two optimal centers according to our approximation to the minimizer of the loss function in Definition Definition~\ref{def: loss function}. The red points are closest to the first center, while the blue points are closest to the second center.
    (\textbf{(right)} the black dots are the two centers resulting by applying $k$-means++~\cite{kmeans++} with $k=2$ for all the points. Again, the red points are closest to the first center, while the blue points are closest to the second center.
    }
    \label{fig: motivation 1}
\end{figure}

In Figure~\ref{fig: motivation 1} we plot an example of the resulting the data generated, along with clustering provided by $k$-means and our solution to the objective function suggested in Definition~\ref{def: loss function}.

We had that $k$-means misclassified the original sample, and mixed part of the large cluster of ``inliers" to the small one of ``outliers".
On the other hand, our method has classified the clusters correctly.
It should be emphasized that this is due to the different objectives, as while our method had a lower loss for our defined loss, it had a noticeably larger loss for the mean squared error that $k$-means++ attempts to minimize.

\begin{figure}[h!] 
    \centering
    \includegraphics[width=0.3275\textwidth]{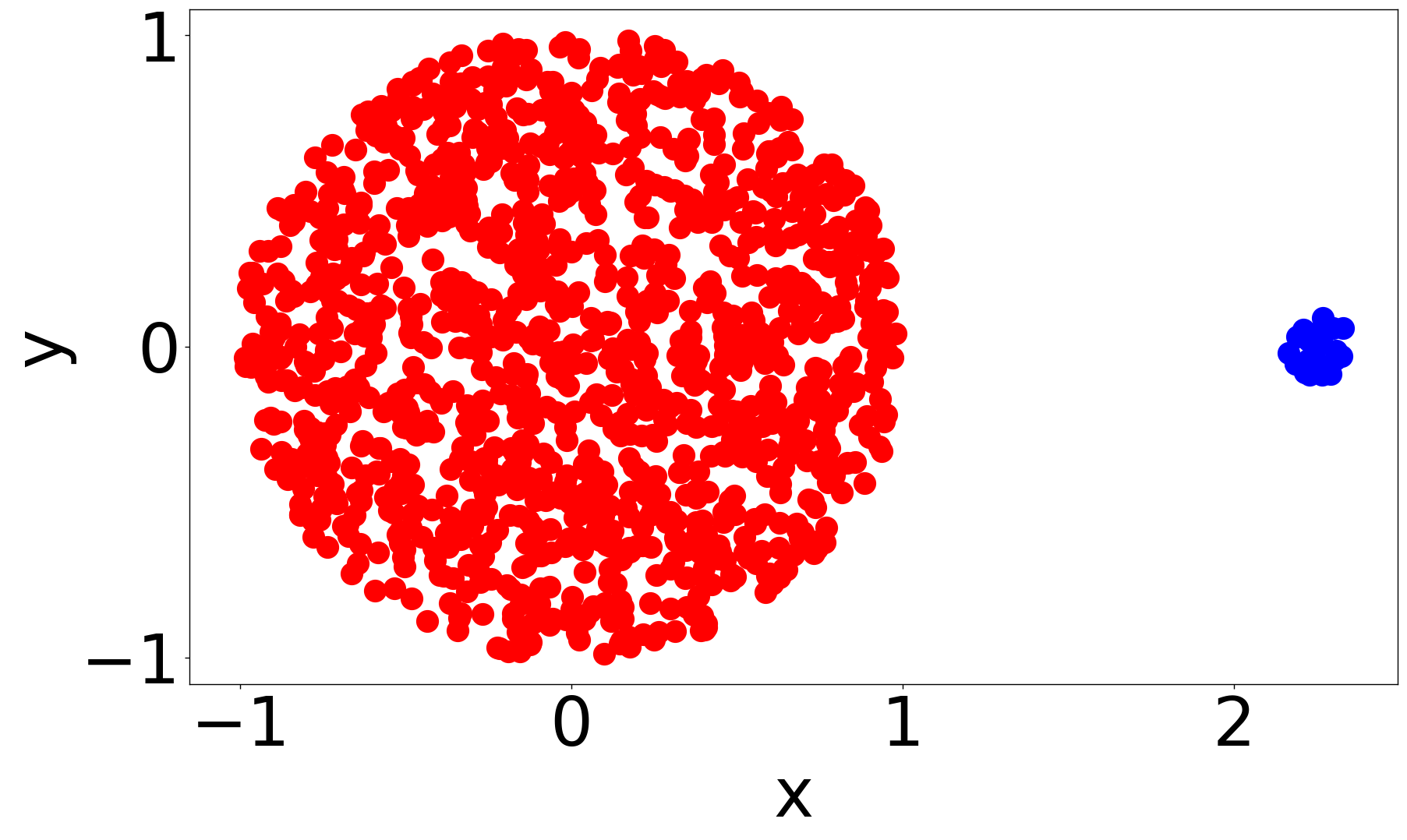}
    \includegraphics[width=0.3275\textwidth]{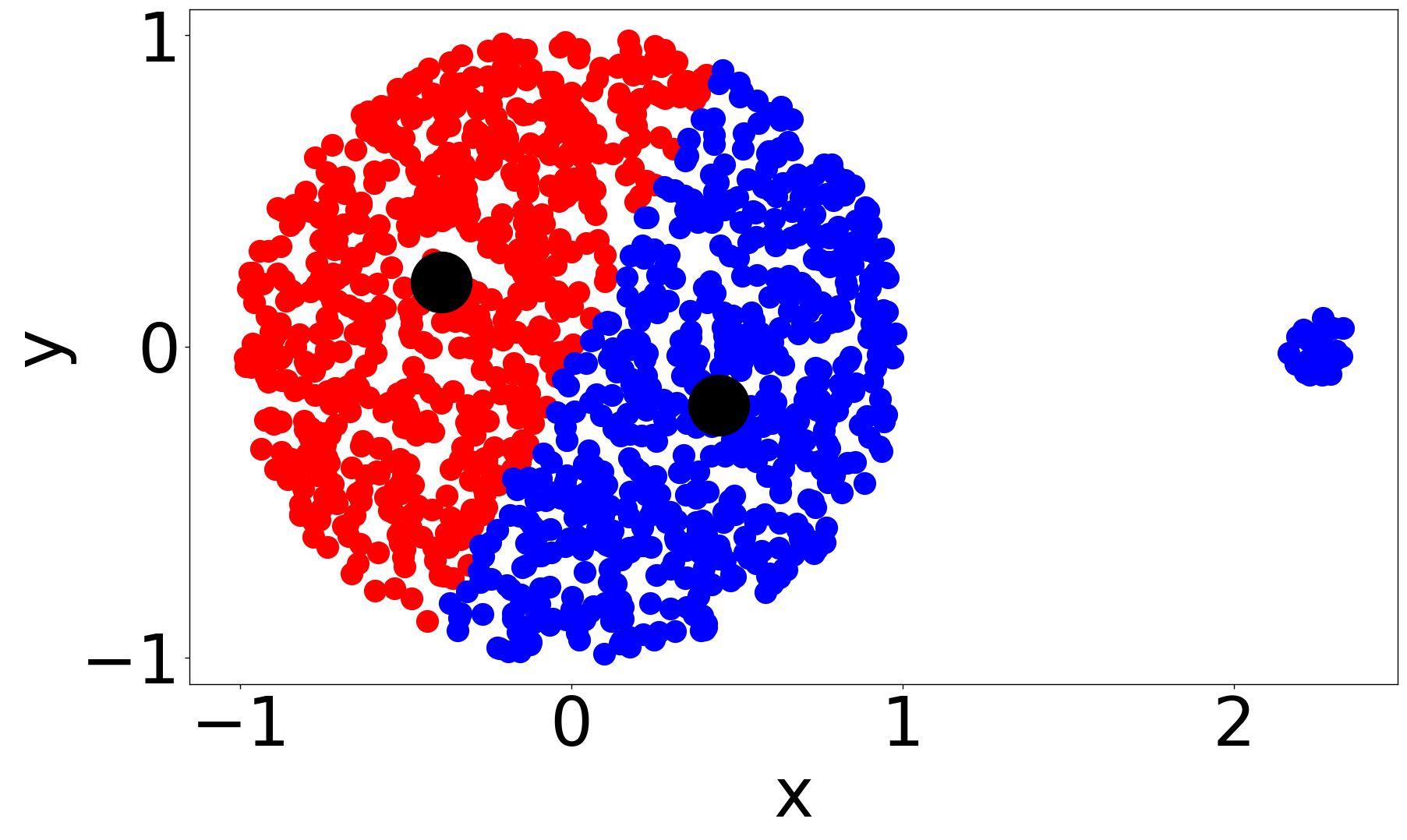}
    \includegraphics[width=0.3275\textwidth]{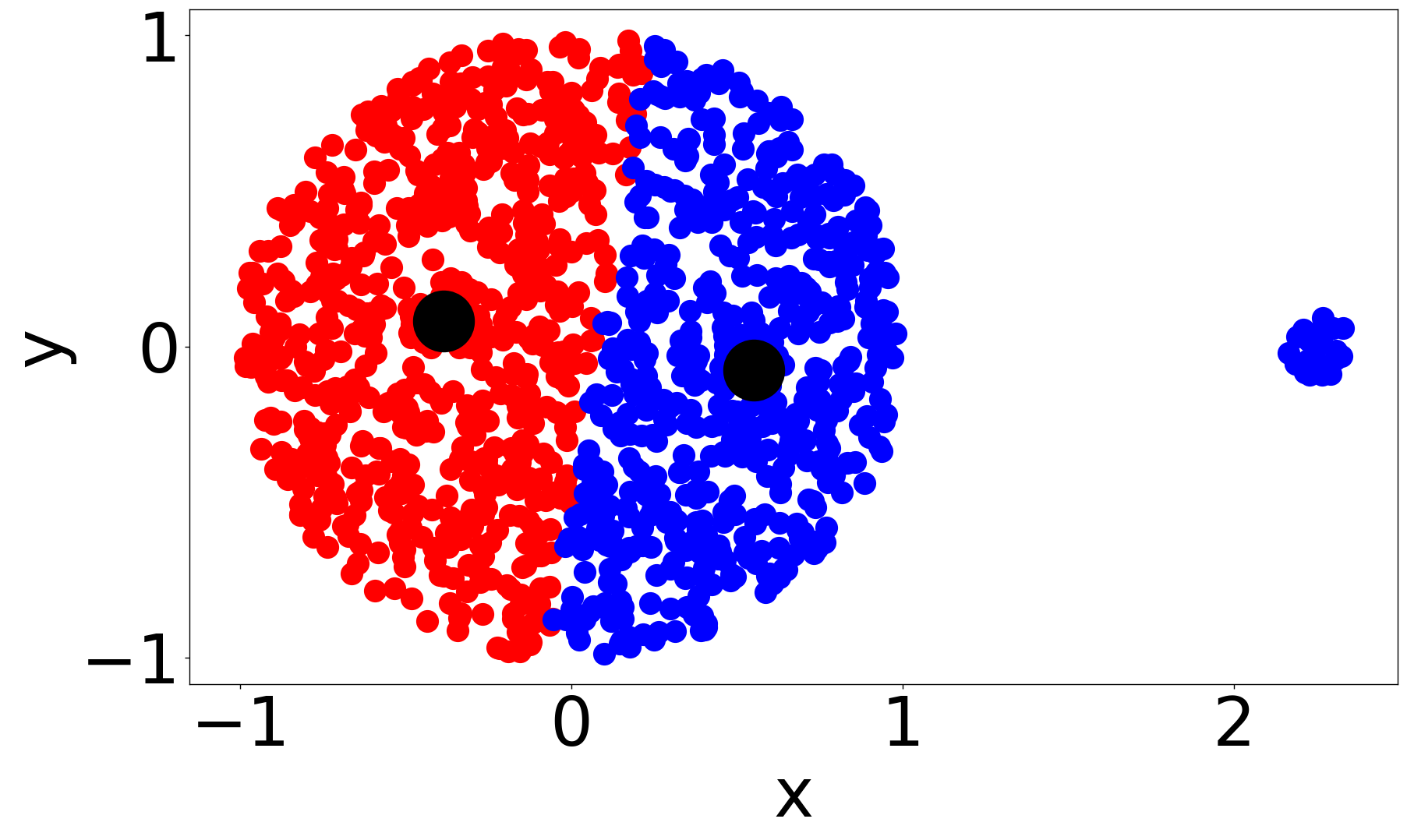}

    \includegraphics[width=0.3275\textwidth]{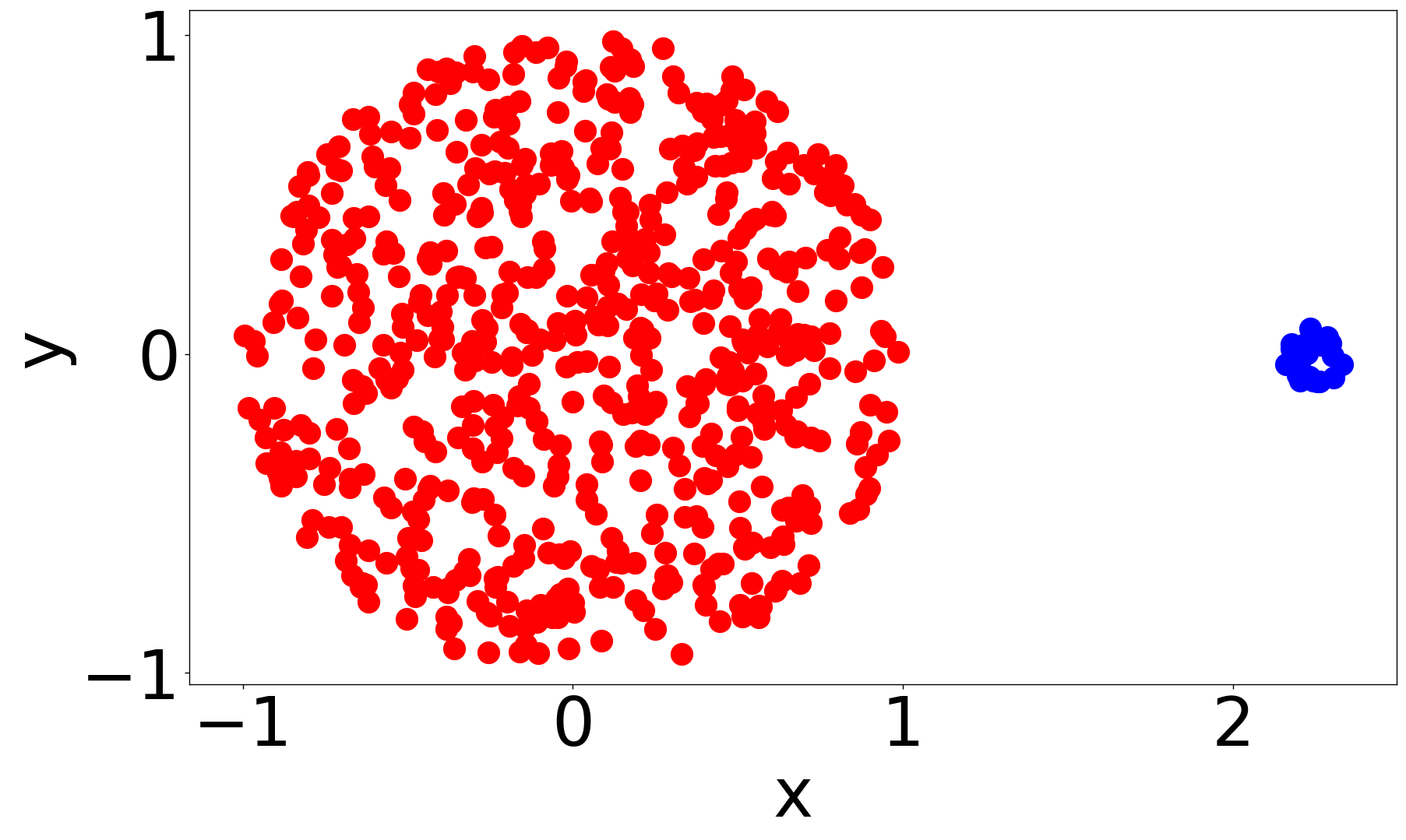}
    \includegraphics[width=0.3275\textwidth]{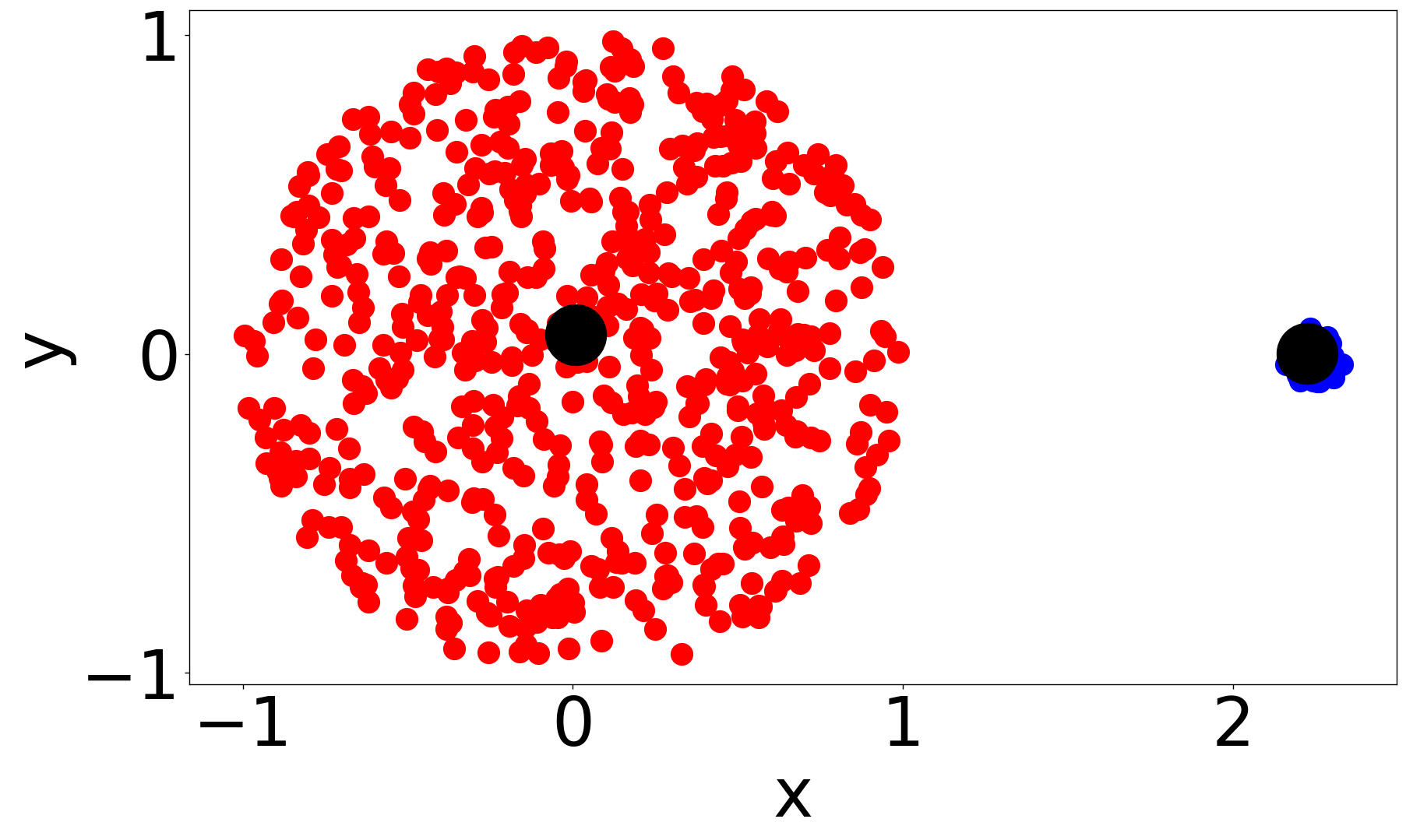}
    \includegraphics[width=0.3275\textwidth]{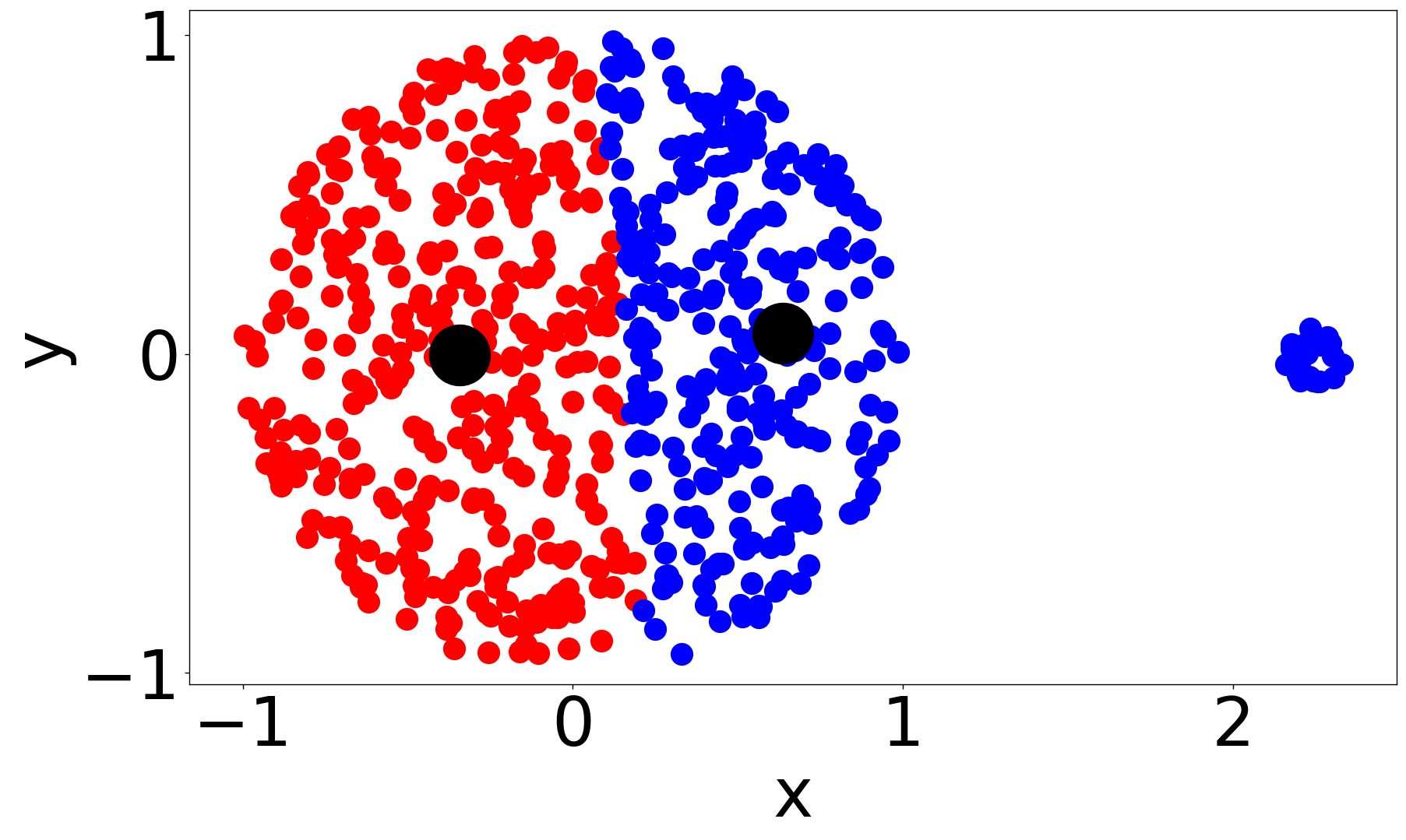}
    \caption{Motivation for the problem suggested in Definition~\ref{def: loss function 2}.
    The top row corresponds to $n=1250$ ``inliers" points along $25$ ``outliers" points, and the bottom row to $n=625$ `'inliers" points along $25$ ``outliers" points.
    The left columns are the data generated for the value of $n$, the color of each point corresponds to the set from which it was generated.
    In the right and middle columns, the black dots are the $2$ centers computed, and the points of each cluster are colored depending on which center they are closest to (red or blue).
    The right figure demonstrates the output of $k$-means++~\cite{kmeans++}.
    The middle figure demonstrates our approximations of the problem suggested in Definition~\ref{def: loss function 2}.
    }
    \label{fig: motivation 2}
\end{figure}

In the following test, we repeat the previous test, where we change our method to solve the problem in Definition~\ref{def: loss function} to a method to solve the problem in Definition~\ref{def: loss function 2}.
Results are provided in Figure~\ref{fig: motivation 2}.

For a large cluster size of $625$ points, we have that our approximation to the problem suggested in Definition~\ref{def: loss function 2} does classify the data correctly, while $k$-means++ fails.
However, when we increase the size of the larger cluster to $1250$ (as in the previous example), our objective function as suggested in Definition~\ref{def: loss function 2} does not partition the data correctly.
This demonstrates that, while the problem suggested in Definition~\ref{def: loss function 2} allows correct classification for imbalanced cases where $k$-means fails, it can still yield miss-classification for sufficiently large imbalances.

\section{Theoretical results} \label{sec: approx}
For presenting our theoretical results, we use the following notations and definitions.

\begin{definition} [$\alpha$-approximation]\label{def: approx}
Let $\alpha \geq 1$ and $P \subset \REAL^d$.
An $\alpha$-\emph{approximation} $C \subset \REAL^d$ of $P$ is a set of size $|C|=k$ such that 
\begin{equation}
\ellt(P,C) \leq \alpha \cdot \min_{C^*\subset \REAL^d, |C^*|=k} \ellt(P,C^*).
\end{equation}
\end{definition}
A coreset for the fitting problem from Definition~\ref{def: loss function 2} is defined as follows.
\begin{definition} [$\eps$-coreset] \label{Def:coreset}
Let $P \subseteq \REAL^d$ and $\eps > 0$ be an error parameter.
A pair $(C,w)$, where $C\subset \REAL^d$ and $w:C\to \REAL$, is an \emph{$\eps$-coreset} of $P$, if, for every set
$C' \subset \REAL^d$ of size $|C'|=k$, we have
\begin{equation}
\Big|\ellt(P,C') - \ellt\big((C,w),C'\big) \Big| \leq \eps \cdot \ellt(P,C').
\end{equation}
\end{definition}
\subsection{Main results}
The following lemma states the main properties of Definition~\ref{def: approx}, which was inspired by the centroid sets method in~\cite{newframework}.
For its proof see Lemma~\ref{l: Approx proof}.
\begin{lemma}\label{l: Approx}
Let $(P,w)$ be a weighted set of size $n\geq k$ where $w:P\to [0,\infty)$.
That is $P\subset\REAL^d$ and $|P|=n$.
Suppose that $\min_{C\subset\REAL^d,|C|=k} \ellt\big((P,w),C\big)$ exists.
Let $Q:=\Approx\big((P,w),k\big)$; see Definition~\ref{alg: approx}.
Then $Q$ is a $2 \log^2(1+n)$-approximation for $(P,w)$; see Definition~\ref{def: approx}.
Moreover, $Q$ can be computed in $O\of{n^{k+1}dk}$ time.
\end{lemma}
The following theorem states the existence of an efficient coreset construction.
Due to space limitations the formal statement, which states how to compute such a coreset, and the proof of the theorem are given at Theorem~\ref{th: coreset} in the appendix.
\begin{theorem} \label{th: coreset: premtive}
There is an algorithm that gets $P$, $k$, $\eps\in (0,1)$, $\delta\in (0,1/10]$, and returns a weighted set $(C,w)$ of size
\begin{equation}
|C|\in O\of{\frac{kd^3\log(k)\log^4(n)}{\eps^2} \left(\log\big(\log(k)\log(n)\big) + \log\left(\frac{1}{\delta}\right)\right)}
\subseteq O\of{\frac{kd^3\log^2k\log^5n+\log(1/\delta)}{\eps^2}},
\end{equation}
such that, with probability at least $1-\delta$, $C$ is an $\eps$-coreset for $P$. 
Moreover, $C$ can be computed in $O\big(ndk \log(1/\delta) \big)$ time.
\end{theorem}

\section{Experimental results}\label{sec: tests}
Using Python 3.8, we implemented a coreset construction algorithm that satisfies Theorem~\ref{th: coreset: premtive} and the approximation from Lemma~\ref{l: Approx}.
Our changes from the theoretically proven algorithms are stated in Section~\ref{sec: gap} after their pseudo-code.
In all of the following tests, we use a PC with an Intel Core i5-12400F, NVIDIA GTX 1660 SUPER (GPU), and 32GB of RAM.

For a set $P\subset \REAL^d$ of size $|P|\geq k$ we denote $\approxoncore(P,k):=Approx((C,w),k)$, where $(C,w)$ is the weighted set stated in Theorem~\ref{th: coreset: premtive}.

\subsection{Real world motivation: image quantization} \label{mot: image}
In the following example, we demonstrate that the previously stated class imbalance can occur in real-world tests.
Specifically, we demonstrate this in the task of image quantization via clustering.
It should be emphasized that our example can be quantized correctly via known existing methods.
This would be done as in OpenCV's~\cite{openCV} K-Means Clustering in OpenCV tutorial, that is, paraphrasing the tutorial:

There are 3 features, say, R, G, and B. 
Reshape the $n\times d \times 3$ image to a matrix of size $M\times 3$ ($M$ is the number of pixels in the image).
Compute two clusters (also R, G, B), details are below.
After the clustering, apply centroid values (it is also R, G, B) to all pixels, such that the resulting image will have a specified number of colors. 
Then reshape the result back to the shape of the original image, i.e., reshape to a $n\times d \times 3$ matrix.

In the following example, we took an image of a white page with a small blue rectangle drawn on it.
We consider two options to compute the two clusters stated in the quantization:\\
(i). Our suggested clustering, which is provided by $\approxoncore$.\\
(ii). The $k$-means approximation clustering, which is applying $k$-means++~\cite{kmeans++} as implemented by Scikit-Learn~\cite{scikit-learn}.

\begin{figure}[h]
\centering
\includegraphics[width=0.3275\textwidth]{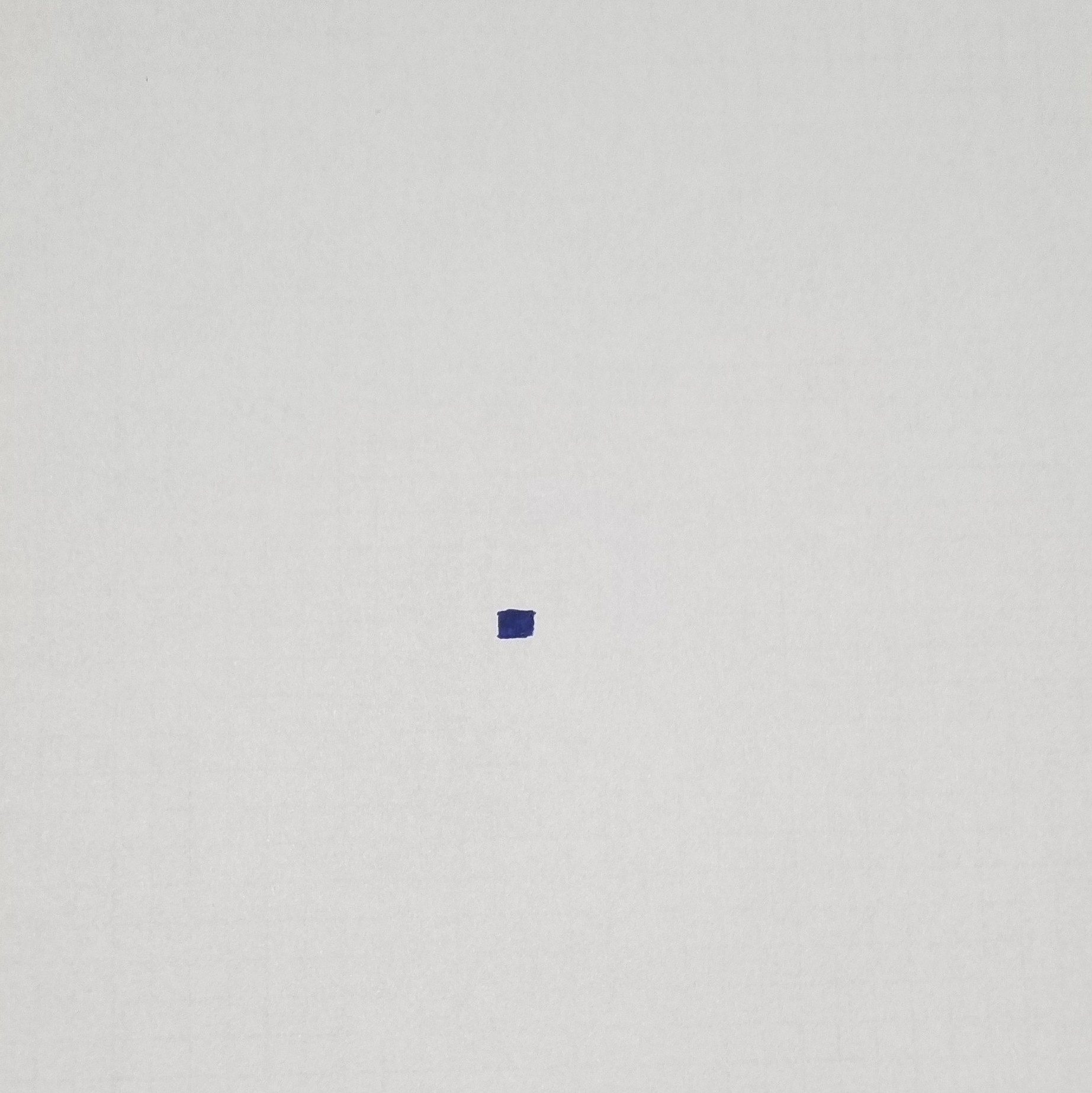}
    \includegraphics[width=0.3275\textwidth]{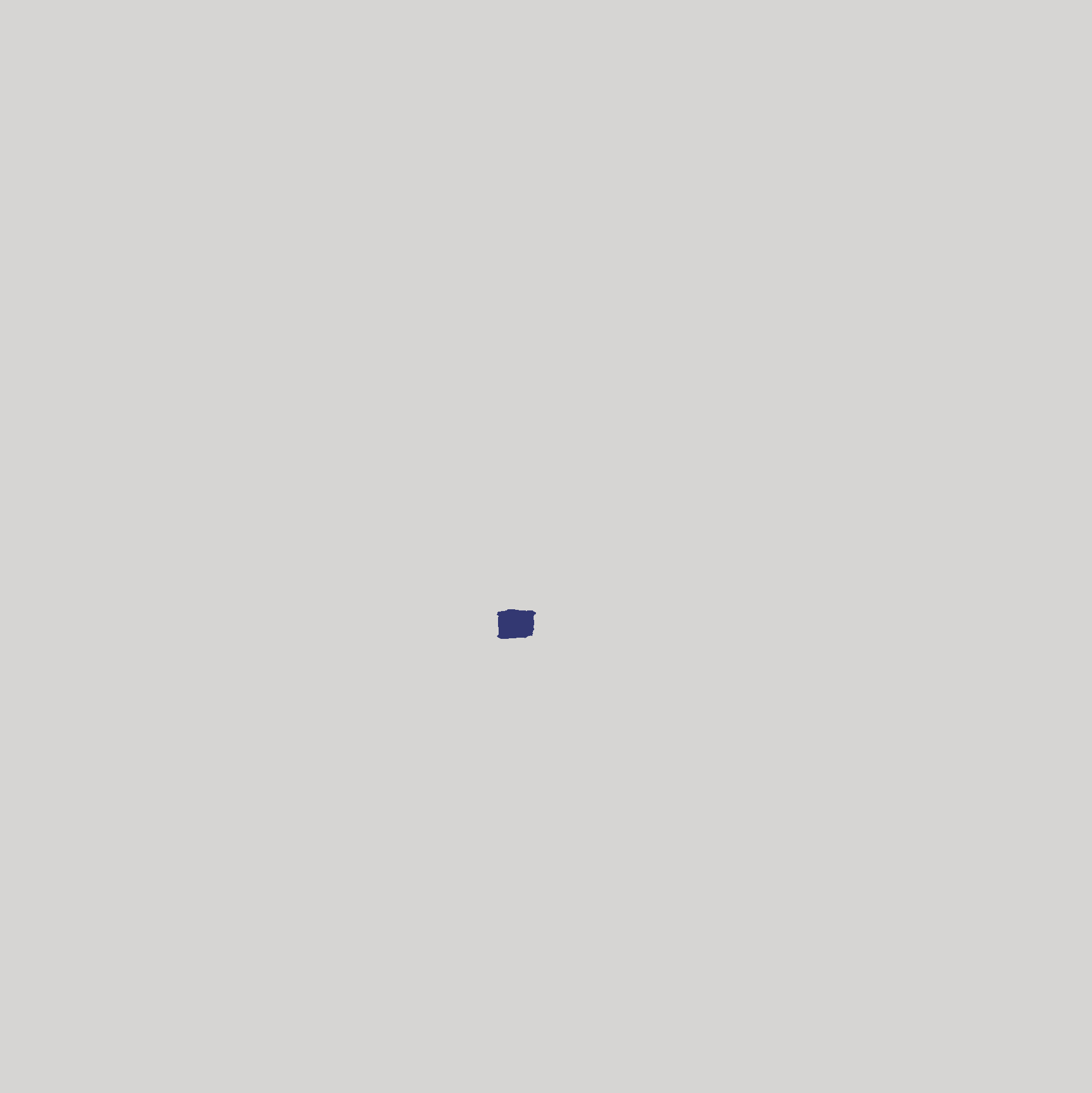}
    \includegraphics[width=0.3275\textwidth]{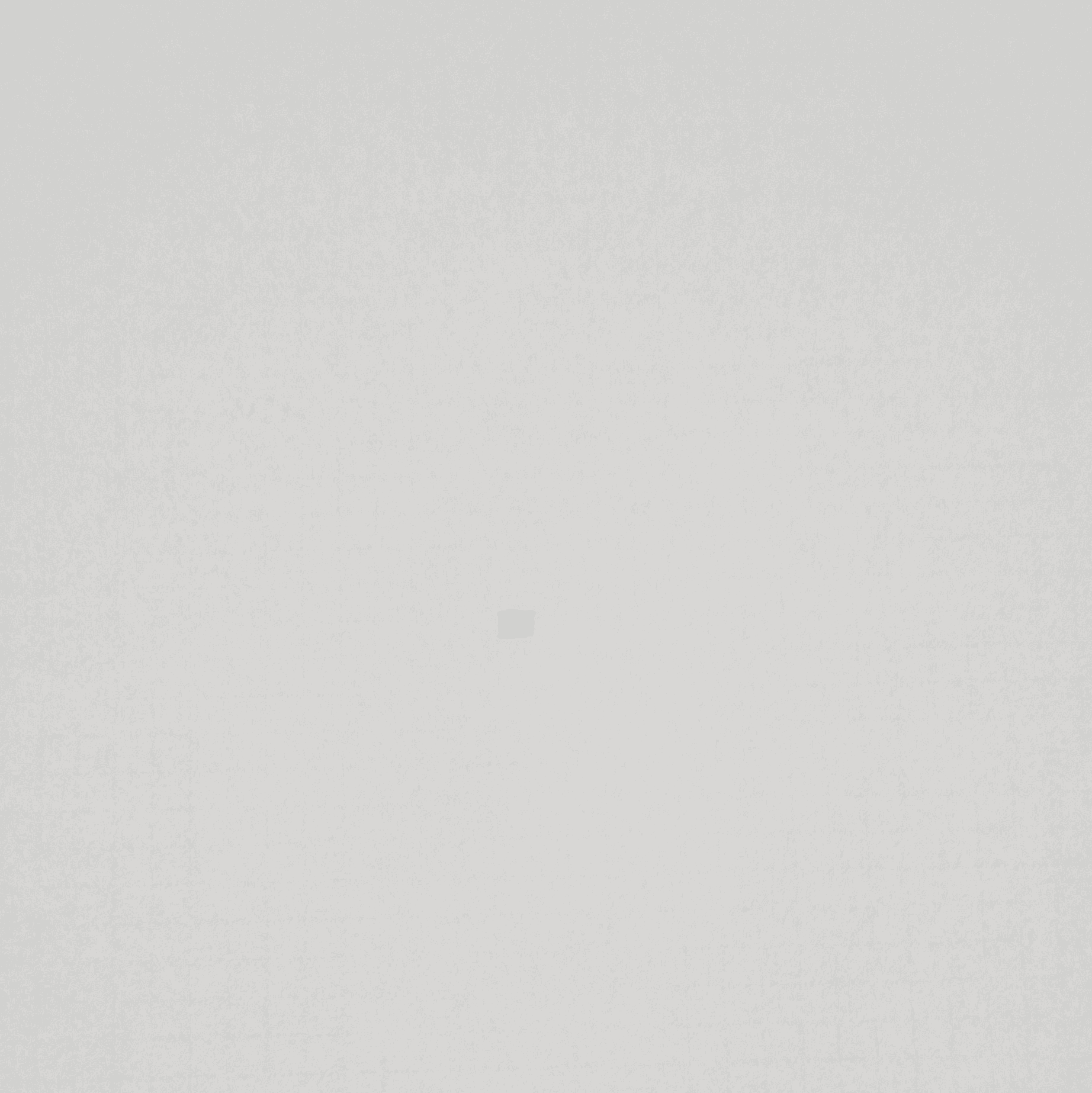}
    \caption{Results for Section~\ref{mot: image}.
    The images are (left to right): the input image to cluster, the result of our quantization, and the result of the Scikit-Learn quantization.
    Note that the right image contains a small near-gray rectangular in place of the blue rectangle of the original image.
    }
    \label{fig: image cluster}
\end{figure}

We refer to the image quantization corresponding to our clustering as our quantization, and similarly, to the image quantization corresponding to Scikit-Learn clustering as Scikit-Learn quantization.
The results of the quantization are provided in Figure~\ref{fig: image cluster}.

As can be seen, our quantization detected the rectangle as a distinct cluster from the page, while the Scikit-Learn quantization did not, and due to this it can barely be seen in the resulting quantization.

As stated in OpenCV's~\cite{openCV} K-Means Clustering in OpenCV tutorial, a motivation for such quantization is to reduce the memory requirement of the images.
The size of the produced image of our quantization is noticeably lower than the original image ($\sim$20kB in ``png" format v.s. $\sim$500kB in ``jpg" format and above 3MB in ``png" format), while no such noticeable decrease is observed for the Scikit-Learn quantization.
This suggests that the competing clustering classified the ``noise" of the image (for example, the inherent ``noise" of the paper page, camera noise, etc.), which is mostly random and as such does not yield any compression.

\subsection{Comparison to common clustering algorithms} \label{sec: Sk-learn comp}
\begin{figure}[h!] 
    \centering
    \includegraphics[width=\textwidth]{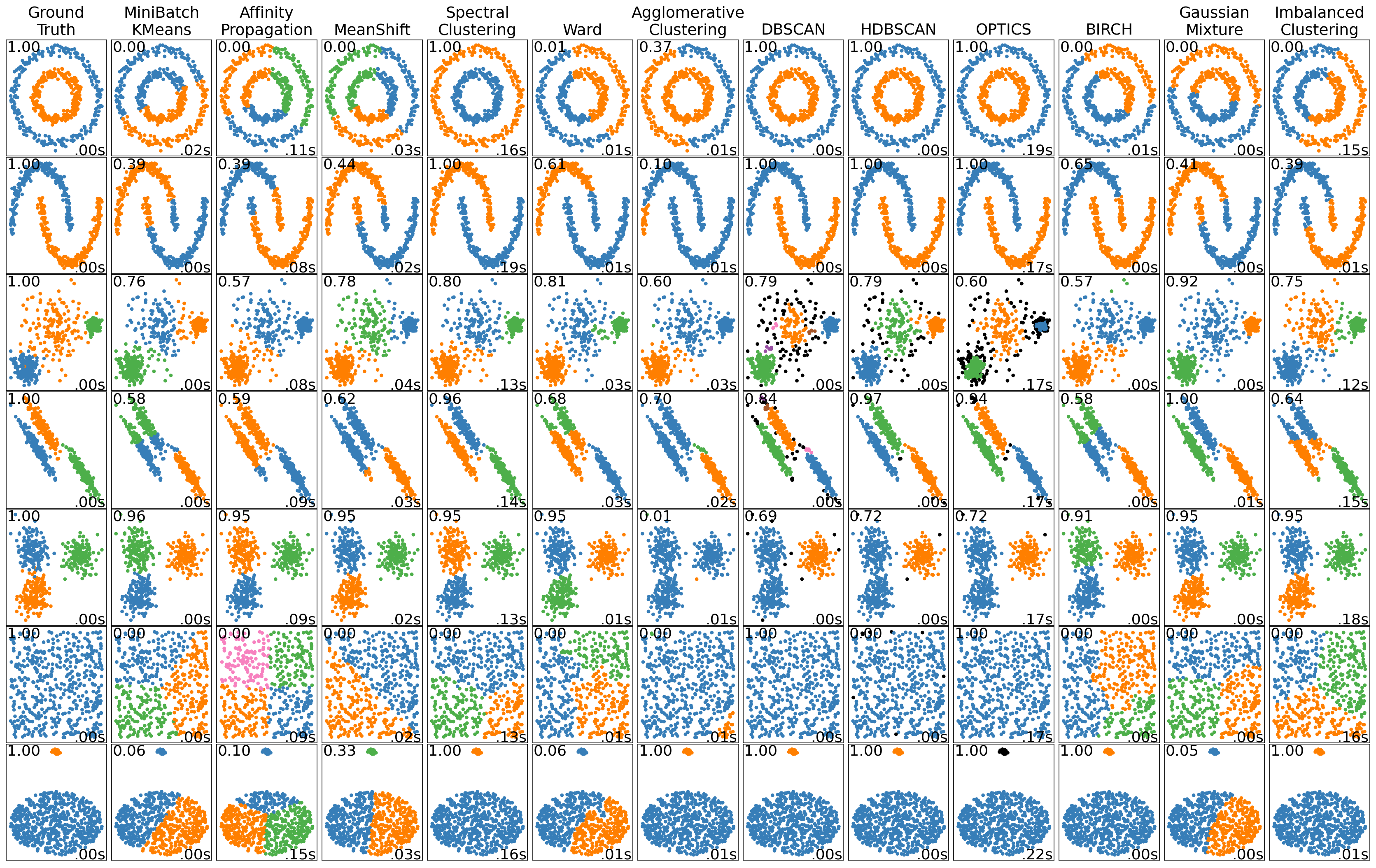}
    \caption{
    Results for the comparison at Section~\ref{sec: Sk-learn comp}.
    The rightmost column is our method and the bottom row is our motivation data from Section~\ref{sec: motivation} for $x:=625$.
    The leftmost column is the ground truth clustering.
    The other rows were copied from the comparison at Scikit-learn~\cite{scikit-learn}, which this comparison is based on.
    }
    \label{fig: motivation 3}
\end{figure}

Inspired by the comparison of various clustering algorithms at Scikit-learn~\cite{scikit-learn}, we compare our method to the ones presented therein.
For a fair comparison, we utilized the open-source code for the comparison (note that the $x$ and $y$ values are presented differently than previously).
Specifically, we added $\approxoncore$, and the previously generated data at Section~\ref{sec: motivation}, for $n:=625$ and with the same pseudo-random number generator as in the open source code.
The results along with running times (bottom right) and V-measure~\cite{V_measure} (top left), which attempts to measure accuracy (larger values are better), are presented in figure~\ref{fig: motivation 3}.
Our method is in the rightmost column, and our additional data is in the bottom row.

Our method is essentially identical to the MiniBatch-KMeans in all the entries beside the bottom row, where it fails similarly to $k$-means in Section~\ref{sec: motivation}.
That is for the top two rows the classification fails, as can be expected since the classes are not convexly separated.
For the other rows as MiniBatch-KMeans, our method was successful, besides the central row (4 from top and bottom).

The only other methods that succeed for the bottom row are hierarchical clustering-based and spectral clustering.
For hierarchical clustering and DBSCAN-related clustering (e.g., Ward to BIRCH), it can be seen that the methods fail, while ours does not, when the clusters are not well separated, such as rows 3,5,6 from the top.
For spectral clustering, the method succeeds in all the tests, but note that the time complexity is very substantial and the method is prohibitively slow for large sets (for example $100,000$ points) while $k$-means computes the output almost instantly for $100,000$ points.
\clearpage
\subsection{``Choice" clustering} \label{sec: choice}
Inspired by the non-decisive results of the compression at Section~\ref{sec: Sk-learn comp}, where the best method (besides spectral clustering, which has large time complexity) for each dataset varies, we propose a novel method (at least to the best of our knowledge), of clustering via various algorithms and choose the best clustering among the results.
In practice, we utilize the silhouette-score~\cite{Silhouettes} to choose the best clustering but other methods can be used.
To emphasize the effectiveness of the method we combine it with hierarchical clustering, as follows.

\textbf{Preliminaries on hierarchical clustering.}
Hierarchical clustering consists of either merging or splitting clusters of the data, recursively, usually splitting into two clusters or merging pairs of clusters. See see~\cite{hyrarcical} an in-depth overview 
Those algorithms are of significant use in practice, as demonstrated by their inclusion in Scikit-learn~\cite{scikit-learn}.
In this work, we mostly focus on the case of splitting clusters, or, as commonly referred to, divisive hierarchical clustering.
The alternative method is merging clusters or as commonly referred to, agglomeration hierarchical clustering.
For the divisive hierarchical clustering, we split each cluster into an equal number of clusters which are then split again up to a certain split depth, which, results in a balanced tree of a predefined depth.
This method is rather similar to common algorithms for Decision trees, such as the CART algorithm~\cite{cart}.
There are many criteria for the split, notably $k$-means, with details in~\cite{Divisive_kmeans}.

\subsection{Examples}\label{sec: choice examples}
In the following example, we apply image quantization as explained in Section~\ref{mot: image}, with the only difference being that instead of the two clustering methods for the RGB values we consider the following three options to compute the divisive clustering:\\
(i). Our clustering, where each split in the divisive clustering is done by utilizing the $\approxoncore$ from Algorithm~\ref{alg: Coreset}.\\
(ii). The Scikit-Learn clustering, where each split in the divisive clustering is done by applying kmeans++~\cite{kmeans++}, as implemented by Scikit-Learn~\cite{scikit-learn}.\\
(iii). Choice clustering, where each split in the divisive clustering is done by computing both the $\approxoncore$ clustering and the kmeans++ clustering, then choosing the clustering with the highest mean silhouette-score as implemented by Scikit-Learn with $\texttt{sample\_size}$ set to $1024$.
In case of an error in the computation of the silhouette-score, caused by its sample, we set $\texttt{sample\_size}$ to $\texttt{None}$ that entitles no sample being done.

We refer to each quantization according to its clustering, that is, choice quantization is the quantization produced by choice clustering, etc.

\begin{figure}[h!]
    \centering
    \includegraphics[width=0.244\textwidth]{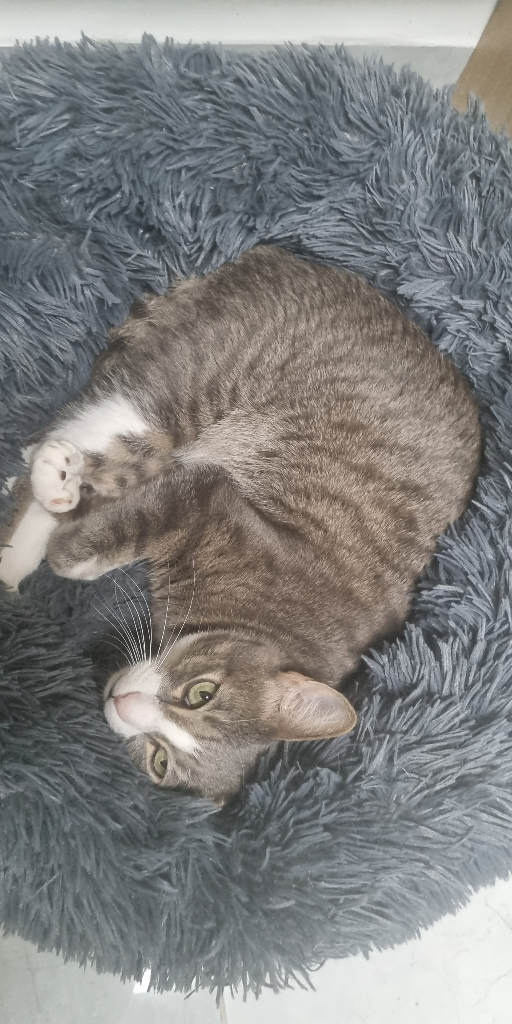}
    \includegraphics[width=0.244\textwidth]{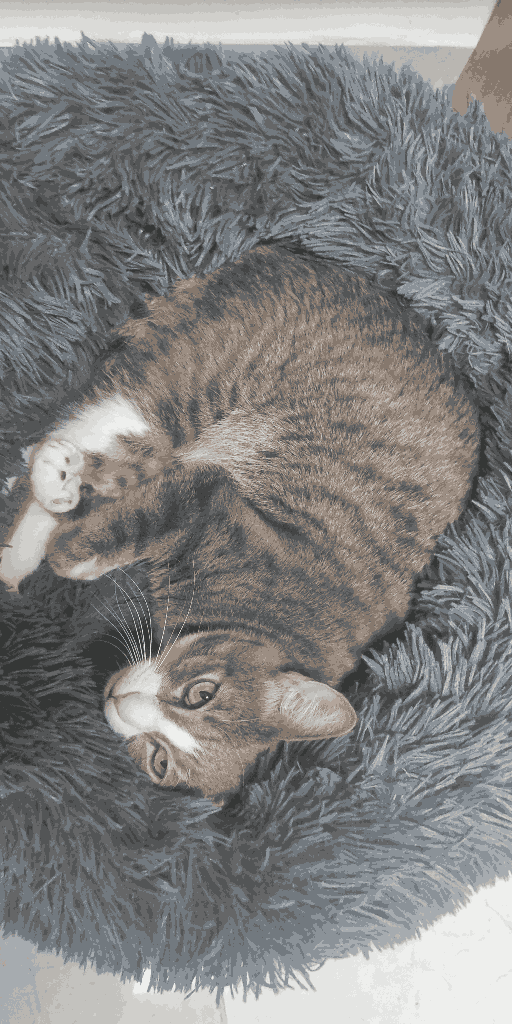}
    \includegraphics[width=0.244\textwidth]{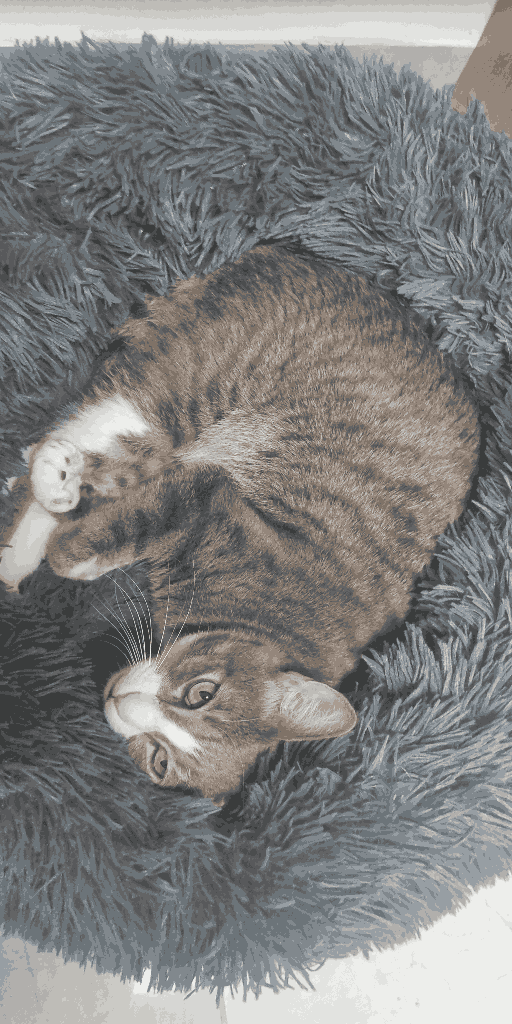}
    \includegraphics[width=0.244\textwidth]{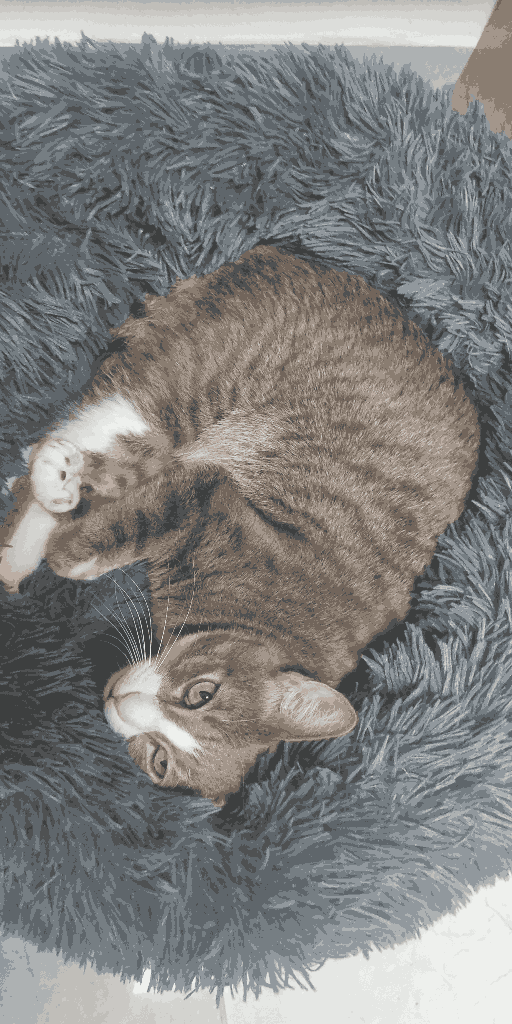}
    
    \includegraphics[width=0.244\textwidth]{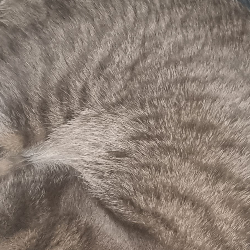}
    \includegraphics[width=0.244\textwidth]{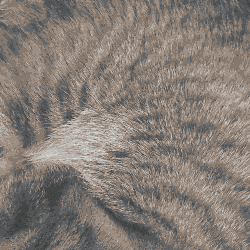}
    \includegraphics[width=0.244\textwidth]{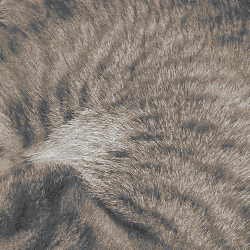}
    \includegraphics[width=0.244\textwidth]{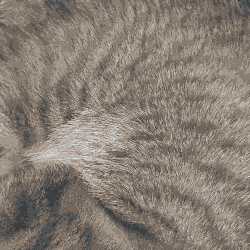}
    \caption{Results for our original cat (Gray) image of Section~\ref{mot: image}.
    The rows correspond to \textbf{(top):} The full images. \textbf{(bottom):} a ``zoom in" on a section of the cat's fur for easier comparison.
    The images (left to right) are: the image we attempted to cluster (ground truth), the result of our quantization, the result of the Scikit-Learn quantization, and the result of the Choice quantization.
    Observe the cat's fur at the ground truth image that has a blue hue in all the quantizations besides the Choice quantization.
    }
    \label{fig: cat cluster}
\end{figure}

In the following example, we have photographed a gray cat (Gray) lying in its blue cat bed, that is the image is novel.
The image was resized to be $1024\times 512$ pixels.
We consider a tree depth of $4$, that is $2^4=16$ levels.
The results of the quantization are provided in Figure~\ref{fig: cat cluster}.

In Figure~\ref{fig: cat cluster}, both our quantization and the Scikit-Learn quantization have resulted in a noticeable miscoloring of the cat's fur to have a bluish hue.
In the Choice quantization, the cat's fur (while still deviating from the ground truth) was noticeably closer to the original color.

\begin{figure}[h!]
    \centering
    \includegraphics[width=0.2425\textwidth]{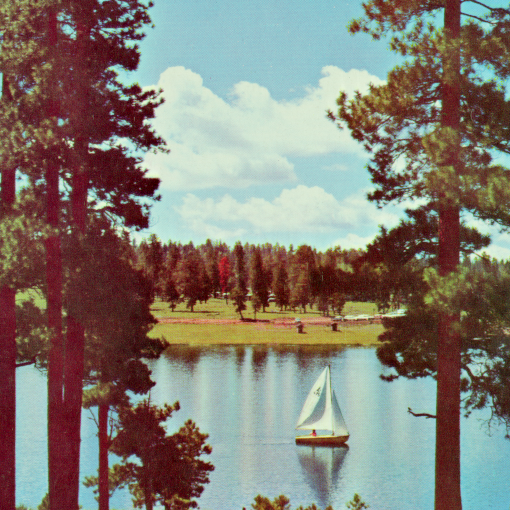}
    \includegraphics[width=0.2425\textwidth]{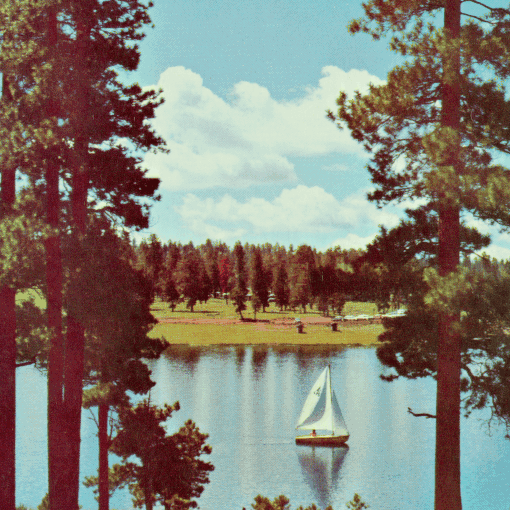}
    \includegraphics[width=0.2425\textwidth]{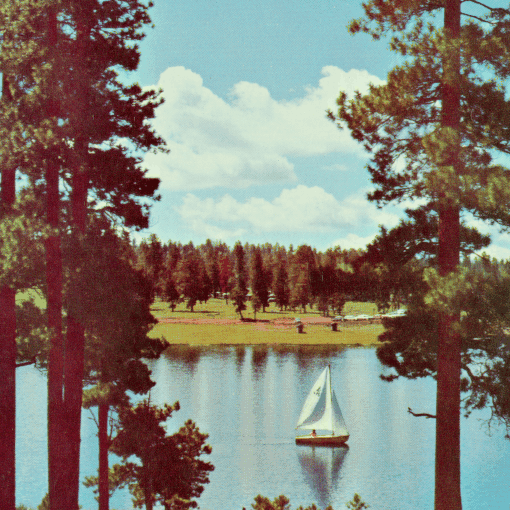}
    \includegraphics[width=0.2425\textwidth]{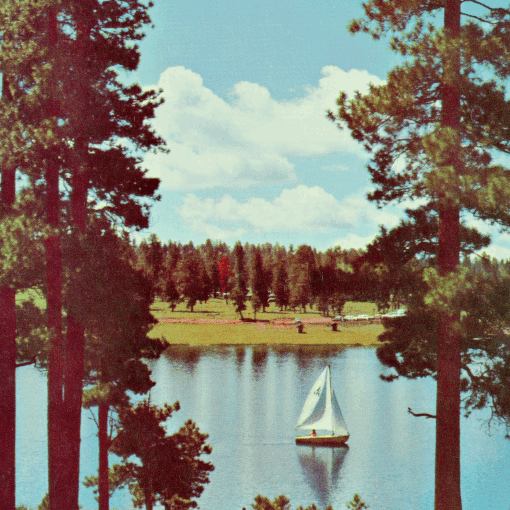}
    \caption{Results for the boat image of Section~\ref{mot: image}.
    The images (left to right) are: the image we attempted to cluster (ground truth), the result of our quantization, the result of the Scikit-Learn quantization, and the result of the Choice quantization.
    Observe the red tree at the background of the ground truth image that is ``missing" from all the quantizations besides the Choice quantization.
    }
    \label{fig: boat cluster}
\end{figure}

The ``USC-SIPI Image Database" contains (but is not limited to) a collection of various reference images commonly used for image processing.
In the following example, we utilized the ``Sailboat on lake" available at~\url{https://sipi.usc.edu/database/preview/misc/4.2.06.png}.
The image is of size $512\times 512$, due to a rather noisy boundary we have removed the pixels across the image border.
That is, the size of the considered image is $510\times 510$.
Then we quantized the image with a tree depth of $7$, that is $2^7=128$ levels.
The results of the quantization are provided in Figure~\ref{fig: boat cluster}.

In Figure~\ref{fig: boat cluster}, both our quantization and the Scikit-Learn quantization have ``missed" the red tree in the background and colored it similarly to the other trees.
In the Choice quantization, the colors of the red tree in the background trees (while still deviating from the ground truth) were noticeably closer to the original colors.
That is, when glancing over the images, the red tree is apparent only for Choice quantization among the quantization methods considered.

Those examples demonstrate the effectiveness of the proposed choice clustering, where the choice allowed better clustering than both the original clustering methods considered.

\section{Conclusion} \label{sec: conclusion}
In this work, we presented a novel method for imbalanced clustering.
This method supports unlabeled data, unlike the numerous previous methods that require labels and show no noticeable improvement over our method; see the additional tests in the appendix.
Our method has competitive running time to~\cite{kmeans++}, while not requiring equal-sized clusters for the classification as known previously and demonstrated in Figures~[\ref{fig: motivation 1},\ref{fig: motivation 2}].
We also proposed choice clustering and demonstrated significant improvement in image quantization over k-means, which is prevalent for this task.

We hope that our work will allow clustering for the numerous cases where despite there being no labels, it is also not the case that the clusters should be equally sized, e.g., as in Figures~[\ref{fig: motivation 1},\ref{fig: motivation 2}].

\section{Future work and limitations} \label{sec: future work}
While the suggested approximation in Definition~\ref{def: approx} performs rather well in practice we believe that there is a significant space for both practical and theoretical improvement.
A consequence of this is that if there is no significant class imbalance, that $k$-means would produce a better center choice, despite that the same clusters will (most likely) be chosen.

A limitation of our coreset construction is that it does not produce a weighted set as an output.
This gives a substantial theoretical limitation to using the suggested coreset in an approximation.
I.e., if the coreset returned a weighted set we could not only in practice but also in theory compute the provable approximation over the coreset.
We hope to address this point in future revisions.

An additional limitation of our method is that it currently supports only Euclidean distances.
We hope that our work will be generalized to absolute distances, robust M-estimators, and other loss functions that satisfy the triangle inequality, up to a constant factor, as in other coreset constructions~\cite{Coreset_survey}.

\section*{Acknowledgements}
David Denisov was (partially) funded by the Israeli Science Foundation (Grant No. 465/22). Shlomi Dolev was (partially) funded by the Israeli Science Foundation (Grant No. 465/22) and by Rita Altura trust chair in Computer Science. Michael Segal was (partially) funded by the Israeli Science Foundation (Grant No. 465/22) and by the Army Research Office under Grant
Number W911NF-22-1-0225. The views and conclusions contained in this document are those of the authors and
should not be interpreted as representing the official policies, either expressed or implied, of the Army Research Office or the U.S. Government. The U.S. Government is authorized to reproduce and distribute reprints for Government purposes notwithstanding any copyright notation herein.

\bibliographystyle{splncs04}
\bibliography{ref}

\begin{thebibliography}{10}
\providecommand{\url}[1]{\texttt{#1}}
\providecommand{\urlprefix}{URL }
\providecommand{\doi}[1]{https://doi.org/#1}

\bibitem{fast_select}
Alexandrescu, A.: Fast deterministic selection. arXiv preprint arXiv:1606.00484
   (2016)

\bibitem{Divisive_Toolbox}
Anagnostou, P., Tasoulis, S., Plagianakos, V.P., Tasoulis, D.: Hipart:
  Hierarchical divisive clustering toolbox. Journal of Open Source Software
  \textbf{8}(84), ~5024 (2023). \doi{10.21105/joss.05024},
  \url{https://doi.org/10.21105/joss.05024}

\bibitem{VC_arithmetic_function}
Anthony, M., Bartlett, P.L.: Neural Network Learning: Theoretical Foundations.
  Cambridge University Press (2009)

\bibitem{kmeans++}
Arthur, D., Vassilvitskii, S.: K-means++: The advantages of careful seeding.
  In: Proceedings of the ACM-SIAM Symposium on Discrete Algorithms (SODA). p.
  1027–1035. SODA '07, Society for Industrial and Applied Mathematics, USA
  (2007)

\bibitem{SMOTETomek}
Batista, G.E., Bazzan, A.L., Monard, M.C., et~al.: Balancing training data for
  automated annotation of keywords: a case study. Wob  \textbf{3},  10--8
  (2003)

\bibitem{openCV}
Bradski, G.: {The OpenCV Library}. Dr. Dobb's Journal of Software Tools  (2000)

\bibitem{SMOTE}
Chawla, N.V., Bowyer, K.W., Hall, L.O., Kegelmeyer, W.P.: Smote: synthetic
  minority over-sampling technique. Journal of artificial intelligence research
   \textbf{16},  321--357 (2002)

\bibitem{webpage}
Ding, Z.: Diversified ensemble classifiers for highly imbalanced data learning
  and its application in bioinformatics. Ph.D. thesis, Georgia State University
  (2011)

\bibitem{Coreset_survey}
Feldman, D.: Core-sets: Updated survey. Sampling Techniques for Supervised or
  Unsupervised Tasks pp. 23--44 (2020)

\bibitem{zahimsc}
Feldman, D., Kfir, Z., Wu, X.: Coresets for {G}aussian mixture models of any
  shape. CoRR  \textbf{abs/1906.04895} (2019),
  \url{http://arxiv.org/abs/1906.04895}

\bibitem{newframework}
Feldman, D., Langberg, M.: A unified framework for approximating and clustering
  data. Proceedings of the Annual ACM Symposium on Theory of Computing  (06
  2011). \doi{10.1145/1993636.1993712}

\bibitem{bigtotiny}
Feldman, D., Schmidt, M., Sohler, C.: Turning big data into tiny data:
  Constant-size coresets for k-means, pca and projective clustering. In:
  Proceedings of the twenty-fourth annual ACM-SIAM symposium on Discrete
  algorithms. pp. 1434--1453. Society for Industrial and Applied Mathematics
  (2013)

\bibitem{outliers-resistance}
Feldman, D., Schulman, L.J.: Data reduction for weighted and outlier-resistant
  clustering. In: Proceedings of the twenty-third annual ACM-SIAM symposium on
  Discrete Algorithms. pp. 1343--1354. Society for Industrial and Applied
  Mathematics (2012)

\bibitem{harpeledsharir}
Har{-}Peled, S., Sharir, M.: Relative (p,epsilon)-approximations in geometry.
  CoRR  \textbf{abs/0909.0717} (2009), \url{http://arxiv.org/abs/0909.0717}

\bibitem{SciPy}
Jones, E., Oliphant, T., Peterson, P., et~al.: {SciPy}: Open source scientific
  tools for {Python} (2001--), \url{http://www.scipy.org/}

\bibitem{imbalanced_clustering_new}
Krawczyk, B.: Learning from imbalanced data: open challenges and future
  directions. Progress in Artificial Intelligence  \textbf{5}(4),  221--232
  (2016)

\bibitem{Divisive_kmeans}
Lamrous, S., Taileb, M.: Divisive hierarchical k-means. In: 2006 International
  Conference on Computational Inteligence for Modelling Control and Automation
  and International Conference on Intelligent Agents Web Technologies and
  International Commerce (CIMCA'06). pp. 18--18. IEEE (2006)

\bibitem{kmeans_smote}
Last, F., Douzas, G., Bacao, F.: Oversampling for imbalanced learning based on
  k-means and smote  (2017)

\bibitem{Imbalanced_learn}
Lema{{\^i}}tre, G., Nogueira, F., Aridas, C.K.: Imbalanced-learn: A python
  toolbox to tackle the curse of imbalanced datasets in machine learning.
  Journal of Machine Learning Research  \textbf{18}(17), ~1--5 (2017),
  \url{http://jmlr.org/papers/v18/16-365.html}

\bibitem{cart}
Lewis, R.J.: An introduction to classification and regression tree (cart)
  analysis. In: Annual meeting of the society for academic emergency medicine
  in San Francisco, California. vol.~14. Citeseer (2000)

\bibitem{LLS01}
Li, Y., Long, P.M., Srinivasan, A.: Improved bounds on the sample complexity of
  learning. Journal of Computer and System Sciences  \textbf{62}(3),  516--527
  (2001)

\bibitem{lucic2017training}
Lucic, M., Faulkner, M., Krause, A., Feldman, D.: Training gaussian mixture
  models at scale via coresets. The Journal of Machine Learning Research
  \textbf{18}(1),  5885--5909 (2017)

\bibitem{kNN_imbalanced}
Mani, I., Zhang, I.: knn approach to unbalanced data distributions: a case
  study involving information extraction. In: Proceedings of workshop on
  learning from imbalanced datasets. vol.~126, pp.~1--7. ICML (2003)

\bibitem{hyrarcical}
Murtagh, F., Contreras, P.: Algorithms for hierarchical clustering: an
  overview. Wiley Interdisciplinary Reviews: Data Mining and Knowledge
  Discovery  \textbf{2}(1),  86--97 (2012)

\bibitem{scikit-learn}
Pedregosa, F., Varoquaux, G., Gramfort, A., Michel, V., Thirion, B., Grisel,
  O., Blondel, M., Prettenhofer, P., Weiss, R., Dubourg, V., Vanderplas, J.,
  Passos, A., Cournapeau, D., Brucher, M., Perrot, M., Duchesnay, E.:
  Scikit-learn: Machine learning in {P}ython. Journal of Machine Learning
  Research  \textbf{12},  2825--2830 (2011)

\bibitem{raschka2020machine}
Raschka, S., Patterson, J., Nolet, C.: Machine learning in python: Main
  developments and technology trends in data science, machine learning, and
  artificial intelligence. arXiv preprint arXiv:2002.04803  (2020)

\bibitem{V_measure}
Rosenberg, A., Hirschberg, J.: V-measure: A conditional entropy-based external
  cluster evaluation measure. In: Proceedings of the 2007 joint conference on
  empirical methods in natural language processing and computational natural
  language learning (EMNLP-CoNLL). pp. 410--420 (2007)

\bibitem{Silhouettes}
Rousseeuw, P.J.: Silhouettes: A graphical aid to the interpretation and
  validation of cluster analysis. Journal of Computational and Applied
  Mathematics  \textbf{20},  53--65 (1987).
  \doi{https://doi.org/10.1016/0377-0427(87)90125-7}

\bibitem{sampleing}
Santoso, B., Wijayanto, H., Notodiputro, K., Sartono, B.: Synthetic over
  sampling methods for handling class imbalanced problems : A review. IOP
  Conference Series: Earth and Environmental Science  \textbf{58},  012031 (03
  2017). \doi{10.1088/1755-1315/58/1/012031}

\end{thebibliography}
\appendix
\section{Algorithms} \label{sec: alg}
For describing our algorithms, we use the following definitions.
First, we extend the $\alpha$-approximation in Definition~\ref{def: approx} to a more rough approximation, called $(\alpha,\beta)$ or Bi-Criteria approximation~\cite{newframework}.
\begin{definition} [$(\alpha,\beta)$-approximation] \label{def:bicretiria}
Let $\alpha \geq 0$, and $\beta\geq 1$ be an integer.
Let $P \subset \REAL^d$ be a finite set.
An $(\alpha,\beta)$-\emph{approximation} $B := \br{c_1,\cdots,c_\beta}\subset \REAL^d$ of $P$ is a set of size $k\beta$ such that 
\begin{equation}
\ell(P,B) \leq \alpha \cdot \min_{C\subset \REAL^d, |C|=k} \ell(P,C).
\end{equation}
\end{definition}

We define the following call to the approximation from~\cite{kmeans++}.
\begin{claim} [Theorem 3.1~\cite{kmeans++}] \label{claim: kmeans++}
There is an algorithm that gets a finite set $P\subseteq\REAL^d$ of size $n\geq 1$, and returns a set $C\subseteq\REAL^d$ of $|C|=k$ points (centers) such that, with probability at least $1/2$,
\begin{equation}
\sum_{p\in P} \min_{c\in C} \|p-c\|_2 \in O(\log k)\cdot \min_{C'\subset \REAL^d,|C'|=k} \sum_{p\in P} \min_{c\in C'} \|p-c\|_2.
\end{equation}
Moreover, $C$ can be computed in $O(dkn)$ time.

We denote this algorithm as $\kmeanspp$, and it receives a finite set $P\subseteq\REAL^d$ and the integer $k$.
\end{claim}

\subsection{$(\alpha,\beta)$-approximation}\label{sec: bi}
The input for the following algorithm is a set $P\subset \REAL^d$ of size $n$, along with an integer $k\geq1$, and an upper limit $\delta\in(0,1/10]$ on the allowed failure probability.
Its output is a set $B\subset \REAL^d$, which is a $O\big(k \log(n),k \log(n)\big)$-approximation of $P$ with probability at least $1-\delta$.

While the algorithm might initially look similar to the $(\alpha,\beta)$-approximation from~\cite{newframework}, it does not entail solving the original problem on the sampled subset (we use a single and not $k$ points as the sample).
This modification enables a non-exponential dependency on $k$, unlike similar algorithms for $k$-means from~\cite{newframework}.
It should be emphasized that there are approximation algorithms for $k$-means that are non-exponential dependency on $k$, notably~\cite{kmeans++}, which is used in this work; see Algorithm~\ref{alg: Coreset}.

The following theorem states the main properties of our algorithm. For its proof of correctness see Theorem~\ref{th: proof}.
\begin{theorem}\label{th}
Let $P\subset \REAL^d$ be a set of size $n\geq 2$.
Put $\delta\in (0,1/10]$, and let $B$ be the output of a call to $\Bicriteria(P,k,\delta)$; see Algorithm~\ref{alg: Bicriteria}.
Then, with probability at least $1-\delta$, we have that $B$ is a $O\big( k \log n,k\log n\big)$-approximation to $P$.
Moreover, the computation time of $B$ is in $O\big(ndk + d k^5 \log^3 (n/\delta)\big)$.
\end{theorem}

\begin{algorithm}[h]
    \caption{$\Bicriteria(P,k,\delta)$; see Theorem~\ref{th}}
    \label{alg: Bicriteria}
    \SetKwInOut{Input}{Input}
    \SetKwInOut{Output}{Output}
    \Input{A finite set $P\subset \REAL^d$ of size $n\geq 2$, an integer $k\geq 1$, and $\delta\in (0,1/10]$.}
    \Output{A set $B\subset \REAL^d$, which, with probability at least $1-\delta$, is an $O\big( k\log n ,k\log n\big)$-approximation to $P$.}

    $P':=P$

    $B:=\emptyset$

    $c:=$ a universal constant (in $O(1)$) that can be derived from the poof of Theorem~\ref{th}.
    
    $\lambda:= \lceil c k^2 \log(n/\delta) \rceil$.
    \label{line: lambda: bi}
    
    \While{$|P'|\geq 2 \lambda$}
    {

        \label{line: while start}
        Pick a sample $S\subseteq P$ of size $\lambda$ from $P'$, where each element in $S$ is sampled i.i.d. and uniformly at random from $P'$.

        \For{every $p\in S$}
        {
        
        Set $S_p$ as the subset of $S$ of the $\displaystyle  \left\lfloor \frac{15|S|}{16k} \right\rfloor$ closest (euclidean distance) points to $p$, ties broken arbitrarily.

        $\ell_p:= \ell(S_p,\br{p})$
        }

        Let $\displaystyle p\in \argmin_{p\in S} \ell_p$.
        \label{line: get robust}

        Set $P'_p$ as the subset of $P'$ of the $\displaystyle  \left\lfloor \frac{3|P'|}{4k} \right\rfloor$ closest (euclidean distance) points to $p$, ties broken arbitrarily.
        \label{line: chose while}

        $B= B\cup \br{p}$

        $P'= P'\setminus P'_p$

        \label{line: update while}
    }

    $B= B \cup P'$ \label{line: end while}
    
    \Return $B$
    
\end{algorithm}

\subsection{Coreset}\label{sec: core}
The following algorithm utilizes the sensitivity framework from~\cite{newframework} to compute an $\eps$-coreset.

The following theorem states the main properties of the Algorithm, for its proof see Theorem~\ref{th: coreset proof}.
\begin{theorem} \label{th: coreset}
Let $P\subset \REAL^d$ be a finite set of size $n\geq 2$.
Put $\delta\in (0,1/10]$, and $\eps\in (0,1)$.
Let $(C,w)$ be the output of a call to $\Coreset(P,k,\delta,\eps)$; see Algorithm~\ref{alg: Coreset}.
Then,
\begin{equation}
|C|\in O\of{\frac{kd^3\log(k)\log^4(n)}{\eps^2} \left(\log\big(\log(k)\log(n)\big) + \log\left(\frac{1}{\delta}\right)\right)}
\subseteq O\of{\frac{kd^3\log^2k\log^5n+\log(1/\delta)}{\eps^2}},
\end{equation}
and, with probability at least $1-\delta$, we have that $(C,w)$ is a $\eps$-coreset for $P$; see Definition~\ref{Def:coreset}.
Moreover, $(C,w)$ can be computed in $O\big(ndk \log(1/\delta) \big)$ time.
\end{theorem}


\begin{algorithm}[h!]
    \caption{$\Coreset(P,k,\delta,\eps)$; see Theorem~\ref{th: coreset}}
    \label{alg: Coreset}
    \SetKwInOut{Input}{Input}
    \SetKwInOut{Output}{Output}
    \Input{A set $P\subset \REAL^d$ of size $n\geq 2$, integer $k\geq 1$,$\delta\in (0,1/10]$, and $\eps\in(0,1)$.}
    \Output{A pair $(C,w)$, where $C\subset \REAL^d$ and $w:C\to \REAL$, which, with probability at least $1-\delta$ is an $\eps$-coreset for $P$.}

    Set $\displaystyle  \lambda \in O\of{\frac{\log(k)\log^4(n)}{\eps^2} \left(k d^3 \log\big(\log(k)\log(n)\big) + \log\left(\frac{1}{\delta}\right)\right)},$ \newline \hspace*{5em} where the exact value can be derived from the proof of Lemma~\ref{th: coreset}.
    \label{line: lambda}

    \If{$n\leq \lambda$}
    {
    Set $w:P\to \br{1}$, i.e., the function that maps every $p\in P$ to $w(p):=1$.

    \Return $(P,w)$.
    }

    $\mathcal{C}:=\emptyset$

    \For{every $i\in \Big[\big\lceil \log(2/\delta)\big\rceil\Big]$ \label{line: core for start}}
    {
        $C_i:=\kmeanspp(P,k)$ \tcp{see Claim~\ref{claim: kmeans++}, $C_i$ is a set of $k$ points in $\REAL^d$.}

        $\mathcal{C} := \mathcal{C}\cup \br{C_i}$ \tcp{note that $\mathcal{C}$ is a set of sets.}
    }
    
    Let $\displaystyle B\in \argmin_{C\in \mathcal{C}} \sum_{p\in P} \min_{c\in C} 
    \| c-p\|_2$, ties broken arbitrarily \tcp{$B$ is a set of $k$ points in $\REAL^d$.}
    \label{line: core for end}
    
    Let $\mathcal{P}:=\br{P_c}_{c\in B}$ be a partition of $P$, such that every $p\in P$ is in $P_c$ if $\displaystyle c \in \argmin_{c'\in B} \| c'-p\|_2$. Ties broken arbitrarily; i.e., for every $c\in B$ we set $P_c$ are the points in $P$ that are closest to $c$.

    \For{every $c\in B$}
    {

    Set $w(c):= |P_c|$.

    For every $p\in P_c$, set $\displaystyle \ell_p:= \frac{\| p-c\|_2}{\log^2(|P_c|+1)} $

    }

    \If{$\ell(P,B)=0$}
    {
     \Return $(B,w)$
    }
    Set $s(p) := \displaystyle \frac{\ell_p}{\sum_{p\in P} \ell_p}$ for every $p\in P$.

    Pick a sample $S$ of $\lambda$ i.i.d. points from $P$, where each $p\in P$ is sampled with probability $s(p)$.
    \label{line: sample}

    $w(p) := 1/\big(\lambda \cdot s(p)\big)$ for every $p\in S$.

    $\displaystyle w(c) := w(c) - \frac{1}{\lambda \cdot s(p)}$ for every $c\in B$ and $p\in P_c$.
    
    $C:= B \cup S$
    
    \Return $(C,w)$
    
\end{algorithm}
\subsection{The gap between theory and practice}\label{sec: gap}
In our code, we apply minor changes to our theoretically provable algorithms, which have seemingly negligible effects on their output quality but aim to improve their running times. Those changes are:\\
\textbf{(i).} During Line~\ref{line: lambda: bi} of Algorithm~\ref{alg: Bicriteria} we compute $\lambda$ given the allowed probability of failure. Instead, we set $\lambda:=64$ and fixing $\delta$.\\
\textbf{(ii).} Line~\ref{line: end while} of Algorithm~\ref{alg: Bicriteria} sets $B:=B\cup P'$.
Instead, to reduce the size of $B$ (which is the output of Algorithm~\ref{alg: Bicriteria}), we set $C:=\Approx((P',w),k)$, where $w:P'\to \br{1}$ (each point is mapped to $1$), and set $B:=B\cup C$; see Definition~\ref{alg: approx}.\\
\textbf{(iii).} During Line~\ref{line: lambda} of Algorithm~\ref{alg: Coreset} we compute $\lambda$ given  the allowed probability of failure. Instead, we set $\lambda:=128$ and fixing $\delta$.\\
\textbf{(iv).} In Lines[\ref{line: core for start}--\ref{line: core for end}] of Algorithm~\ref{alg: Coreset} we compute $B\subset \REAL^d,|B|=k$ by rerunning $\kmeanspp(P,k)$ $O(\log (1/\delta)$ times.
Again, we fix $\delta$ so that the number iteration would be one, which amounts to setting $B:=\kmeanspp(P,k)$.

\section{Constant factor approximation}\label{sec: analysis of approx}
In this section, we prove that we can compute an $\alpha$-approximation, as stated in Definition~\ref{alg: approx}, for a set $P\subset \REAL^d$ of size $n\geq 2$, and $\alpha:=2^k\log^2 (n+1)$.

This is formally stated as follows, where the proof is inspired by~\cite{outliers-resistance}.
\begin{lemma}\label{l: Approx proof}
Let $(P,w)$ be a weighted set of size $n\geq k$ where $w:P\to [0,\infty)$. That is, $P\subset\REAL^d$ and $|P|=n$.
Suppose that $\min_{C\subset\REAL^d,|C|=k} \ellt\big((P,w),C\big)$ exists.
Let $Q:=\Approx\big((P,w),k\big)$; see Definition~\ref{alg: approx}.
Then $Q$ is a $2 \log^2(1+n)$-approximation for $(P,w)$; see Definition~\ref{def: approx}.
Moreover, the computation time of $Q$ is in $O\of{n^{k+1}dk}$.
\end{lemma}
\begin{proof}
We will prove that there is a set $C'\subset P$ of size $k$, which is a $2^k \log(1+n)^{2}$-approximation for $(P,w)$.
Since we iterate in $\Approx\big((P,w),k\big)$ over the sets $C\subset P$ of size $k$, and take one of these subsets with the smallest loss, this proves that $C$ is a $2^k \log^2(1+n)$-approximation for $(P,w)$.

Let $C^* \in \argmin_{C\subset\REAL^d,|C|=k} \ellt\big((P,w),C\big)$, which was assumed to exist.
If $C^*\subset P$, then it immediately proves the existence of the stated $C'$.
Hence, we assume this is not the case.
That is, $C^*\not \subset P$.
Let $\mathcal{P}:=\br{P_c}_{c\in C^*}$ be a partition of $P$, such that every $p\in P$ is in $P_c$ if $\displaystyle c \in \argmin_{c'\in C^*} \| c'-p\|_2$. Ties broken arbitrarily; i.e., for every $c\in C^*$ we set $P_c$ as the points in $P$ that are closest to $c$.
For every $c\in C^*$ let $c_P$ be the closest (Euclidean distance) point in $P$ to $c$. Ties broken arbitrarily.

Let $c\in C^*$.
By the triangle inequality, for every $p\in P$, we have
\begin{equation}
\| c_P-p\|_2 \leq \| c_P-c\|_2 + \| c-p\|_2 \leq 2 \| c-p\|_2 .
\end{equation}
Summing this over every $p\in P_{c}$ yields
\begin{equation}
\sum_{p\in P_{c}} \| p-p_C\|_2 \leq 2 \sum_{p\in P_{c}} \| c-p\|_2.
\end{equation}
Let $C'':=\br{c_P \mid c\in C^*}\subset P$.
We have
\begin{equation}
\sum_{p\in P} w(p) \min_{c\in C''} \| p-c\|_2 \leq 2 \sum_{p\in P} w(p) \min_{c\in C^*} \| p-c\|_2.
\end{equation}
Hence, by the choice of $\ellt$ in Definition~\ref{def: loss function 2}, it follows that
\begin{equation}
\ellt\big((P,w),C''\big) \leq 2 \log(1+n)^2 \ellt\big((P,w),C^*\big),
\end{equation}
which proves the existence of the sated $C'$.

\textbf{Running time.}
Let $C\subset \REAL^d$ be a set of size $k$.
Observe that for each $p\in P$ we can compute $c\in \argmin_{C} \|p-c\|_2$, ties broken arbitrarily, in $O(dk)$ time.
Let $f:P\to C$ that maps every $p$ to its computed $c\in \argmin_{C} \|p-c\|_2$.
Observe that in the computation of $\ellt\big((P,w), C\big)$, the mapping of $f$ yields the desired partition of $P$.
Hence, we can compute $\ellt\big((P,w),C\big)$ in $O(ndk)$ time.

Observe that the number of sets of size $k$ from $P$ is in $O\of{n^k}$.
Hence, we can compute the output of a call to $\Approx\big((P,w),k\big)$ in $O\of{n^{k+1}dk}$ time.
\end{proof}

\section{Efficient $(\alpha,\beta)$-approximation} \label{sec: bi proof}
For our analysis of Algorithm~\ref{alg: Bicriteria} we require the following preliminaries on \emph{robust medians}.
\subsection{Robust median} \label{sec: bi proof robust}
In the following section, we will define robust median and prove that we can compute it efficiently.
Our method would be based on two parts, the initial method and its speed-up.
The initial method entails taking the best center from a uniform sample of the points and is inspired by~\cite{outliers-resistance}.
The speed-up entails computing the previous robust median on a uniform sample and is based on Section \textit{From $\eps$-approximation to robust medians} of~\cite{newframework}.

\textbf{Initial method}.
For the statement, we utilize the following definitions.
\begin{definition}[\cite{outliers-resistance}]
Let $P \subset \REAL^d$ be a non-empty finite set of points. 
For $q \in \REAL^d$ and $\gamma\in[0,1]$, we denote by $\closest(P, q, \gamma)$ the set that consists of
the $\lceil \gamma |P| \rceil$ points $p\in P$ with the smallest values of $\| q-p\|_2 $, ties broken arbitrarily.
\end{definition}

\begin{definition}[\cite{outliers-resistance}]
For $\gamma\in [0, 1]$, $\eps\in (0,1)$, and $\alpha>0$, the point $q\in \REAL^d$ is a $(\gamma,\eps,\alpha)$-median of the non-empty finite set $Q \subset \REAL^d$ if
\begin{equation}
\sum_{p\in \closest(Q,q,(1-\eps)\gamma)} \| q-p\|_2 
\leq \alpha \cdot \inf_{q'\in \REAL^d} \sum_{p\in  \closest(Q,q',\gamma)} \| q'-p\|_2.
\end{equation}
\end{definition}

Utilizing those definitions we can state the main result of our initial method to compute the robust median.

\begin{lemma}\label{lemma 1}
Let $P\subset \REAL^d$ be a finite and non-empty set.
There is $q\in P$, which is an $(15/16k,1/16,2)$-median of $P$.
\end{lemma}
\begin{proof}
Let $q^*\in \REAL^d$ be a $(1/k,0,1)$-median of $P$.
Let $q$ be the closest point in $P$ to $q^*$. Ties are broken arbitrarily.
By the triangle inequality, for every $p\in P$, we have
\begin{equation}
\| q-p\|_2 \leq \| q-q^*\|_2 + \| q^*-p\|_2 \leq 2 \| q^*-p\|_2 .
\end{equation}
Let $C^*:=\closest(P,q^*,15/16k)$.
Summing this over every $p\in \closest(P,q,15/16k)$ yields
\begin{equation}
\sum_{p\in \closest(P,q,15/16k)} \| q-p\|_2 \leq \sum_{p\in C^*} \| q-p\|_2 \leq 2 \sum_{p\in C^*} \| q^*-p\|_2.
\end{equation}
Hence, $q$ is a $(15/16k,0,2)$-median of $P$, which is also a $(15/16k,1/16,2)$-median of $P$.
\end{proof}

Hence, as suggested by Lemma~\ref{lemma 1}, our initial method to compute a robust median for a set is via an exhaustive search over all the points in the set.

\textbf{Speed up.}
A consequence of Lemma 5.1 from~\cite{newframework} is the following result for our case.
\begin{lemma}\label{lemma 2}
Let $P\subset \REAL^d$ be a finite and non-empty set of points.
Let $\delta\in (0,1/10]$.
Let $S$ be a uniform random sample of i.i.d. points from $P$, of size $\lambda:= c k^2 \log(1/\delta)$, for universal constant $c$ that can be determined from the proof.
With probability at least $1-\delta$, any $(15/16k,1/16,2)$-median of $S$ is a $(1/k,1/4,2)$-median of $P$.
\end{lemma}

\subsection{Analysis of Algorithm~\ref{alg: Bicriteria}} \label{sec: bi proof 2}
In this section, we prove the Theorem~\ref{th}.
\begin{theorem}\label{th: proof}
Let $P\subset \REAL^d$ be a set of size $n\geq 2$.
Put $\delta\in (0,1/10]$, and let $B$ be the output of a call to $\Bicriteria(P,k,\delta)$; see Algorithm~\ref{alg: Bicriteria}.
Then, with probability at least $1-\delta$, we have that $B$ is a $O\big( k \log n,k\log n\big)$-approximation to $P$.
Moreover, the computation time of $B$ is in $O\big(ndk + d k^5 \log^3 (n/\delta)\big)$.
\end{theorem}
\begin{proof}
Let $\lambda:=b k^2 \log(n/\delta)$ where $b$ is as defined in Lemma Lemma~\ref{lemma 2}.
Consider a single iteration of the ``while" loop in the call to $\Bicriteria(P,k,\delta)$.

Substituting $\delta:=\delta/n$ in Lemma~\ref{lemma 2} yields that, with probability at least $1-\delta/n$, any $(15/16k,1/16,2)$-median of $S$ is a $(1/k,1/4,2)$-median of $P'$.
Assume that this event indeed occurs.

By Lemma~\ref{lemma 1}, there is $p\in S$ which is a $(15/16k,1/16,2)$-median of $S$.
Hence, by the definition of $p\in S$ in the current ``while" iteration it follows that $p$ is a $(15/16k,1/16,2)$-median of $S$. 
Thus, we have that $p$ is a $(1/k,1/4,2)$-median of $P'$.

Let $P'_p$ be computed in this ``while" iteration.
We have that
\begin{equation}
\sum_{p'\in P'_p} \| p'-p\|_2 \leq 2 \min_{q^*\in \REAL^d} \sum_{p'\in \closest(P,q^*,1/k)} 
\| q^*-p\|_2.
\end{equation}
Observe that due to the pigeonhole principle, for any possible $k$ centers, there is a center with at least $1/k$ of the points assigned to it.
This yields
\begin{equation}
\frac{k}{|P|+1}\cdot \min_{q^*\in \REAL^d} \sum_{p'\in \closest(P,q^*,1/k)} \leq \min_{C\subset \REAL^d, |C|=k} \ell(P,C).
\end{equation}
Hence, we have that 
\begin{equation}
\ell(P'_p,\br{p}) \leq 2 \min_{C\subset \REAL^d, |C|=k} \ell(P,C).
\end{equation}

Observe that at each iteration of the ``while" loop in Algorithm~\ref{alg: Bicriteria}, the size of $P'$ decreases by a multiplicative factor of $1-1/(2k)$.
Hence, every $2k$ iterations, the size of $P'$ decreases by a multiplicative factor of $\left(1-1/(2k)\right)^{2k}$.
Let $f:\REAL \to \REAL$ be the function that maps every $x\in \REAL$ to $f(x):=\left(1-1/x\right)^{x}$.
A known property of the function is that for every $x\geq 1$, we have that $f(x)\leq 1/e$.
Hence, due to $k$ being a positive integer we have that $\left(1-1/(2k)\right)^{2k}\leq 1/e$.
As such, for every $2k$ iteration of the ``while" loop, the size of $P'$ decreases by a constant multiplicative factor.
Hence, we have that there are $O\big(k \log n\big)$ iterations in the ``while" loop.

Therefore, since there are $O(k \log n)$ iterations of the ``while" loop, for $B$ computed at the end of the ``while" loop (Line~\ref{line: end while} of Algorithm~\ref{alg: Bicriteria}) we have 
\begin{equation}
\ell(P\setminus P',B) \in O(k \log n) \min_{C\subset \REAL^d, |C|=k} \ell(P,C).
\end{equation}
Thus since the algorithm returns $B:=B\cup P'$ it immediately follows that $B$ is a $O\big( k \log n,k \log n\big)$-approximation to $P$.

We assumed, that at each iteration of the ``while" loop, there is $q\in S$ which is a $(15/16k,1/16,2)$-median of $S$ and any such median is also a $(1/k,1/4,2)$-median of $P'$.
This even occurs with a probability of at least $1-\delta/n$.
Hence, by the union bound and that $\delta\leq 1/10$, we have that all the even occurs with probability at least $\displaystyle \left(1-\frac{\delta}{n}\right)^n \leq 1-\delta$.

Thus, with probability at least $1-\delta$, we have that $B$ is a $O\big( k \log n ,k \log n\big)$-approximation to $P$.

\textbf{Running time.}
Consider a single iteration of the ``while" loop in the call to $\Bicriteria(P,k,\delta)$.

It immediately follows that every ``for" iteration can be done in $O(\lambda d)$; note that the choice of $S_p$ does not require sorting of the $S$, for implementation example see~\cite{fast_select}.
Hence, since there are $\lambda$ iterations, it immediately follows that Lines~\ref{line: while start}--Lines~\ref{line: get robust} can be computed in $O\big(\lambda^2 d\big)$ time.
Hence, since $\lambda\in O\big(k^2 \log(n/\delta)\big)$, Lines~\ref{line: while start}--Lines~\ref{line: get robust} can be computed in $O\big(d k^4 \log^2 (n/\delta)\big)$ time.

Let $n'$ be the size of $P'$ in the ``while" iteration considered.
It can be seen that all the computations in Lines~\ref{line: chose while}--Line~\ref{line: update while} can be done in $O\big(n' d \big)$ time; note that the choice of $P'_p$ does not require sorting of the $P$, for implementation example see~\cite{fast_select}.

Hence, each ``while" iteration in the call to $\Bicriteria(P,k,\delta)$, where $n':=|P'|$, can be computed in $O\big(n' d + d k^4 \log^2 (n/\delta)\big)$ time.

Hence, since there are $O\big(k \log n\big)$ iterations in the ``while" loop, and in each iteration the size of $P'$ decreases by a factor of $\displaystyle \left(1-\frac{1}{2k}\right)$, i.e., we have a geometric sum, the total time of the ``while" loop is in
\begin{equation}
    \frac{O\big(n d\big)}{1-\displaystyle \left(1-\frac{1}{2k}\right)} + 
    O\big(k \log n \big) \cdot d k^4 \log^2 (n/\delta).
\end{equation}
Hence, the total time of the ``while" loop is in $O\big(n d k + d k^5 \log ^3(n/\delta) \big)$.
Thus, by the construction of Algorithm~\ref{alg: Bicriteria}, it follows that the computation time is dominated by the computation time of the ``while" loop.
That is, the call to $\Bicriteria(P,k,\delta)$ can be computed in $O\big(n d k + d k^5 \log^3 (n/\delta) \big)$ time.
\end{proof}

\section{Coreset for the relaxed problem}\label{sec: analysis of Core}
The coreset construction that we use in Algorithm~\ref{alg: Coreset} is a non-uniform sample from a distribution, which is known as sensitivity, that is based on the~$(\alpha,\beta)$-approximation defined in Definition~\ref{def: approx}.
To apply the generic coreset construction we need two ingredients:
\begin{enumerate}
    \item A bound on the dimension induced by the query space (``complexity") that corresponds to our problem as formally stated and bounded in subsection~\ref{subsection - VC dim}. This bound determines the required size of the random sample picked in Algorithm~\ref{alg: Coreset}.
    \item A bound on the sensitivity as formally stated and bounded in the proof of Lemma~\ref{th: coreset}. This bound on the sensitivity determines the required size of the random sample that is picked in Algorithm~\ref{alg: Coreset}.
\end{enumerate}

\subsection{Bound on the VC-Dimension}
\label{subsection - VC dim}

We first define the classic notion of VC-dimension, which is used in Theorem 8.14 in \cite{VC_arithmetic_function}, and is usually related to the PAC-learning theory~\cite{LLS01}.
The following definition is from \cite{lucic2017training}.
\begin{definition}
Let $F \subset \br{\REAL^d \to \br{0,1}}$ and let $X\subset \REAL^d$.
Fix a set $S=\br{x_1,\cdots,x_n}\subset X$ and a function $f\in F$.
We call $S_f=\br{x_i \in S \mid f(x_i)=1}$ the induced subset of $S$ by $f$.
A subset $S=\br{x_1,\cdots,x_n}$ of $X$ is shattered by $F$ if $|\br{S_f\mid f\in F}|=2^n$.
The VC-dimension of $F$ is the size of the largest subset of $X$ shattered by $F$.
\end{definition}

\begin{theorem}
\label{vc t}
Let $h$ be a function from $ \REAL^m \times  \REAL^d$ to $\{0, 1\}$, and let \begin{equation}
\Hstyled = \{ h_{\theta}: \REAL^d \to \{0, 1\}\mid \theta \in \REAL^m\}.\end{equation}
Suppose that $h$ can be computed by an algorithm that takes as input the pair $\theta \in \REAL^m \times  \REAL^d$ and returns $h_{\theta} (x)$ after no more than $t$ of the following operations:
\begin{itemize}
    \item the arithmetic operations $+,-,\times,$ and $/$ on real numbers,
    \item jumps conditioned on $>,\leq,<,\geq,=,$ and $\neq$ comparisons of real numbers, and
    \item output 0, 1.
\end{itemize}
Then the VC-dimension of $\Hstyled$ is $O\big(m^2+mt \big) $.
\end{theorem}
For the sample mentioned at the start of Section~\ref{sec: analysis of Core}, we utilize the following generalization of the previous definition of VC-dimension.
This is commonly referred to as VC-dimension, but to differentiate this definition from the previous, and to be in line with the notations in~\cite{zahimsc} we abbreviate it to \emph{dimension}.
This is the dimension induced by the query space which would be assigned in Theorem~\ref{theorem - the output of coreset is coreset} to obtain the proof of desired properties of Algorithm~\ref{alg: Coreset}.
\begin{definition} [range space \cite{bigtotiny}]
A range space is a pair $(L, \ranges)$ where $L$ is a set, called ground set and $\ranges$ is a family (set) of subsets of $L$.
\end{definition}
\begin{definition} [dimension of range spaces \cite{bigtotiny}]
The dimension of a range space $(L, \ranges)$ is the size
$\abs{S}$ of the largest subset $S \subseteq F$ such that
\begin{equation}
\abs{\br{S \cap \range \mid \range \in \ranges}} = 2^{\abs{S}}.
\end{equation}
\end{definition}

\begin{definition} [range space of functions \cite{bigtotiny},\cite{harpeledsharir},\cite{newframework}]\label{def: range space}
Let $F$ be a finite set of functions from a set $\Q$ to $[0,\infty)$. For every $Q \in \Q$ and $r \geq 0$, let $\range(F, Q, r) = \br{f \in F \mid f(Q) \geq r}$.
Let $\ranges(F) = \br{\range(F, Q, r) \mid Q \in \Q, r \geq 0}$.
Let $\R_{\Q,F} = \of{F, \ranges(F)}$ be the range space induced by $\Q$ and $F$.
\end{definition}

In the following lemma, which is inspired by 
Theorem 12 in \cite{lucic2017training}, we bound the VC-dimension which would be assigned in Theorem~\ref{theorem - the output of coreset is coreset} to obtain the proof of Algorithm~\ref{alg: Coreset}.
\begin{lemma}\label{vc lemma} 
Let $ \br{p_1,\cdots,p_n}:=P\subset\REAL^d$, and let $\Q$ be the union over all the sets of size $k$ in $\REAL^d$.
For every $i\in[n]$ let $f_i:\Q \to [0,\infty)$ denote the function that maps every $C \in \Q$ to $\displaystyle f_i(C)=\frac{1}{|P''_c|} \min_{c\in C} \| p-c\|_2$, where for every $c\in C$ we set $P''_c$ are the points in $P$ that are closest to $c$.
Let $F = \br{f_1, \ldots, f_n}$.
We have that the dimension of the range space $\R_{\Q,F}$
The dimension of the range space $\R_{\Q, F}$ that is induced by $\Q$ and $F$ is in $O(k d^3)$.
\end{lemma}
\begin{proof}
Let $\Q'$ be the union of the weighted sets of size $k$.
For every $i\in[n]$ let $f_i:\Q \to [0,\infty)$  denote the function that maps every $C \in \Q'$ to $\displaystyle f_i(C):=\min_{c\in C} w(c) \| p-c\|_2$.
It immediately follows that the dimension of the range space $\R_{\Q, F'}$ is upper bound by the dimension of the range space $\R_{\Q', F'}$.
Hence, from now on we prove that the dimension of the range space $\R_{\Q', F'}$ is in $O(k d^3)$.

Let $f': \Q' \to \REAL^{d^2+d}$ be the function that maps every $(C,w)\in \Q',\br{c_i}_{i=1}^d :=C$ to $f(C,w):=(c_1|c_2|\cdots|c_d) \mid \br{w(c)}_{c\in C}$.
For every $(q,r) = \big((C,w),r\big)\in \Q' \times  \REAL$, let $h_{(f(C,w) \mid r)} : \REAL \to \br{0,1} $ that maps every $p \in P$ to $h_{(f(C,w) \mid r)}(x) = 1 $ if and only if $\min_{c\in C} w(c) \| p-c\|_2 \leq r$, and every $p \in \REAL^d \setminus P$ to $h_{(f(C,w) \mid r)}(x) = 0 $.
Let $\Hstyled = \{ h_\theta \mid \theta \in \REAL^{d^2+d+1}\} $.
For every $(C,w)\in \Q$, and $p\in \REAL^d$ we can calculate $\min_{c\in C} w(c)^2 \| p-c\|_2^2$ with $O\big(k d\big)$ arithmetic operations on real numbers and jumps conditioned on comparisons of real numbers.

Therefore, for every $p\in P$ and any $\theta\in \REAL^{d^2+d+1}$ we can calculate $h_\theta(i)$ with $O(k d)$ arithmetic operations on real numbers and jumps conditioned on comparisons of real numbers.
Hence, substituting $d:=n$, $m:=d^2+d+1$, $h:=h$, $\Hstyled:=\Hstyled$, and $t\in O(k d)$ in Theorem~\ref{vc t} yields that the VC-dimension of $\Hstyled$ is in $O(k d^3)$.
Hence, by the construction of $\Hstyled$ and the definition of range spaces in Definition~\ref{def: range space}, we have that the dimension of the range space $\R_{\Q', F'}$ that is induced by $\Q'$ and $F'$ is in $ O(k d^3)$.
\end{proof}

\subsection{Sensitivity of functions} \label{sec: sen}
For the self-containment of this work, we state previous work on the sensitivity of functions.

\begin{definition} [query space \cite{zahimsc}] \label{definition - k query space}
Let $P \subset \REAL^d$ be a finite nonempty set.
Let $f : P \times \Q \to [0,\infty)$ and $\loss : \REAL^{|P|} \to [0, \infty)$ be a function. The tuple
$(P,\Q,f,\loss)$ is called a
\emph{query space}.
For every $q \in \Q$ we define the overall fitting error of $P$ to $q$ by
$$
f_{\loss}(P,q) := \loss \of{f(p,q)_{p \in P}}
= \loss\of{f(p_1,q),\ldots,f(p_{\abs{P}},q)}.
$$
\end{definition}

\begin{definition} [$\eps$-coreset \cite{zahimsc}] \label{rat eps coreset}
Let $(P,\Q,f,\loss)$ be a query space as in Definition~\ref{definition - k query space}. For an approximation error $\eps > 0$, the pair $S' = (S,u)$ is called an \emph{$\eps$-coreset} for the query space $(P,\Q,f,\loss)$, if $S \subseteq P, u : S \to [0, \infty)$, and for every $q \in \Q$ we have
$$
(1 - \eps)f_{\loss}(P, q) \leq f_{\loss}(S',q) \leq (1 + \eps)f_{\loss}(P, q).
$$
\end{definition}

\begin{definition} [sensitivity of functions] \label{definition - sensitivity rat}
Let $P\subset\REAL^d$ be a finite and nonempty set, and let
$F \subset \br{ P \to [0,\infty]}$ be a possibly infinite set of functions.
The \emph{sensitivity} of every point $p \in P$ is
\begin{align} \label{single line sensitivity}
S_{(P,F)}^*(p)
= \sup_{f\in F} \frac{f(p)}{\displaystyle \sum_{p\in P} f(p)},
\end{align}
where $\sup$ is over every $f \in F$ such that the denominator is positive. The \emph{total sensitivity} given a \emph{sensitivity} is defined to be the sum over these sensitivities, $S_F^*(P) = \sum_{p \in P} S_{(P,F)}^*(p)$. The function $S_{(P,F)} : P \to [0, \infty)$  is a \emph{sensitivity bound} for $S_{(P,F)}^*$, if for every $p \in P$ we have $S_{(P,F)}(p) \geq S_{(P,F)}^*(p)$. The \emph{total sensitivity bound} is then defined to be $\displaystyle S_{(P,F)}(P)= \sum_{p \in P} S_{(P,F)}(p)$.
\end{definition}
The following theorem proves that a coreset can be computed by sampling according to the sensitivity of functions. The size of the coreset depends on the total sensitivity and the complexity (dimension) of the query space, as well as the desired error $\eps$ and probability $\delta$ of failure.

\begin{theorem} [coreset construction \cite{zahimsc}] \label{theorem - the output of coreset is coreset}
Let
\begin{itemize}

\item $P = \br{p_1,\cdots,p_n}\subset \REAL^d$ be a finite and non empty set, and $f:P \times \Q \to [0,\infty)$.

\item $F = \br{f_1, \ldots, f_n}$, where $f_i(q)= f(p_i, q)$ for every $i \in [n]$ and $q\in \REAL^d$

\item $d'$ be the dimension of the range space that is induced by $\REAL^d$ and $F$.

\item $s^* : P \to [0, \infty)$ such that $s^*(p)$ is the sensitivity of every $p \in P$, after substituting $P=P$ and $F=\br{f':P\to[0,\infty] \mid \forall p\in P,q\in \REAL^d : f'(p) := f(p,q)}$ in Definition~\ref{definition - sensitivity rat}, and $s : P \to [0, \infty)$ be the sensitivity bound of $s^*$.


\item $t = \sum_{p \in P} s(p)$.

\item $\eps, \delta \in (0,1)$.

\item $c > 0$ is a universal constant that can be determined from the proof.

\item
$ \displaystyle
\lambda \geq c(t+1) \big(d' \log(t+1) + \log(1/\delta)\big/\eps^2.
$

\item $w:P\to \br{1}$, i.e. a function such that for every $p\in P$ we have $w(p)=1$.

\item $(S,u)$ be the output of a call to $\textsc{Coreset-Framework}(P, w, s, \lambda)$ (Algorithm 1 in \cite{zahimsc}).
\end{itemize}

Then the following holds
\begin{itemize}
    \item With probability at least $1- \delta$, $(S,w)$ is an subset-$\eps$-coreset of size $\abs{S}\leq \lambda$ for the query space $(F, \Q, f, \|\cdot\|_1)$; see Definition~\ref{rat eps coreset}.
\end{itemize}
\end{theorem}
\subsection{Proof of Theorem~\ref{th: coreset}} \label{sec: core proof}
In the following section, we prove Theorem~\ref{th: coreset} that states the desired properties of Algorithm~\ref{alg: Coreset}.
\begin{theorem}\label{th: coreset proof}
Let $P\subset \REAL^d$ be a set of size $n\geq 2$.
Put $\delta\in (0,1/10]$, and $\eps\in (0,1)$.
Let $(C,w)$ be the output of a call to $\Coreset(P,k,\delta,\eps)$; see Algorithm~\ref{alg: Coreset}.
Then, $|C|$ is in
\begin{equation}
|C|\in O\of{\frac{kd^3\log(k)\log^4(n)}{\eps^2} \left(\log\big(\log(k)\log(n)\big) + \log\left(\frac{1}{\delta}\right)\right)}
\subseteq O\of{\frac{kd^3\log^2k\log^5n+\log(1/\delta)}{\eps^2}},
\end{equation}
and with probability at least $1-\delta$, we have that $(C,w)$ is a $\eps$-coreset for $P$.
Moreover, the computation time of $C$ is in  $O\big(dkn \log(1/\delta) \big)$.
\end{theorem}
\begin{proof}
Let $B$ as computed in the call to $\Coreset(P,k,\delta,\eps)$.
Let $k'\in O(\log(k))$ as stated in Claim~\ref{claim: kmeans++}.
By Claim~\ref{claim: kmeans++}, for every $C_i$ computed in the call to $\Coreset(P,k,\delta,\eps)$, with probability at least $1/2$, we have
\begin{equation}
\sum_{p\in P} \min_{c\in C_i} \| p-c\|_2 \leq k' \min_{C'\subset \REAL^d,|C'|=k} \sum_{p\in P} \min_{c\in C'} \| p-c\|_2.
\end{equation}
Hence, the event that
\begin{equation}
\sum_{p\in P} \min_{c\in B} \| p-c\|_2 > k' \min_{C'\subset \REAL^d,|C'|=k} \sum_{p\in P} \min_{c\in C'} \| p-c\|_2,
\end{equation}
occurs with probability at most $(1/2)^{\big\lceil \log(2/\delta)\big\rceil}\leq \delta/2$.
Hence, with a probability of at least $1-\delta/2$, we have that
\begin{equation}
\sum_{p\in P} \min_{c\in B} \| p-c\|_2 \leq k' \min_{C'\subset \REAL^d,|C'|=k} \sum_{p\in P} \min_{c\in C'} \| p-c\|_2.
\end{equation}
Suppose this even indeed occurs.

Observe, by the choice of $\ellt$ in Definition~\ref{def: loss function 2}, that for every $Q\subset \REAL^d,|Q|=k$ we have
\begin{equation}
\ellt(Q,P) \leq \sum_{p\in P} \min_{c\in Q} \| p-c\|_2 \leq \log^2 (n) \ellt(Q,P).
\end{equation}
Hence, $B$ is an $O(\log (k) \log^2 (n))$-approximation for $P$.

Let $\alpha \in O\of{\log (k)  \log^2 (n)}$ such that $B$ is an $\alpha$-approximation to $P$.
Let $s: P \to [0,\infty)$, and $\mathcal{P}:=\br{P_c}_{c\in B}$ as computed in the call to $\Coreset(P,k,\delta,\eps)$.
Let $w': B \to [0,\infty)$ that maps every $c\in B$ to $|P_c|$.

Suppose $\ell(B,P)=0$.
That is for every $c\in B$ and $p\in P_c$ we have $p=c$.
By the construction of Algorithm~\ref{alg: Coreset}, we have that $C=B$, and for every $c\in C$ we have $w(c)=|P_c|$.
Hence, it immediately holds that $(C,w)$ is an $\eps$-coreset for $P$.
Therefore, from now on we assume this is not the case, i.e., $\ell(B, P)> 0$.

Let $\Q$ be the union over all the sets of size $k$ in $\REAL^d$.
Let $Q\in \Q$.
We have
\begin{equation}
\label{sen:1}
 \Big| \ellt(P,Q) - \ellt\big((C,w),Q\big) \Big|
 =
 \ellt(P,Q) \cdot \bigg| \frac{ \displaystyle \ellt(P,Q)-\ellt\big((B,w'),Q\big)}{ \ellt(P,Q)} 
 - \frac{ \ellt\big((C,w),Q\big) -\ellt\big((B,w'),Q\big)}{ \ellt(P,Q)} \bigg| ,
\end{equation}
which follows from reorganizing the expression and taking $\ell(P,Q)$ out of the sum.

Let $\mathcal{P}':=\br{P'_c}_{c\in Q}$ as a partition of $P'$ such that every $p\in P$ is in $P'_c$ if $\displaystyle c \in \argmin_{c'\in Q} \| c'-p\|_2$. Ties broken arbitrarily; i.e., for every $c\in Q$ we set $P'_c$ are the points in $P'$ that are closest to $c$.
Consider $q\in Q$, where $P'_q$ is non-empty.

Let $s^* :P'_q\to [0,\infty)$ such that for every $c\in B$ and $p\in P_c \cap P'_q$ we have
\begin{gather}
\label{sen:2}
s^*(p) = \frac{\displaystyle \Big| \| q-p\|_2 - \| q-c\|_2\Big|}{\log^2\big(|P'_q|+1\big) \cdot \ellt(P,Q)},
\end{gather}
which, due to the definition of $w'$, is an upper bound on the contribution of every point in $P'_q$ to the sum in Equation~\eqref{sen:1}.

Let $c\in B$ and $p\in P_c \cap P'_q$, so that
\begin{align}
\label{sen:3}
 s^*(p) & \leq  \frac{\displaystyle \Big| \| q-p\|_2 - \| q-c\|_2 \Big|}{\log^2\big(|P'_q|+1\big) \cdot \ellt(P,Q)} \\
\label{sen:4}
& \leq \frac{\| p-c\|_2}{\log^2\big(|P'_q|+1\big) \cdot \ellt(P,Q)}  \\
\label{sen:5}
&\leq \alpha \cdot \frac{\log^2\big(|P'_q|+1\big)}{\log^2\big(|P_c|+1\big)} \cdot s(p)\\
\label{sen:6}
&\leq \alpha \cdot \log^2(n+1) \cdot s(p),\\
\label{sen:7}
&\leq 4 \alpha \cdot \log^2 (n) \cdot s(p),
\end{align}
where Equation~\ref{sen:3} is by the definition of $s^*$ from Equation~\eqref{sen:2}, Equation~\eqref{sen:4} is by the reverse triangle inequality, Equation~\eqref{sen:5} is by the choice of $s$ and the definition of $B$ as an $\alpha$-approximation of $P$, Equation~\eqref{sen:6} is by reorganizing, and Equation~\eqref{sen:7} is by assigning that $n\geq 2$.

Let $\tilde{s}: P \to [0,\infty)$ such that for every $q\in Q$ and $p\in P_q$ we have $\displaystyle \tilde{s}(p) = 4 \alpha \cdot \log^2 n \cdot s(p)$.
By Equations~[\ref{sen:3}--\ref{sen:5}] we have that $\Tilde{s}$ is a sensitivity bound for $s^*$, for every $q\in Q$.

Let $\br{p_1,\cdots,p_n}:=P$, and for every $i\in[n]$ let $f_i:\Q \to [0,\infty)$ denote the function that maps every $C' \in \Q$ to $\displaystyle f_i(C')=\frac{1}{|P''_c|} \min_{c\in C'} \| p-c\|_2$, where for every $c\in C'$ we set $P''_c$ are the points in $P$ that are closest to $c$.
Let $F = \br{f_1, \ldots, f_n}$, and $d' \in O(k d^3)$ be the dimension of the range space $\R_{\Q,F}$ from Lemma~\ref{vc lemma} when assigning $P$.

Substituting the query space $\left(P,\Q, F,\|\cdot \|_1\right)$, $d'\in O\of{k d^3}$ the VC-dimension induced by $\Q$ and $F$ from Lemma~\ref{vc lemma}, the sensitivity bound $\tilde{s}$, and the total sensitivity $t = 4 \alpha \log^2(n) \sum_{p\in P} s(p) = 4 \alpha \log^2(n) \in O\of{\log(k)\cdot \log^4(n)}$ in Theorem~\ref{theorem - the output of coreset is coreset}, combined with the construction of Algorithm~\ref{alg: Coreset}, yields that there is $\lambda$ as defined in the lemma such that with probability at least $1-\delta/2$, for every $Q\in \Q$
\begin{equation}
\label{sen:6.2}
\bigg| \frac{ \displaystyle \ellt(P,Q)-\ellt\big((B,w'),Q\big)}{ \ellt(P,Q)}
 - \frac{ \ellt\big((C,w),Q\big) -\ellt\big((B,w'),Q\big)}{ \ellt(P,Q)}\bigg| \leq \eps.
\end{equation}

Combining Equation~\ref{sen:6.2} and Equation~\ref{sen:1} yields that with probability at least $1-\delta/2$ we have 
\begin{equation} \label{eq: sample coreset: first res}
    \Big| \ellt(P,Q) - \ellt\big((C,w),Q\big) \Big| \leq \eps  \ellt(P,q).
\end{equation}
That is, $(C,w)$ is an $\eps$-coreset.

Hence, by the union bound, we have with probability at least $(1-\delta/2)^2\leq 1-\delta$ we have that $(C,w)$ is an $\eps$-coreset.

\textbf{Size of $C$.}
By its construction in the call to $\Coreset(P,k,\delta,\eps)$, we have that $|B| =k$.
Recall the choice of $\lambda$ in Line~\ref{line: lambda} of Algorithm~\ref{alg: Coreset}, that is
\begin{equation}
\lambda \in O\of{\frac{\log (k) \log^4 (n)}{\eps^2} \left(k d^3 \log(k n) + \log\left(\frac{1}{\delta}\right)\right)}.
\end{equation}
The size of the sample $S$ taken in Line~\ref{line: lambda} of Algorithm~\ref{alg: Coreset} is $\lambda$.

Hence, since $C:=S\cup B$ we have that the claim on the size of $C$ holds.

\textbf{Running time.}
By Claim~\ref{claim: kmeans++}, we have that each call to $\kmeanspp(P,k)$ can be computed in $O(ndk)$ time.
Hence, since there are $\big\lceil \log(2/\delta)\big\rceil$ calls to $\kmeanspp(P,k)$, we have that $B$ can be computed in $O(ndk \log(1/\delta))$ time.
All the other computations in the call to $\Coreset(P,k,\delta,\eps)$ can be computed in $O(ndk)$ time.
Hence, the total running time of the call to $\Coreset(P,k,\delta,\eps)$ is in $O\big(ndk \log(1/\delta) \big)$.
\end{proof}

\section{Additional tests}\label{sec: appedix tests}
In the following section, we include additional tests, which were not in the main text due to space limitations.
\subsection{Tests of stand-alone imbalanced clustering}\label{test: clustering}
In the following section, we compare the proposed approximation and compression techniques to clustering methods, which aim to compute $k$ clusters in one form or another, such as $k$-means~\cite{kmeans++} and Gaussian Mixtures~\cite{zahimsc}.

We put tests for hierarchical clustering in Section~\ref{test: hierarchical} since in the following section we consider clustering as a means to an end. In contrast, in Section~\ref{test: hierarchical} the clustering is a subroutine of the algorithm considered.

\textbf{Labels. }
All the methods used from the imbalanced-learn library~\cite{Imbalanced_learn} require labels/targets, which are supplied.
In simple terms, we provide for each point to which class it belonged or was generated from.

\textbf{Approximation methods. }
In this section, we consider the following three sets of approximation methods.
In all the sets methods (i) and (ii) are identical, the name of the sets is defined by the other methods.
I.e., \textbf{approximation} consists of methods that do not change the data set, \textbf{under-sample} consists of methods that decrease the data set to obtain balanced sets, and \textbf{over-sample} consists of methods that increase the data set to obtain balanced sets.

\textbf{Approximation.}\\
(i). The output of $\Approx((P,w),k)$, where $P$ denotes the data and $w:P\to \br{1}$ (each point is mapped to $1$); see Definition~\ref{alg: approx}. This is denoted by $\Approx$.\\
(ii). The output of $\Approx((P',w'),k)$, where $(P',w')$ is the output of a call to Algorithm~\ref{alg: Coreset} over the data. This is denoted by $\approxoncore$.\\
(iii). kmenas++~\cite{kmeans++} as implemented in scikit-learn~\cite{SciPy}, denoted by $\kmeans$.\\
(iv). kmenas++ as implemented by RAPIDS to utilize GPU~\cite{raschka2020machine}, denoted by $\kmeansgpu$.\\
(v). Our implementation of Algorithm~\ref{alg: Bicriteria}, denoted by $\Bicriteria$.\\
(vi). Gaussian mixture fitting as implemented in scikit-learn, with the initialization set to kmenas++, denoted by $\gaussian$.

\textbf{Under-sample.}\\
From now on, when we state kmenas++, we refer to the scikit-learn implementation.\\
(i). The previously stated $\Approx$.\\
(ii). The previously stated $\approxoncore$.\\
(iii). kmenas++ computed over \texttt{RandomUnderSampler} from the imbalanced-learn library~\cite{Imbalanced_learn}, denoted by $\randU$.\\
(iv--vi). kmenas++ computed over \texttt{NearMiss}, which is based on~\cite{kNN_imbalanced}, from the imbalanced-learn library.
Since at the time of writing, \texttt{NearMiss} at the imbalanced-learn library had $3$ versions, we utilized all the versions, denoted by $\nearO,\nearTw,\nearTh$.

\textbf{Over-sample.}\\
(i). The previously stated $\Approx$.\\
(ii). The previously stated $\approxoncore$.\\
(iii). kmenas++ computed over \texttt{RandomOverSampler} from the imbalanced-learn library, denoted by \\$\randO$.\\
(iv). kmenas++ computed over the commonly used \texttt{SMOTE}~\cite{SMOTE}, from the imbalanced-learn library. Denoted by $\SMOTE$.\\
(v). kmenas++ computed over \texttt{KMeansSMOTE}~\cite{kmeans_smote}, from the imbalanced-learn library. Denoted by \\$\KMeansSMOTE$.\\
(vi). kmenas++ computed over \texttt{SMOTETomek}, which was presented at~\cite{SMOTETomek}, from the imbalanced-learn library. Denoted by $\SMOTETomek$.
Note that this method utilizes a combination of over-sampling and under-sampling.

\textbf{Loss function.}
In this section, for each dataset $P\subset\REAL^2$ and its approximation $C\subset \REAL^d$, we use $\ell(P, C)$ as the loss.
For $\gaussian$, for a dataset $P\subset\REAL^2$, for each set in the partition of $P$ corresponding to the model (each point goes to its most probable Gaussian probability, ties broken arbitrarily), we set the center of the partition to be its mean.
For readability, for each set generated, we divide all the losses by the loss of $\Approx$.
It should be emphasized that our methods attempt to optimize the loss function from Definition~\ref{def: loss function 2}, and not Definition~\ref{def: loss function}.

\subsubsection{Synthetic.} \label{test: clustering: synthetic}
In the following test, inspired by the motivation in Section~\ref{sec: motivation}, we consider the following data sets in $\REAL^2$, where we use different values for $n$, where we vary the size of $n$ across the test, which are the union of $2$ uniform samples from discs generated as follows.\\
(i). A sample of $25 n$ ``inliers" points, where each point is chosen uniformly inside the unit disc.\\
(ii). A sample of $n$ ``outliers" points, where each point is chosen uniformly inside a disc of radius $0.1$, centered at $(2.25,0)$.

We utilize the set $\displaystyle \br{5 i}_{i=1}^{10}$ as the values for $x$.

\begin{figure}[h!]
    \centering
\includegraphics[width=0.495\textwidth]{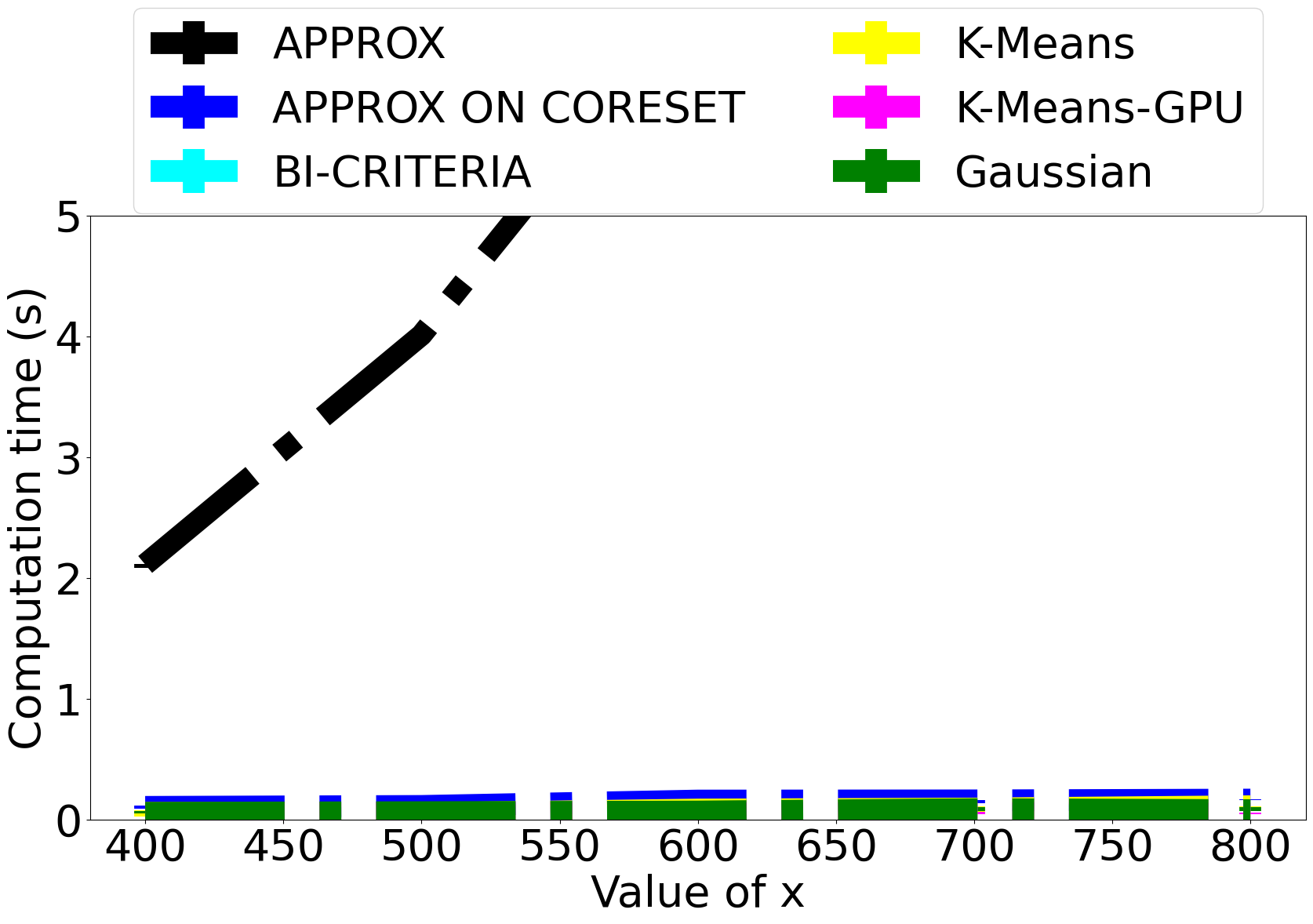}
\includegraphics[width=0.495\textwidth]{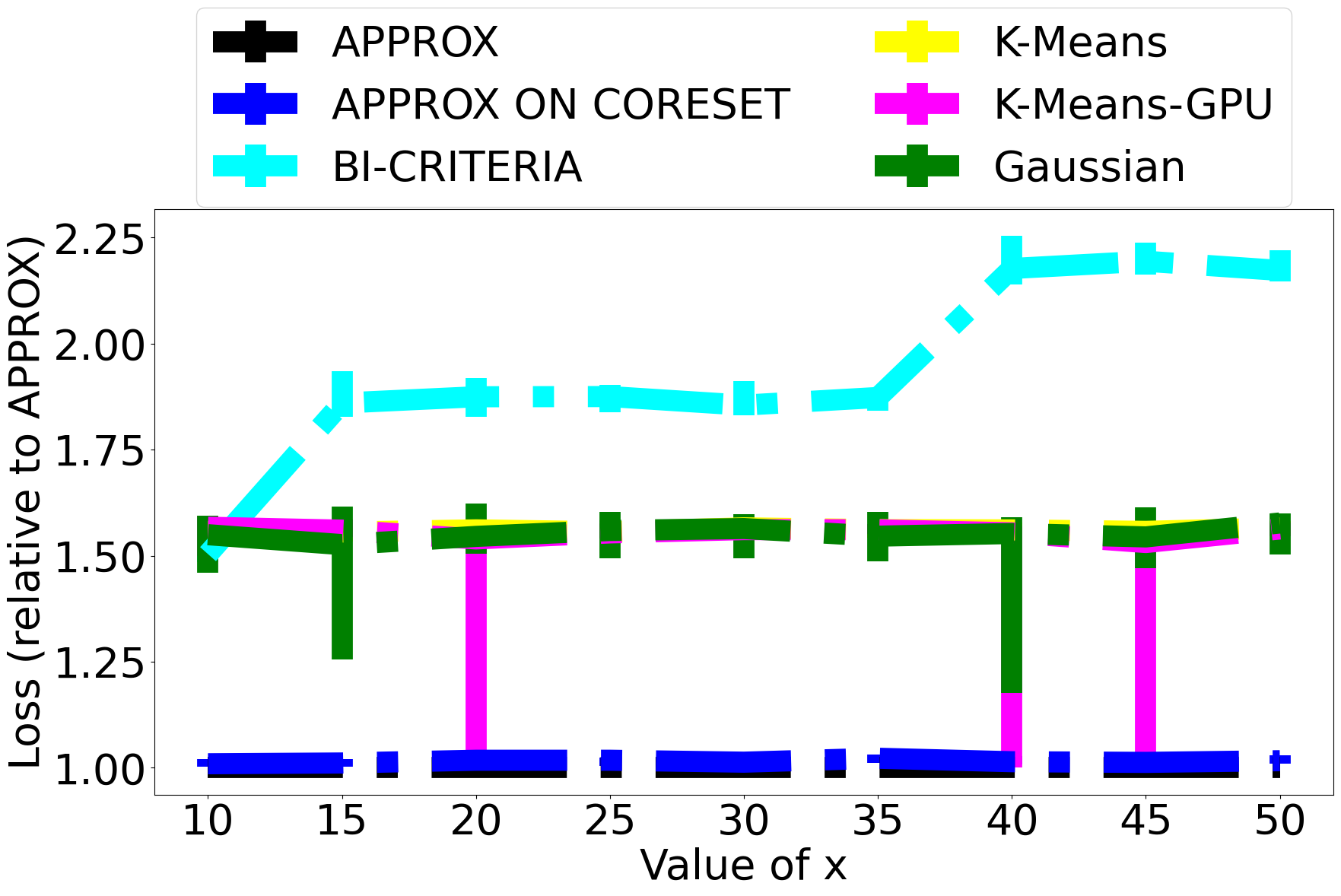}
    
\includegraphics[width=0.495\textwidth]{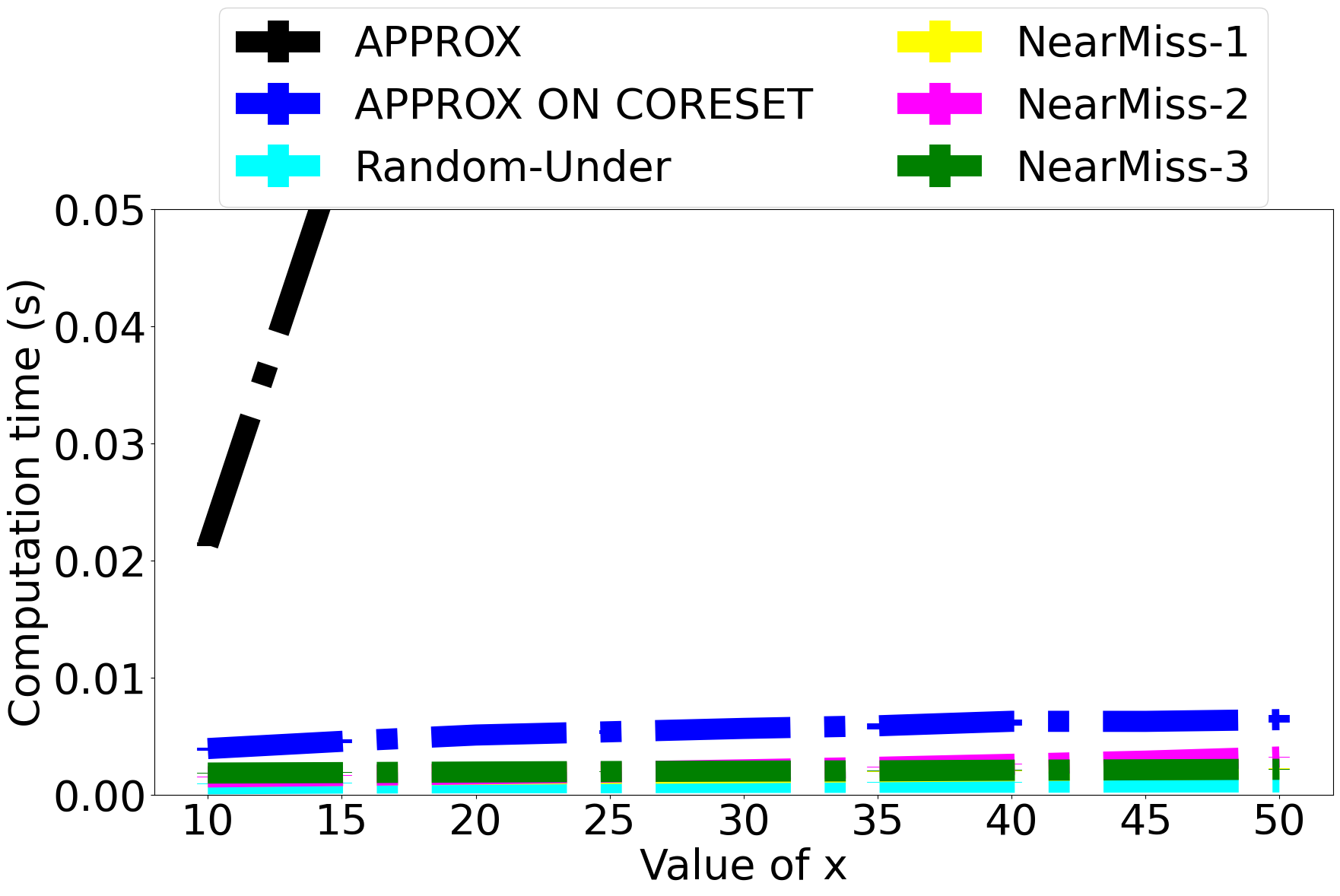}
\includegraphics[width=0.495\textwidth]{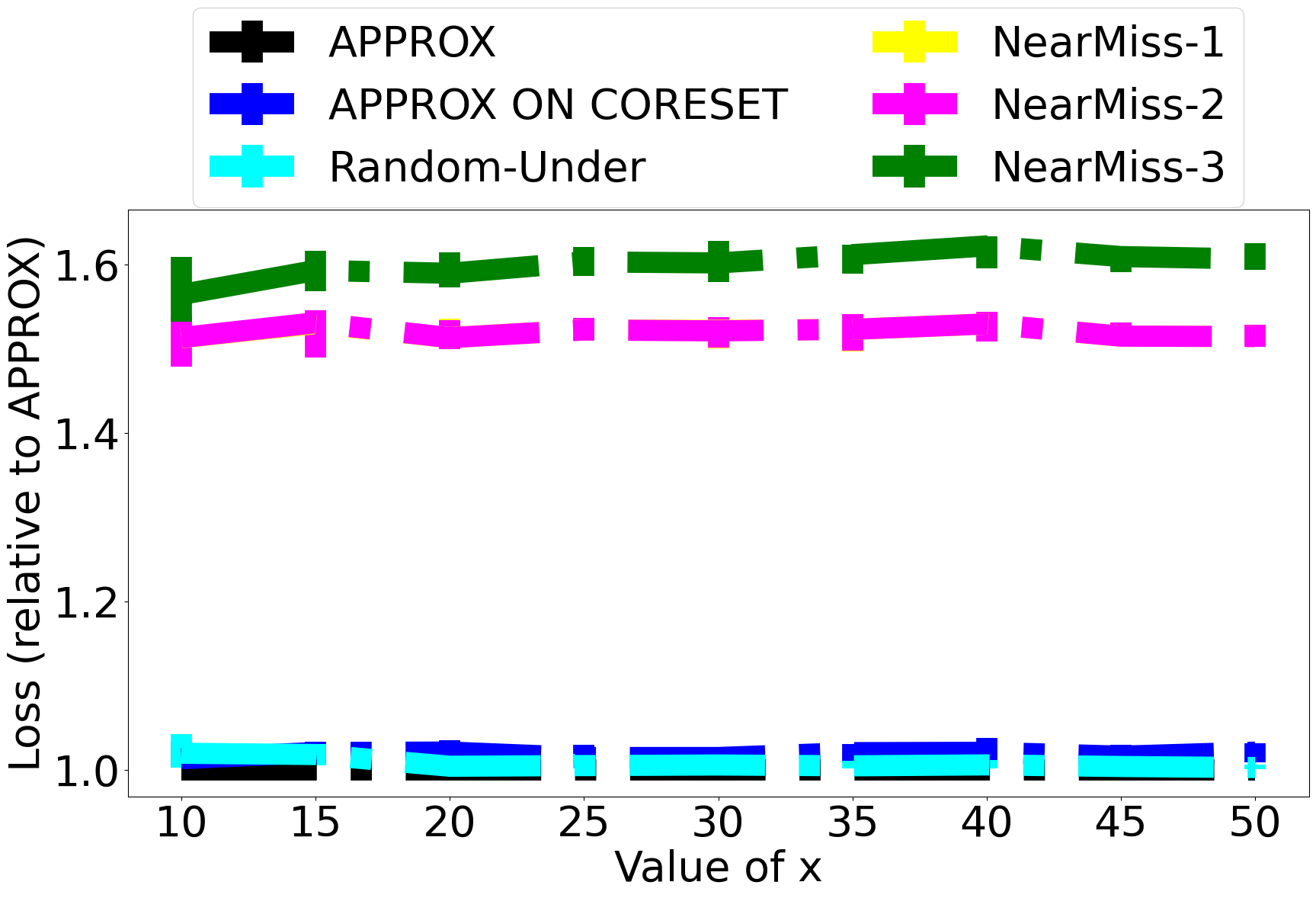}
    
\includegraphics[width=0.495\textwidth]{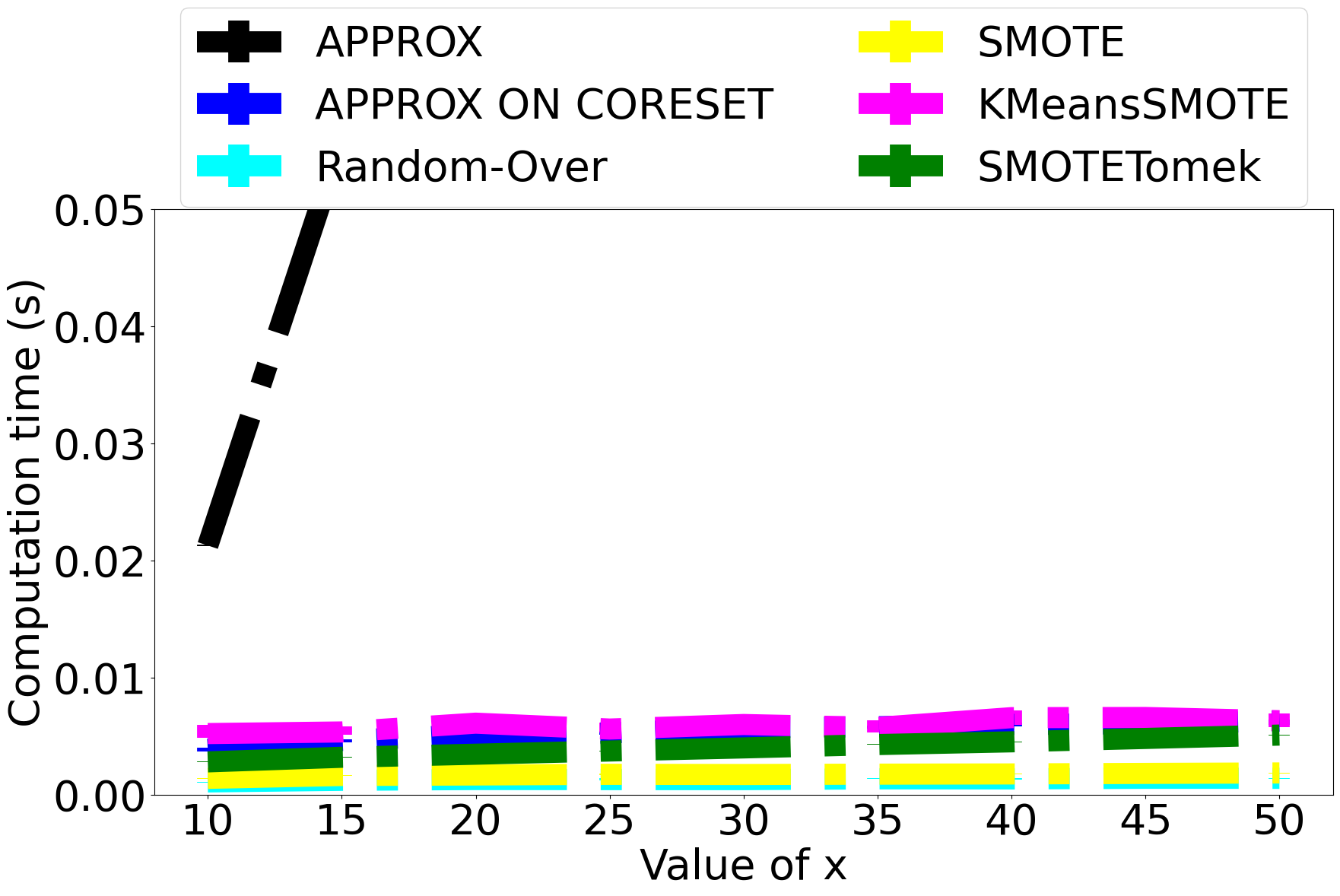}
\includegraphics[width=0.495\textwidth]{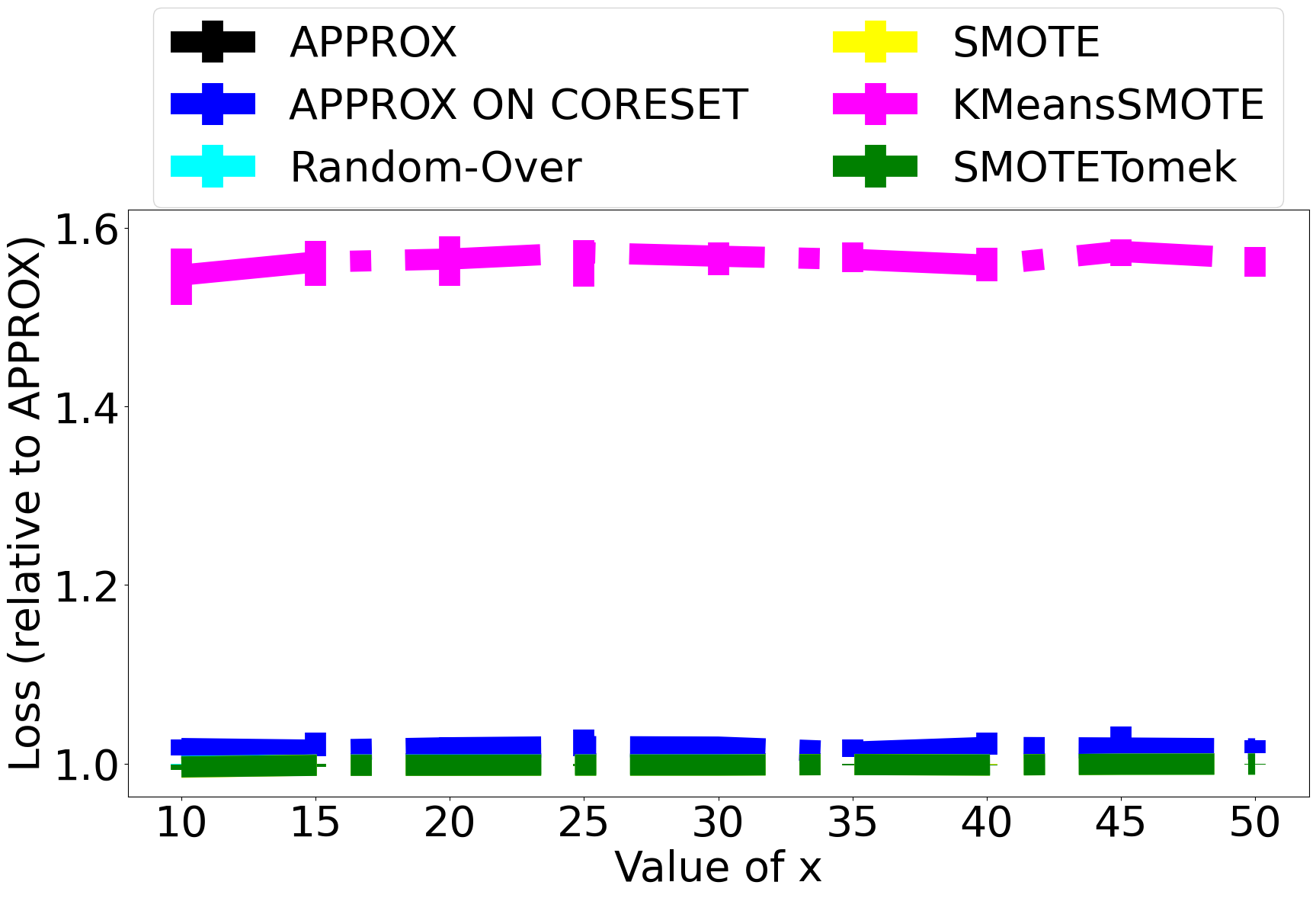}
    \caption{The results for Section~\ref{test: clustering: synthetic}.
The left column corresponds to the running time of the method, and the right column corresponds to the loss of the methods, normalized for $\Approx$ to be $1$.
The rows correspond, top to bottom, to \textbf{Top:} Approximation, \textbf{Middle:} Under-sample, and \textbf{Bottom:} Over-sample.
    }
    \label{fig: res: synthetic mean}
\end{figure}

The results are presented in Figure~\ref{fig: res: synthetic mean}.
The test was repeated $40$ times, and in all the figures the values presented are the medians across the tests, along with error bars that present the $25\%$ and $75\%$ percentiles.

The time taken to compute the optimal solution is rather large.
Nonetheless, the time taken for $\Bicriteria$ and $\approxoncore$ is significantly lower and is in line with the other method tested.

For the Approximation methods, $\approxoncore$ and $\Approx$ have yielded a noticeable quality improvement compared to $\kmeans$.
Notably, $\approxoncore$ has yielded an almost unnoticeable quality decrease compared to $\Approx$.
While, $\Bicriteria$, due to returning more clusters yielded a higher loss than the other methods.

For the under-sampling methods, we obtained that $\randU$ yielded essentially identical loss to our methods tested.
This is rather logical since the method is given the labels, which are essentially the circles to which each point corresponds.
Thus, by resampling the points such that in each circle we would have an equal number of points, kmeans++ should classify the data correctly and yield low loss.
Nonetheless, the other resampling methods have not yielded such improvement, and seem to have made the approximation worse than if they were not used at all.

For the over-sampling methods, we obtained that all the competing methods yielded essentially the same performance as $\Approx$, besides $\KMeansSMOTE$ which had a noticeably larger loss.
Nonetheless, note that all the competing methods were given the labels of the points, and have increased the size of the data.

\subsubsection{Real world data.} \label{test: clustering: real}
In the following tests, we utilize the webpage data set, which was proposed in~\cite{webpage}.
All the $34780$ entries in this data have $300$ features.

In each test iteration (done separately across the approximation sets) we uniformly, without repetitions, sample values from the data set.
Specifically, we sample the values in the set $\displaystyle \br{100 i}_{i=4}^{8}$; chosen to not obtain error for most of the competing methods from the imbalanced-learn library.

\begin{figure}[h!]
    \centering
    \includegraphics[width=0.32925\textwidth]{figures/res/mean/real/approx_time.png}
    \includegraphics[width=0.32925\textwidth]{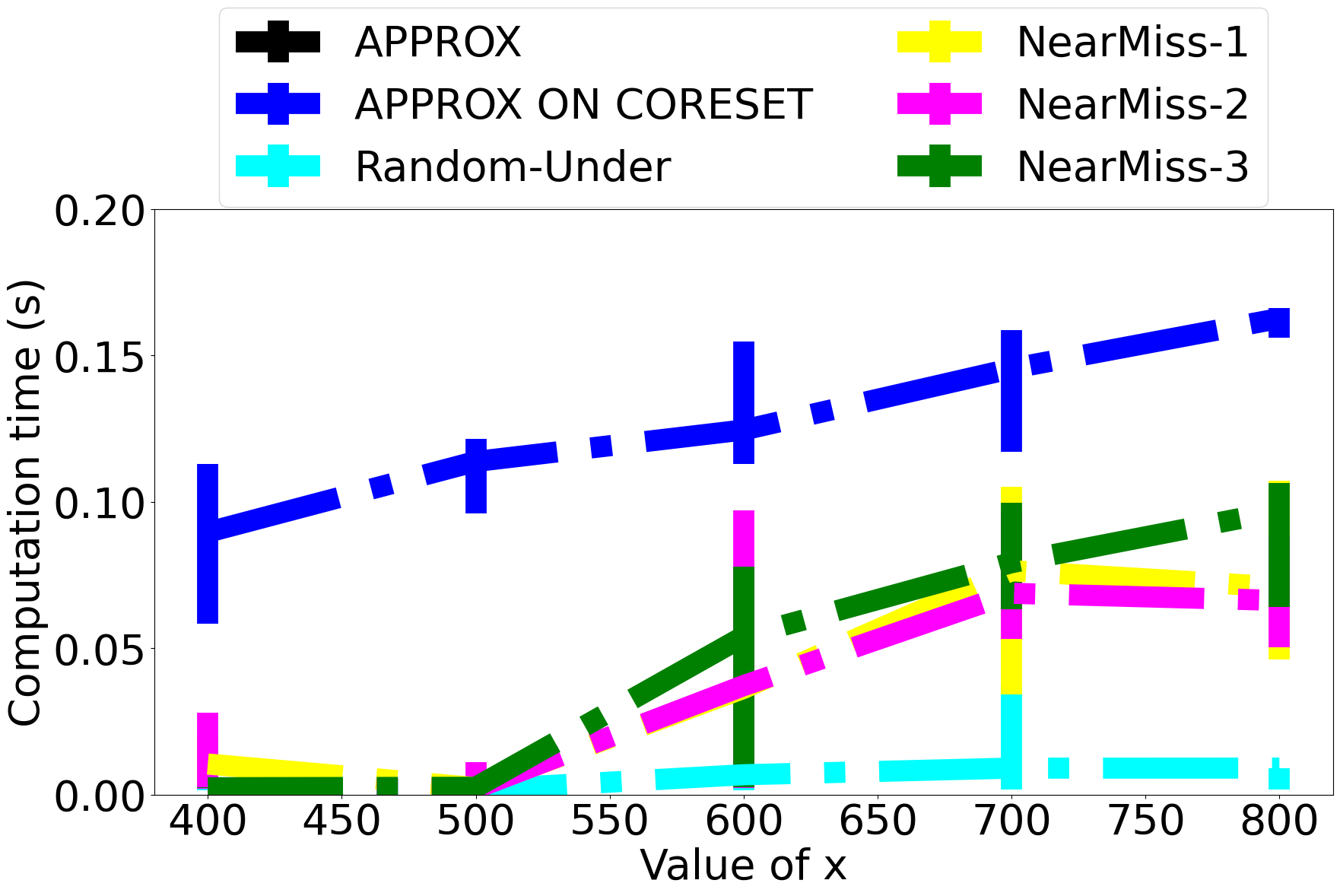}
    \includegraphics[width=0.32925\textwidth]{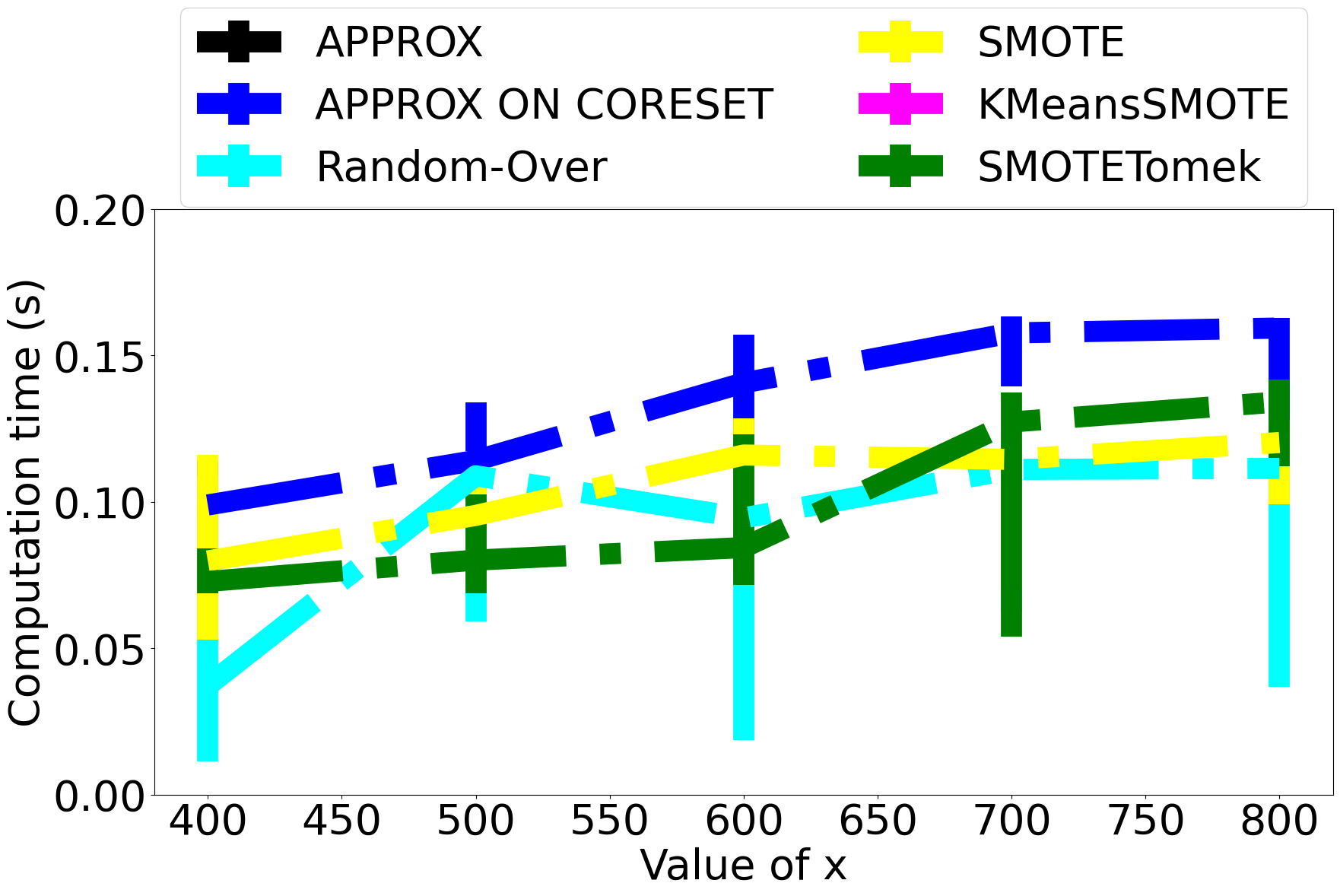}
    
    \includegraphics[width=0.32925\textwidth]{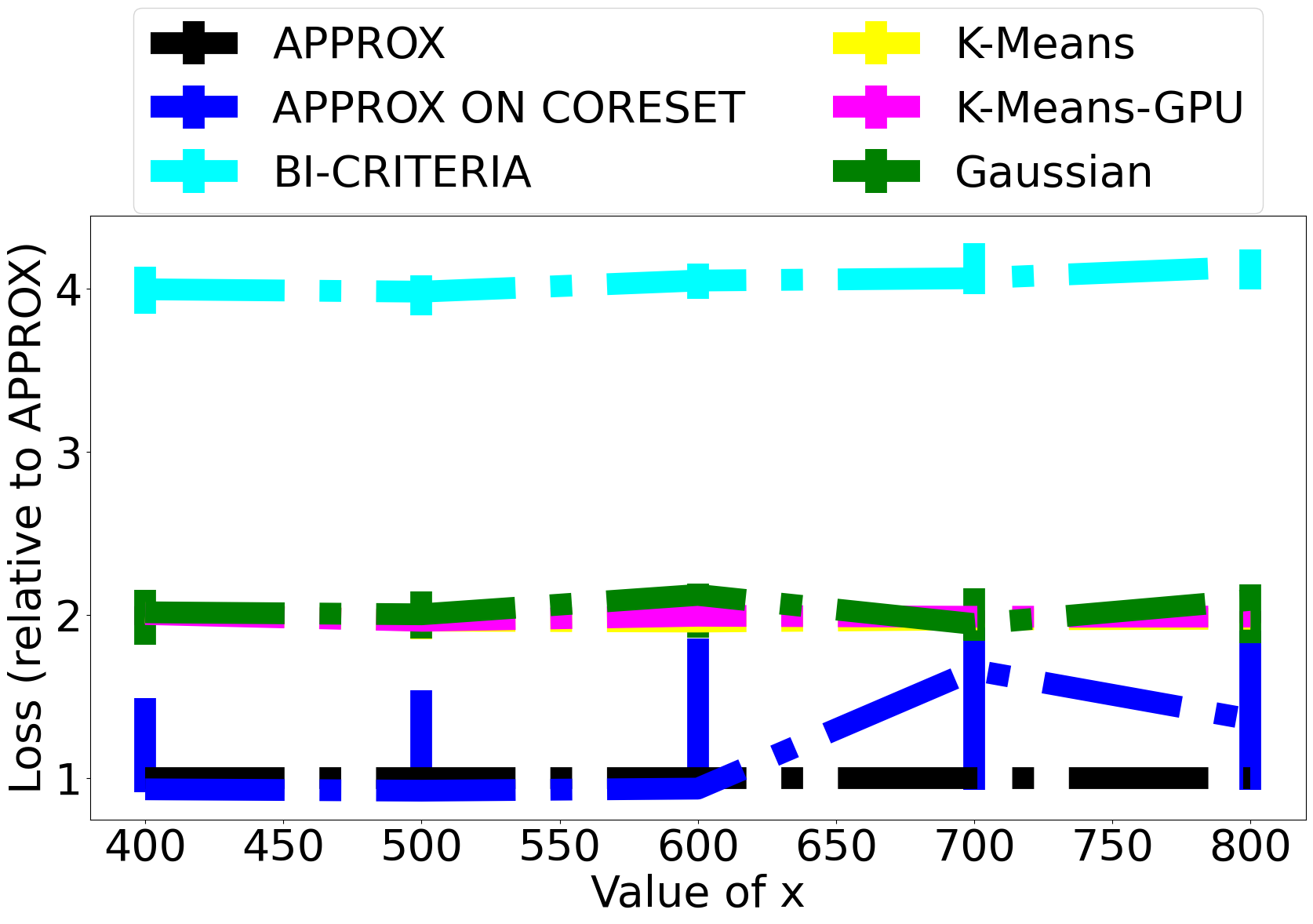}
    \includegraphics[width=0.32925\textwidth]{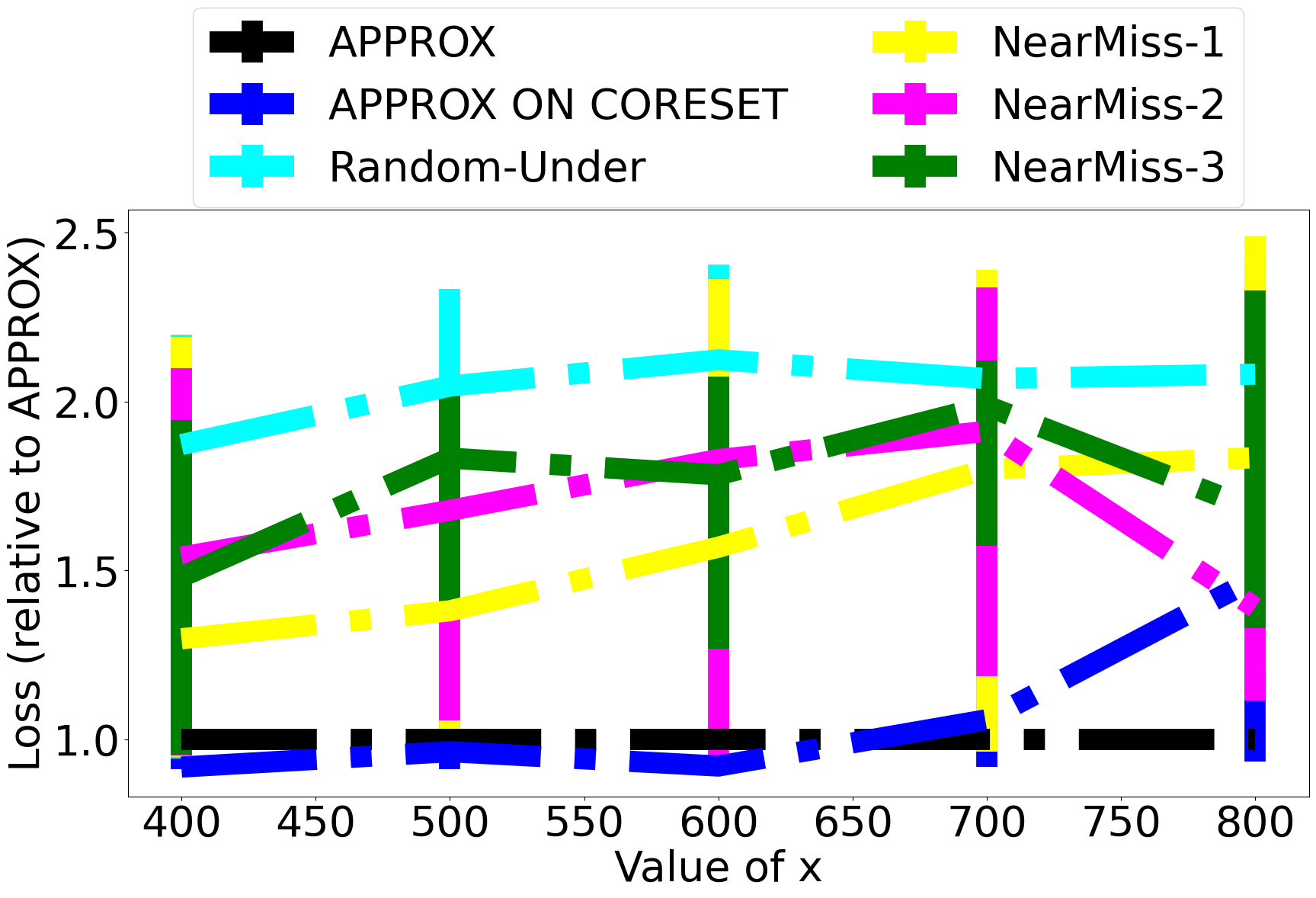}
    \includegraphics[width=0.32925\textwidth]{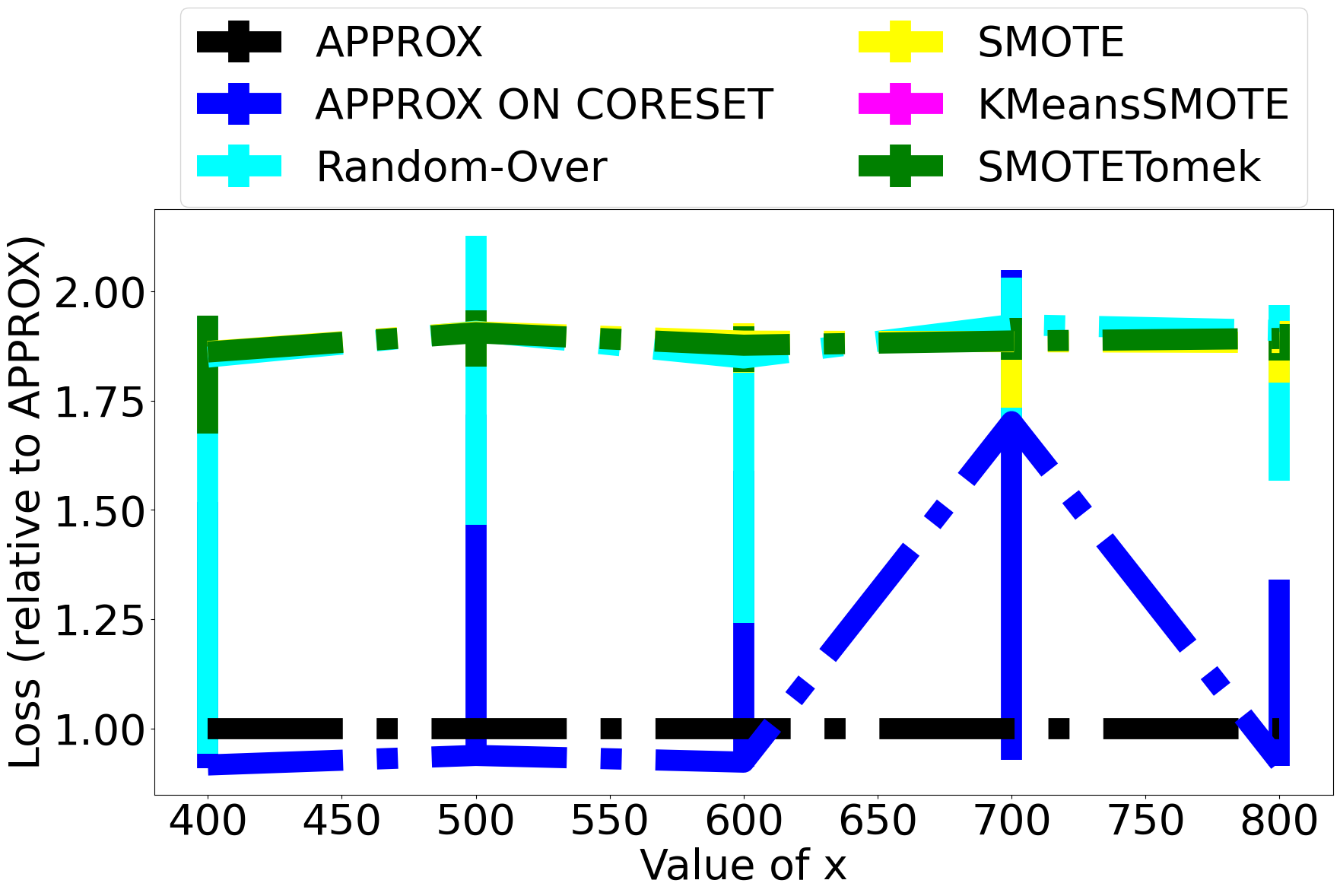}
    
    \caption{The results for Section~\ref{test: clustering: real}.
    The top row corresponds to the running time of the method, and the bottom row corresponds to the loss of the methods, normalized for $\Approx$ to be $1$.
    The columns correspond, left to right, to \textbf{Left:} Approximation, \textbf{Middle:} Under-sample, and \textbf{Right:} Over-sample.
    }
    \label{fig: res: real mean}
\end{figure}

The results are presented in Figure~\ref{fig: res: real mean}.
The test was repeated $40$ times, and in all the figures the values presented are the medians across the tests, along with error bars that present the $25\%$ and $75\%$ percentiles.

We have obtained parallel results to the test in Section~\ref{test: clustering: synthetic}.
To summarise, $\approxoncore$ and $\Approx$ yielded the lowest loss and improved the results noticeably comparable to all the other methods (including the ones that are given the data set classes, while we do not use them).
This is in contrast to Section~\ref{test: clustering: synthetic}, where a few methods (that utilized labels) from the imbalanced-learn library~\cite{Imbalanced_learn} gave comparable results to $\approxoncore$ and $\Approx$.

Notably, our approximation over the coreset gave better results than the approximation over the entire data set.
That is, $\approxoncore$ gave better quality than $\Approx$.
A similar occurrence, where coreset improves the results, was previously noted in the coreset literature.
\clearpage
\subsection{Imbalanced clustering in hierarchical clustering} \label{test: hierarchical}
In the following section, we demonstrate the improvement of our proposed approximation and compression techniques to hierarchical clustering methods, specifically divisive hierarchical clustering.

Note that we do not address comparisons to agglomerative clustering, and use our method as a subroutine of divisive hierarchical clustering and not as a stand-alone one.
As noted by a reviewer such a comparison has merit and we hope to add such one in future versions.

\subsubsection{Tests setup}
In this section, we compare the following hierarchical clustering methods.\\
(i). Divisive clustering, where each split is done via $\approxoncore$.
That is, we compute $2$ centers with $\approxoncore$, and split the data by to which center each point is closest (euclidean distance).
Denoted by \\$\approxoncore$.\\
(ii). Divisive clustering, where each split is done via $\kmeans$.
That is, we compute $2$ centers with $\kmeans$, and split the data by to which center each point is closest.
Denoted by $\kmeans$.\\
(iii). Divisive clustering, where each split is done via $\Bicriteria$.
That is, we compute centers with $\Bicriteria$ and split the data by to which center each point is closest.
Denoted by $\Bicriteria$.\\
(iv). Divisive clustering, where each split is done via by computing $2$ centers by a call to.
That is, we compute $2$ centers with a call to \\$\texttt{Clustering.SpectralClustering}$ from skict-learn~\cite{scikit-learn}, and split the data by to which center each point is closest.
Denoted by $\spectral$.\\
(v). The output of the dePDDP divisive clustering from ~\cite{Divisive_Toolbox}, denoted by $\dePDDP$.\\
(vi). The output of the ward version of the agglomeration clustering~\cite{Divisive_Toolbox}, denoted by $\ward$.

In all the tests the loss is the sum of the variances of the clusters computed.

In all the following tests we utilize $3$ splits for all the methods, i.e., each method partitions the data to $8$ sets.
\subsubsection{Synthetic data.} \label{test: hierarchical: synthetic}
In the following test, we generate the data by a union of the following $3$ sets, where we vary the size of $x$ across the test.\\
(i). A sample of $250 x$ points from a circle in $\REAL^2$, where each point is chosen uniformly from the area inside the unit circle.\\
(ii). A sample of $10 x$ points from a circle in $\REAL^2$, where each point is chosen uniformly from the area inside a circle with center in $(2.25,0)$ and radius $0.1$.\\
(ii). A sample of $10 x$ points from a circle in $\REAL^2$, where each point is chosen uniformly from the area inside a circle with center in $(2.25,2.25)$ and radius $0.1$.\\

We utilize the set $\displaystyle \br{i}_{i=1}^{10}$ as the values for $x$.

\begin{figure}[h!]
    \centering
    \includegraphics[width=0.495\textwidth]{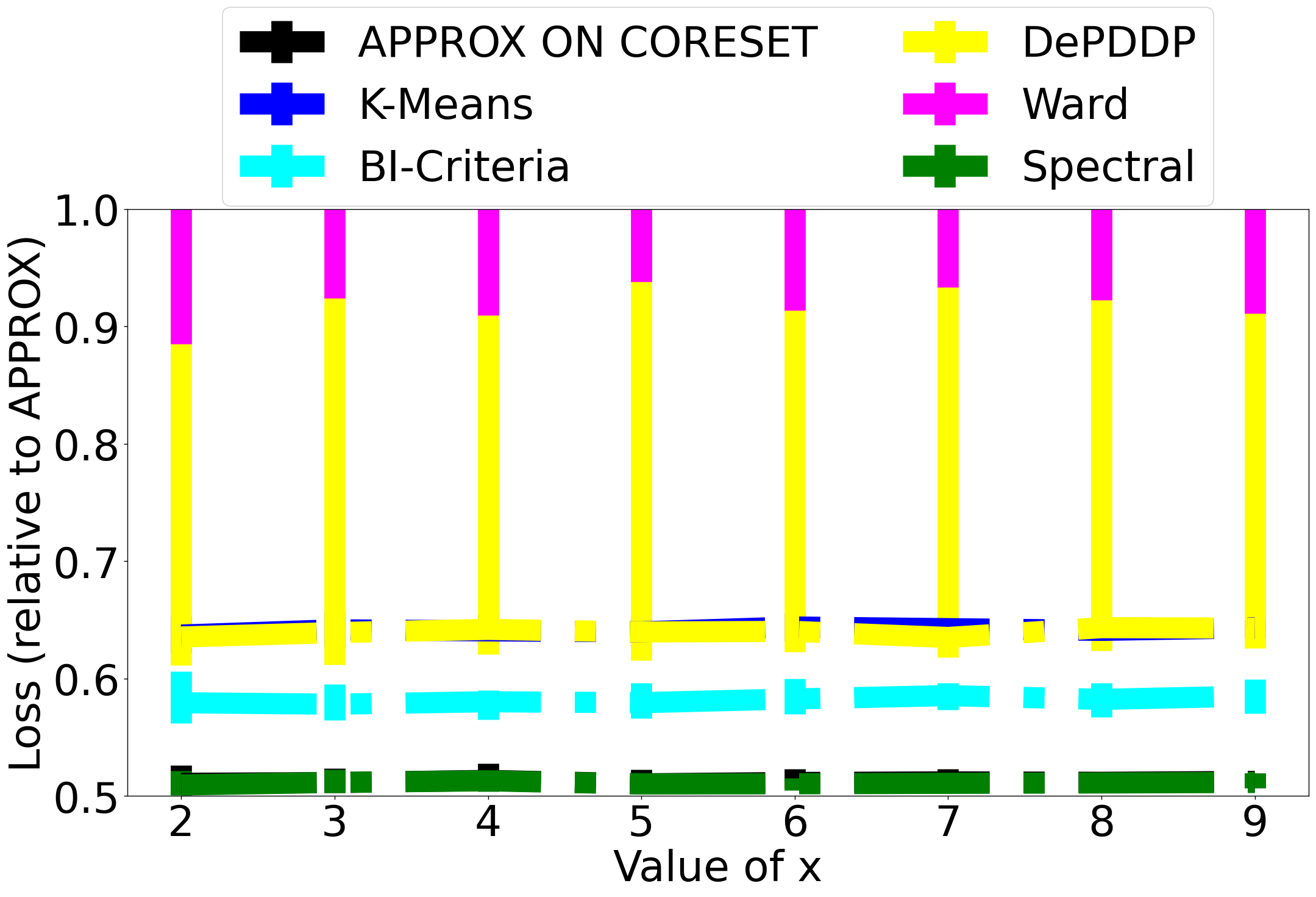}
    \includegraphics[width=0.495\textwidth]{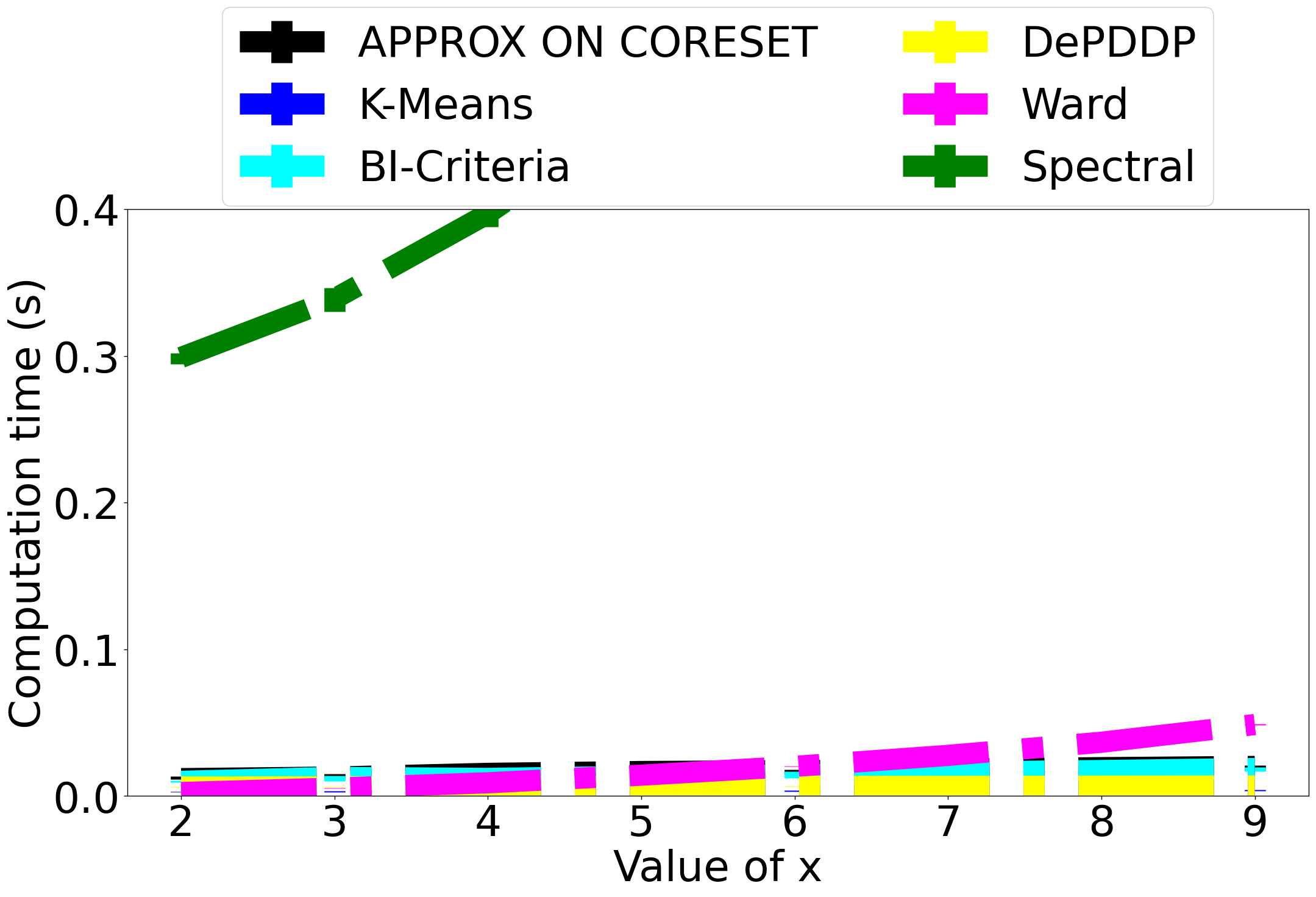}

    \caption{The results for Section~\ref{test: hierarchical: synthetic}.
    The left figure is the loss (sum of cluster variance) plot, and the right plot is the time plot.
    }
    \label{fig: hierarchical: synthetic}
\end{figure}

The results are presented in Figure~\ref{fig: hierarchical: synthetic}.
The test was repeated $100$ times, and in all the figures the values presented are the medians across the tests, along with error bars that present the $25\%$ and $75\%$ percentiles.

As can be seen, the best results were for $\spectral,\approxoncore$, followed by $\Bicriteria$ and $\kmeans,\dePDDP$ consecutively.
Nonetheless, $\spectral$ and $\approxoncore$ yield noticeably lower loss.
Moreover, observe that most of the competing methods have similar running times, except $\spectral$ which is notably slower, and $\ward$ which for the larger values of $x$ becomes noticeably slower.
It is expected that $\ward$ takes noticeably longer for larger values of $x$, due to its quadratic time complexity, while $\kmeans,\approxoncore,\approxoncore$ all have semi-linear time complexity.

Hence, the only other method which has a similar quality to our method, i.e., $\spectral$, is noticeably slower.
\clearpage

\subsubsection{Real world data.} \label{test: hierarchical: real}
In the following tests, we utilize the webpage data set, which was proposed in~\cite{webpage}.
All the $34780$ entries in this data have $300$ features.

In each test iteration (done separately across the tests) we uniformly, without repetitions, sample values from the data set.
Specifically, we sample the values in the set $\displaystyle \br{250 i}_{i=2}^{10}$; chosen to not obtain error for most of the competing methods.

\begin{figure}[h!]
    \centering
    \includegraphics[width=0.495\textwidth]{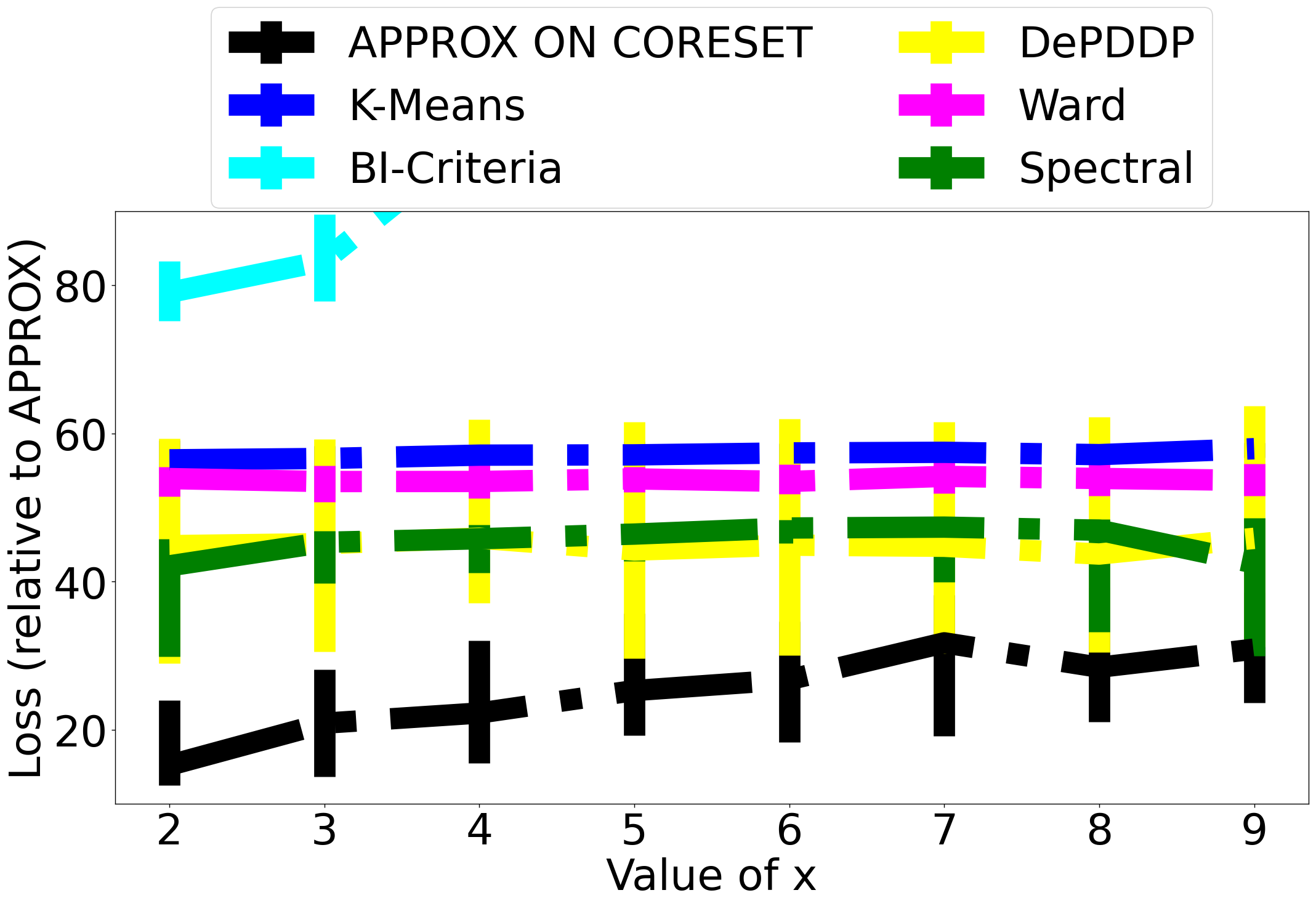}
    \includegraphics[width=0.495\textwidth]{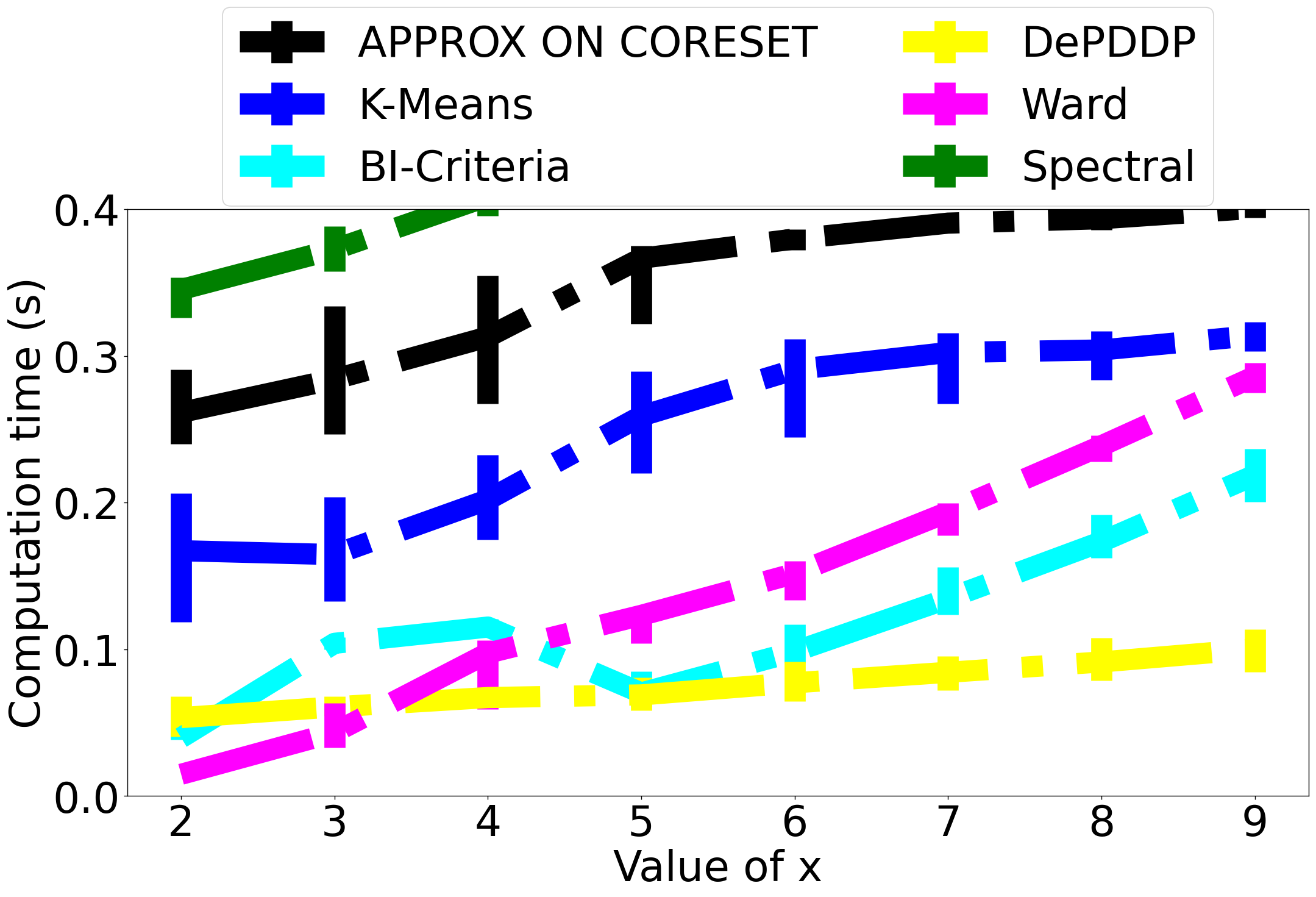}

    \caption{The results for Section~\ref{test: hierarchical: real}.
    The left figure is the loss (sum of cluster variance) plot, and the right plot is the time plot.
    }
    \label{fig: hierarchical: real}
\end{figure}

The results are presented in Figure~\ref{fig: hierarchical: real}.
The test was repeated $100$ times, and in all the figures the values presented are the medians across the tests, along with error bars that present the $25\%$ and $75\%$ percentiles.

As can be seen, $\approxoncore$ has yielded the best results by a noticeable margin.
While \\ $\approxoncore$ has a noticeably slower running time than most other methods, as can be expected since the next faster method was $\kmeans$, where the computation of the $k$-means is utilized internally in $\approxoncore$.
Hence, we obtained the best results at the price of a longer running time.
Nonetheless, note that our implementation is less optimized than the other competing methods, such as $\kmeans$ which consists mostly of a call to k-means++ implemented by Scit-Learn.

\end{document}